\documentclass[twocolumn]{article}

\usepackage[english]{babel}
\usepackage[a4paper,top=2.5cm,bottom=2.5cm,left=1.5cm,right=1.5cm,marginparwidth=1.25cm,columnsep=0.75cm]{geometry}
\usepackage{svg}
\usepackage{amsmath}
\usepackage{graphicx}     
\usepackage{caption}     
\usepackage{subcaption}   
\usepackage{float}     
\usepackage{multirow}
\usepackage[colorlinks=true, allcolors=blue]{hyperref}
\usepackage{float}
\usepackage{authblk}
\usepackage{booktabs}
\usepackage{threeparttable} 
\usepackage{float} 
\usepackage[ruled,vlined]{algorithm2e}
\usepackage{cuted}
\usepackage{tabularx} % in preamble
\usepackage{makecell}
\usepackage{algpseudocode}
\usepackage{geometry}
\usepackage{amssymb}
\usepackage{pgfplots}
\pgfplotsset{compat=1.18}
\usepackage{tikz}
\usepackage{booktabs}
\usepackage{longtable,array}
\usepackage{calc}
\providecommand{\keywords}[1]
{
  \small	
  \textbf{\textit{Key words---}} #1
}

\title{\textbf{BrainRotViT: Transformer-ResNet Hybrid for Explainable Modeling of Brain Aging from 3D sMRI}}
\author[1]{Wasif Jalal}
\author[1]{Md Nafiu Rahman}
\author[1,2]{Atif Hasan Rahman}
\author[1,2]{M. Sohel Rahman}

\affil[1]{\small Department of Computer Science and Engineering, Bangladesh University of Engineering and Technology}

\affil[2]{\small Corresponding authors: 
\href{mailto:atif@cse.buet.ac.bd}{atif@cse.buet.ac.bd},
\href{mailto:msrahman@cse.buet.ac.bd}{msrahman@cse.buet.ac.bd}}
\date{}

\begin{document}

\twocolumn[

\begin{@twocolumnfalse}

\maketitle

\begin{abstract}
Accurate brain-age estimation from structural MRI provides a valuable biomarker for studying healthy aging and neurodegenerative disease. Conventional CNN-based approaches are constrained by limited receptive fields and susceptibility to overfitting on heterogeneous multi-site data, whereas pure transformer architectures are computationally demanding and require large-scale training data. We propose \textit{\textbf{Brain} \textbf{R}esNet \textbf{o}ver \textbf{t}rained \textbf{Vi}sion \textbf{T}ransformer (\textbf{BrainRotViT})}, a two-stage, slice-based (2.5D) framework that approximates volumetric context. A Vision Transformer is first pretrained on an auxiliary age--sex composite classification task and subsequently frozen; its per-slice embeddings are stacked into a 160$\times$768 feature matrix and processed as a pseudo-image by a lightweight 2D residual CNN regressor, with subject sex incorporated at the final prediction layer. Trained and validated on 5,108 samples from 11 multi-site datasets and tested on four independent fully held-out cohorts comprising 948 samples, the proposed model achieves a mean absolute error (MAE) of 3.43 years (Pearson $r = 0.98$, $R^2 = 0.96$, Spearman $\rho = 0.97$) and generalizes across the independent cohorts with MAEs of 4.79--5.35 years, outperforming retrained 3D-ResNet, SFCN, 3D-ViT, Global-Local Transformer, Triamese-ViT, and TSAN baselines. Following age-bias correction and covariate adjustment, the brain-age gap is significantly associated with Alzheimer's disease, mild cognitive impairment, and autism spectrum disorder. An integrated interpretability pipeline combining ViT attention maps with guided backpropagation, validated through saliency sanity checks and anatomical localisation using the AAL3 atlas, identifies aging-related regions including the occipital lobe, cerebellum, precentral and postcentral gyri, temporal lobes, and medial superior frontal gyrus. These findings demonstrate that BrainRotViT provides an efficient and interpretable bridge between CNN and transformer paradigms, achieving moderate to strong cross-cohort generalization while yielding biologically meaningful insights into brain aging. 
\\
\end{abstract}

% \keywords{Deep learning, Biomedical image analysis, Brain aging, MRI, Vision transformer, ResNet, Convolutional neural network, Hybrid model, Neurodegeneration}

\vspace{5em} % optional space before 2-column text starts

\end{@twocolumnfalse}
]

\section{Introduction}\label{sec:introduction}
The human brain undergoes continuous transformations across the lifespan, representing a natural component of aging that does not inherently signal pathological conditions \cite{FOX2004392}. Neurodegenerative disorders such as dementia can compromise the brain structure and accelerate aging processes. Understanding and characterizing healthy brain aging patterns therefore becomes essential for distinguishing normal aging from pathological neurodegeneration, potentially enabling earlier detection of neurodegenerative diseases. The Brain Age-Gap (BAG), i.e. the discrepancy between predicted brain age and chronological age, has emerged as a robust biomarker that captures pathological brain processes and offers insights into the rate at which an individual's brain ages in comparison to others in the population \cite{franke2019ten,zhang2025brain}. It is not only associated with various neurological disorders, such as Alzheimer's disease, cognitive impairment, and Autism Spectrum Disorder, but also serves as an indicator of all-cause mortality \cite{millar2023multimodal,wittens2024brain,torenvliet2023longitudinal,wang2021predicting,cole2020multimodality} 
Brain age estimation has been approached through both conventional and machine learning techniques, analyzing either the whole brain, specific regions, or localized patches \cite{FRANKE2010883,behesti2021patch,COLE2017115}. One particular study presented a method using T1-weighted MRI to predict age through region-level and voxel-level metrics \cite{Beheshti2022BrainAge}. Regression-based machine learning has shown promise for the brain age prediction, with kernel regression applied to whole-brain MRI across diverse age ranges \cite{FRANKE20121305}. Various algorithms including Support Vector Regression and Binary Decision Trees have been compared for their brain age prediction capabilities \cite{9439893}. Additional regression techniques such as Relevance Vector Regression, Twin Support Vector Regression, and Gaussian Process Regression have been explored across different imaging modalities for age estimation and mortality prediction \cite{COLE2017115,10.3389/fnagi.2018.00317,Ganaie2024,Cole2018}. 
A more recent addition to this family, the improved least-squares twin SVR \cite{ganaie2023lstsvr}, reformulates the twin-SVR objective to avoid explicit matrix inversion, lowering  omputational cost while retaining competitive accuracy on T1-weighted brain-age estimation. Together with the comprehensive benchmarking of conventional regressors in \cite{9439893} and the twin-SVR study in \cite{Ganaie2024}, these works indicate that carefully designed classical models can approach deep-network accuracy at a fraction of the training cost. This favourable trade-off between accuracy and computational cost directly motivates the lightweight regression head adopted in our framework. However, while these conventional machine learning approaches achieve reasonable performance in brain age estimation, they are limited by the need for manual feature engineering. Deep learning offers automatic feature extraction capabilities, making it particularly suited for handling unstructured data and addressing complex analytical challenges.\\

Recent advances have focused on developing deep learning approaches for brain age estimation from T1-weighted structural MRI \cite{TANVEER2023130,FENG202015,9740203}.
Deep methods can be further organized by their input representation into whole-volume, ROI/parcellation-based, and slice-based pipelines, as described in \cite{TANVEER2023130}. Our approach belongs to the slice-based branch of this taxonomy, but varies from earlier slice methods by learning a compact ordered inter-slice embedding matrix that approximates volumetric continuity, rather than pooling independent per-slice predictions. Research has demonstrated that traditional methods such as Lasso regression and Support Vector Regression fail to capture complex brain structural relationships, necessitating deep learning models \cite{10.3389/fnagi.2019.00115}. Various architectures have been explored: Convolutional Neural Networks (CNNs) have been employed on both raw and preprocessed MRI data to capture fine-grained structural patterns of the brain \cite{COLE2017115}. Extensions to three-dimensional CNN architectures further enhanced spatial context modeling compared to their 2D counterparts \cite{ueda20193dcnn}, while Fully Convolutional Networks (SFCNs) inspired by VGG Net achieved notable success on large-scale datasets \cite{PENG2021101871}. Beyond CNNs, attention-based architectures including graph transformers utilizing regions of interest \cite{Cai2023GraphTransformer}, global-local transformers \cite{He2022} and 3D-Resnet based architectures \cite{Aghaei2024} have demonstrated decent performance through their attention mechanisms.\\

Beyond age regression, the broader field of deep learning for brain MRI has increasingly prioritized compute-efficient architectures that balance accuracy against parameter count and inference cost. Within brain-tumor classification, compact convolutional and small transformer models have been shown to match far larger backbones while using only a few million parameters \cite{attallah2025lightweight}; directional-split-convolution designs such as DiSCNet \cite{ganie2026discnet} attain strong recognition at a reduced computational budget; systematic comparisons of transformer families clarify the accuracy and efficiency trade-offs across ViT, DeiT and Swin variants \cite{cakmak2025transformer}; and attention-fusion architectures such as T-FSPANNet \cite{pacal2026tfspan} pursue interpretability alongside efficiency. Related efficiency-oriented designs have likewise been applied to automated ischemic stroke lesion segmentation \cite{pacal2025adaptive}. These works address tumour detection, segmentation or stroke analysis rather than age regression; however, they highlight the broader movement toward compute-efficient brain-MRI modeling, which our two-stage ViT-CNN framework extends to the brain-age setting.\\

Nevertheless, despite these developments, significant research gaps persist. Pure transformer–based and global–local transformer models (e.g., He et al.\cite{He2022}) typically fuse global and local information across entire volumes or heavy patch-based pipelines, which increases computational cost, memory footprint, and data requirements, thus hindering scalability across moderate-sized, heterogeneous multi-site MRI cohorts. Graph-transformer and ROI-based methods effectively model inter-regional relationships but often rely on parcellation (i.e.\ predefined ROIs), potentially discarding voxel- or slice-level signals that are crucial for fine-grained brain age prediction \cite{Beheshti2022BrainAge,Cai2023GraphTransformer}. Recent multi-oriented and hybrid designs (e.g., permuted transformer–convolution hybrids \cite{Yang2022} and multi-view ViT fusions, such as Triamese-ViT \cite{Zhang2024}) improve 3D context modeling but do not implement a pipeline that trains on 3D voxel data of whole-brain images in a domain-agnostic approach across a diverse set of independent multi-site cohorts. \\

We first experimented with native-3D architectures, namely 3D-ResNet and 3D-ViT, but found these to be resource-heavy while performing relatively poorly (Tables~\ref{tab:validation-comparison} and~\ref{tab:train_time}), which motivated a departure toward joint-2D or 2.5D formulations. However, the comparatively weak performance of pure 2D-ensemble approaches such as Triamese-ViT \cite{Zhang2024} and the Global-Local Transformer \cite{He2022}, also shown in Table~\ref{tab:validation-comparison}, led us to instead explore 2.5D designs that approximate volumetric continuity without incurring the cost of full 3D modeling. We settled on a two-stage formulation in which slice-level representations are first learned along a given anatomical plane, and a second 2D backbone then operates over the resulting representation space of slices along the orthogonal axis, thus approximating volumetric context while minimizing computational overhead. Empirical justification for the specific backbone choices, the incorporation of sex information, and the manner in which it is incorporated is provided in Section~\ref{sec:ablation}. \\

\begin{figure*} [h]
    \centering
        \includegraphics[width=\linewidth]{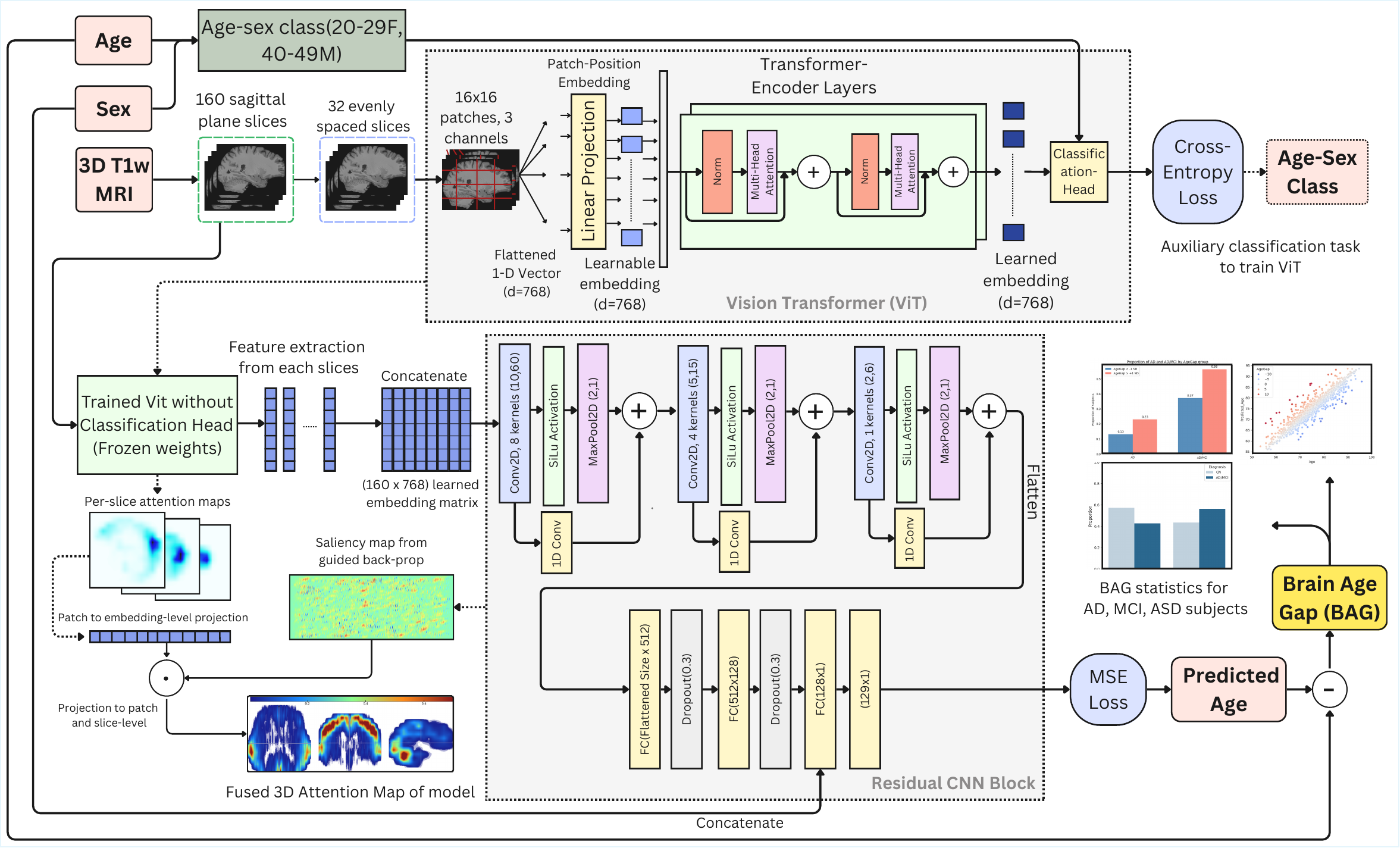}
        \caption{\textbf{Overview of the proposed 2.5D, slice-based representation-learning framework that approximates volumetric context.} The model integrates a Vision Transformer (ViT) encoder to extract slice-level embeddings with a lightweight 2D convolutional regression head to aggregate volumetric information. By applying 2D convolutions over a (slice × embedding) matrix derived from the transformer, the framework jointly models global contextual and local spatial patterns, enabling subsequent analyses on brain age gaps and thus linking aging with neurological conditions such as Alzheimer’s disease, mild cognitive impairment, and autism spectrum disorder.}
        \label{fig:methodology}
\end{figure*}

To the best of our knowledge, no prior transformer-based or hybrid CNN–ViT study on brain age estimation has simultaneously (a) reduced data demands of end-to-end ViT regression while preserving inter-slice continuity in a compact, learnable representation, that approximates a volumetric context (b) been evaluated on as many fully held-out independent cohorts as in our study, (c) enabled efficient regression without reliance on ROI graphs or full 3D attention, (d) provided biologically interpretable explanations consistent with existing literature, and (e) analyzed the association of the brain age-gap (BAG) with neurological disorders. To address these gaps, we introduce a hybrid architecture (Figure~\ref{fig:methodology}) that integrates Vision Transformer (ViT) and Convolutional Neural Network (CNN) components. Our method leverages the global attention capabilities of the transformer, which enhance prediction accuracy by encoding meaningful features, and combines them with the local feature extraction efficiency of a lightweight CNN, which improves computational efficiency compared to an entirely transformer-based setup. Training across diverse multi-site cohorts is intended to improve performance across acquisition sites. This balance of accuracy, efficiency, and clinical applicability forms the foundation of our contributions outlined below.

\begin{itemize}
    \item We propose a two-stage hybrid framework for brain age prediction that decouples representation learning from volumetric aggregation: a Vision Transformer (ViT) encoder is first trained on an auxiliary age–sex composite-classification task and then frozen; its stacked slice-level embeddings are treated as a pseudo-image and passed to a lightweight 2D residual CNN regression head. This design combines global, attention-based slice representations with a computationally cheap mechanism for aggregating inter-slice spatial patterns, achieving competitive accuracy (MAE = 3.43 years; Table~\ref{tab:validation-comparison}) at low training cost (Table~\ref{tab:train_time}), while remaining conceptually distinct from prior CNN--ViT hybrids and 2.5D/multi-view fusion approaches to brain age estimation.

    \item We train and validate our model across 11 independent multi-site brain MRI cohorts using a unified preprocessing pipeline without explicit site harmonization, and evaluate it zero-shot on 4 additional fully held-out cohorts. This provides an empirical assessment of cross-cohort performance, though we note that performance under site-specific variation is assessed empirically rather than guaranteed by design (Tables~\ref{tab:independent-testing},~\ref{tab:independent-testing-comparison}).

    \item We introduce an interpretability pipeline that fuses ViT attention maps with guided backpropagation and maps the resulting saliency onto AAL3 brain regions, identifying regions most associated with aging. These patterns show strong alignment with established neuroanatomical findings and offer additional insight for empirical aging research (Section~\ref{subsec:attention-map}).

    \item We analyze associations between the brain age-gap (BAG) produced by our model and neurological conditions, including Alzheimer's disease (AD), mild cognitive impairment (MCI), and autism spectrum disorder (ASD), demonstrating the clinical relevance and translational potential of the approach (Sections~\ref{subsec:adni_analysis},~\ref{subsec:abide_analysis}). The observed associations between brain age-gap and clinical disorders should, however, be interpreted as cross-sectional and associational rather than causal.
\end{itemize}

\section{Methodology}\label{sec:methodology}
Our proposed framework combines transformer-based and convolutional approaches to achieve accurate and interpretable brain age estimation from 3D MRI scans. It covers the entire process, from data collection and preparation to feature learning, prediction, and interpretation. The workflow starts with gathering multiple MRI datasets and applying a consistent preprocessing pipeline. Then, we use a Vision Transformer (ViT) \cite{dosovitskiy2020image} to learn informative features from MRI slices and employ a residual CNN module to predict brain age. Finally, we generate attention and saliency maps to understand how the model makes its predictions. The following sections describe each of these stages in detail.
% ===================== DATASET ================================ %
\subsection{Dataset}\label{subsec:dataset}

In this study, we utilize fifteen publicly available structural MRI datasets, including prominent collections such as ADNI, IXI, Cam-CAN, and ABIDE, providing a diverse pool of subjects across various age ranges, cognitive statuses, and acquisition protocols. This diversity supports training and evaluation across heterogeneous acquisition conditions. Table~\ref{tab:dataset_info} summarizes the demographic information of the 6,056 total samples used in this study. A brief description of each dataset follows.\\

\begin{table*}[h]
\centering
\footnotesize
\caption{Demographic information of datasets used for training, validation, and independent cohort evaluation.}
\label{tab:dataset_info}
\begin{tabularx}{0.72\textwidth}{lccc}
\hline
\textbf{Dataset} & \textbf{\# Total Samples} & \textbf{\# Samples Used} & \textbf{Age (Mean $\pm$ SD)} \\
\hline
\multicolumn{4}{l}{\textbf{Training + Validation Datasets (Total = 5,108; Train = 4,086, Val = 1,022)}} \\
\hline
ADNI \cite{Mueller2005}      & 5881 & 900 & 63.04 $\pm$ 3.54 \\
IXI \cite{ixi_dataset}      & 565  & 565 & 48.69 $\pm$ 16.46 \\
ABIDE \cite{abideII}    & 1114 & 750 & 17.99 $\pm$ 9.72 \\
DLBS \cite{dlbs, park2025dlbs}     & 315  & 315 & 54.62 $\pm$ 20.09 \\
COBRE \cite{cobre}    & 147  & 146 & 36.98 $\pm$ 12.78 \\
FCON1000 \cite{re3data_1000functionalconnectomes_2025} & 1355 & 763 & 27.86 $\pm$ 13.74 \\
CORR \cite{Zuo2014_CoRR}     & 3357 & 377 & 52.95 $\pm$ 13.22 \\
OASIS-1 \cite{oasis_data}  & 416  & 300 & 64.17 $\pm$ 19.12 \\
CamCAN \cite{taylor2016camcan, shafto2014camcan}   & 655  & 653 & 54.31 $\pm$ 18.58 \\
NIMH \cite{nimh, nugent2022nimh}     & 189  & 181 & 34.62 $\pm$ 11.52 \\
BOLD variability \cite{rieck2024boldvar, rieck2021reconfiguration, rieck2021contributions} & 158 & 158 & 49.28 $\pm$ 19.01 \\
\hline
\multicolumn{4}{l}{\textbf{Independent Cohort Test Datasets (Total = 948)}} \\
\hline
SALD \cite{sald_dataset}        & 494 & 494 & 45.18 $\pm$ 17.44 \\
SUDMEX-CONN \cite{sudmex, angelesvaldez2022sudmex} & 136 & 135 & 30.81 $\pm$ 7.70 \\
AgeRisk \cite{agerisk}     & 189 & 186 & 45.05 $\pm$ 19.27 \\
TrueCrime \cite{truecrime}   & 133 & 133 & 26.98 $\pm$ 8.80 \\
\hline
\textbf{Overall Total} &  & \textbf{6056} & -- \\
\hline
\end{tabularx}
\end{table*}

Descriptions of each dataset, with its source and the subset used, are given in Supplementary Section~\ref{supp:datasets}.

To evaluate generalization, we further conduct independent cohort testing on 948 samples using four datasets entirely excluded from training and validation: SALD, SUDMEX-CONN, AgeRisk, and TrueCrime.\\

For model development, we aggregate eleven datasets into a balanced, unified pool, randomly split into training and validation sets at an 80\%/20\% ratio: 5,108 samples total, with 4,086 for training and 1,022 for validation. In case of longitudinal studies, we make sure that the split is subject-disjoint; no subject contributes scans to more than one partition. Details on subset selection and age-balancing of the training pool are in Supplementary Section~\ref{supp:subset_selection}

\begin{figure*}[t]
    \centering
    % First row
    \begin{subfigure}[b]{0.17\textwidth}
        \centering
        \includegraphics[width=\linewidth]{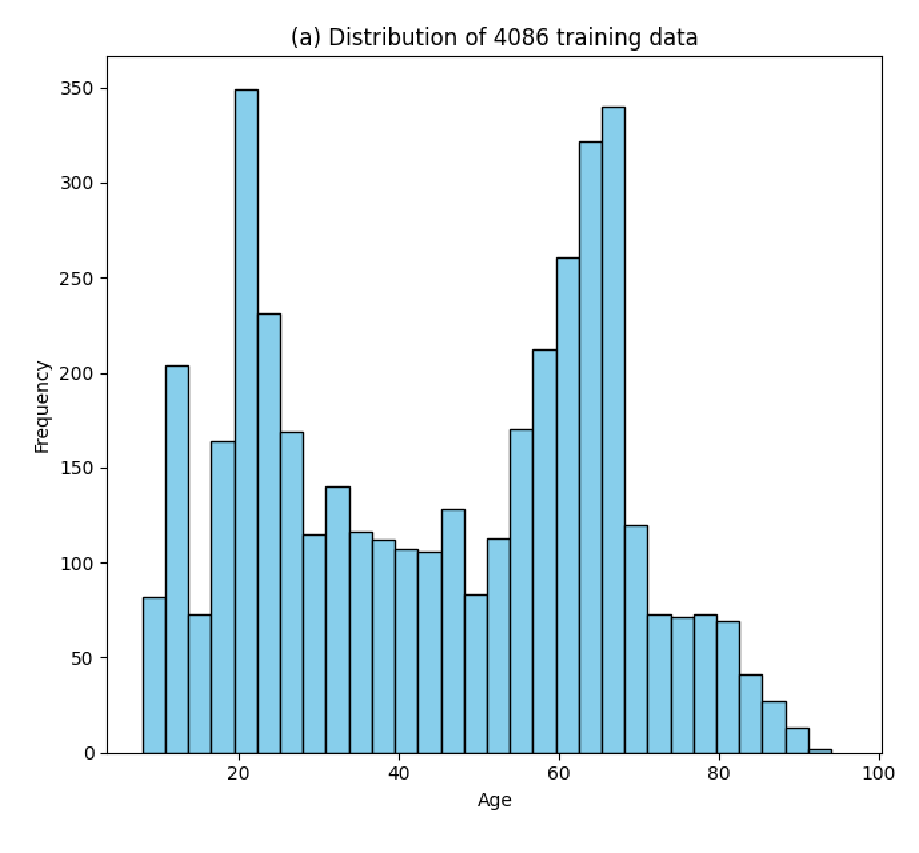}
        \caption{}
        \label{fig:age_train}
    \end{subfigure}%
    % No space here — remove \hfill or any space
    \begin{subfigure}[b]{0.17\textwidth}
        \centering
        \includegraphics[width=\linewidth]{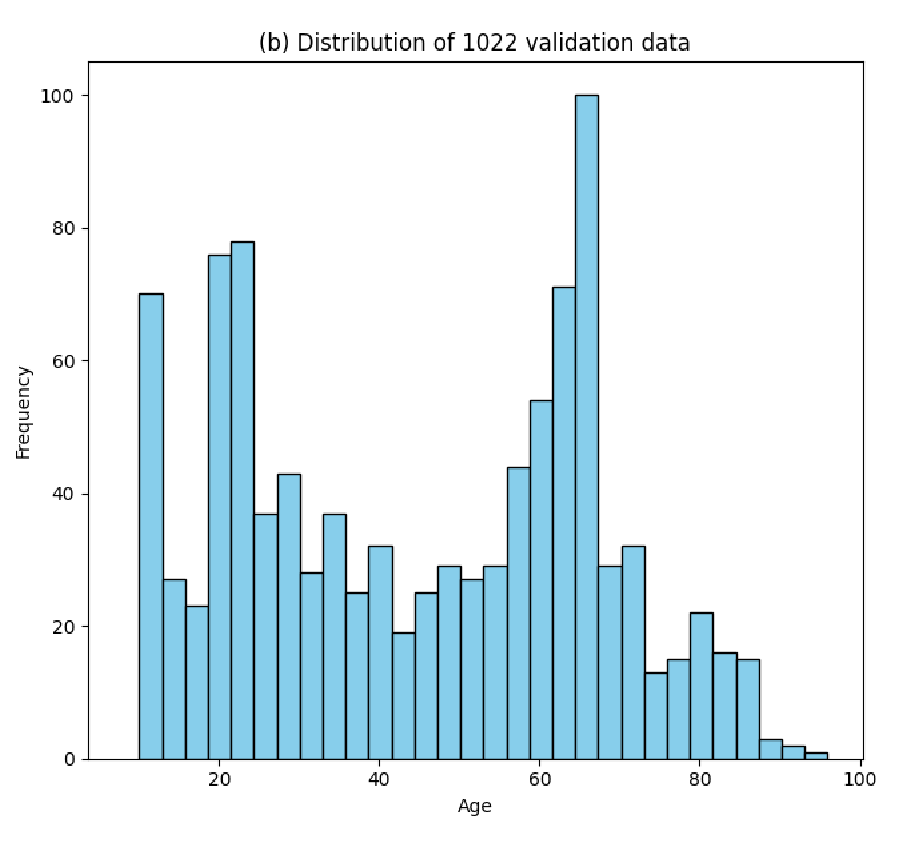}
        \caption{}
        \label{fig:age_val}
    \end{subfigure}
    % Add horizontal gap before the third subfigure
    \hspace{0.02\textwidth}
    \begin{subfigure}[b]{0.62\textwidth}
        \centering
        \includegraphics[width=\linewidth]{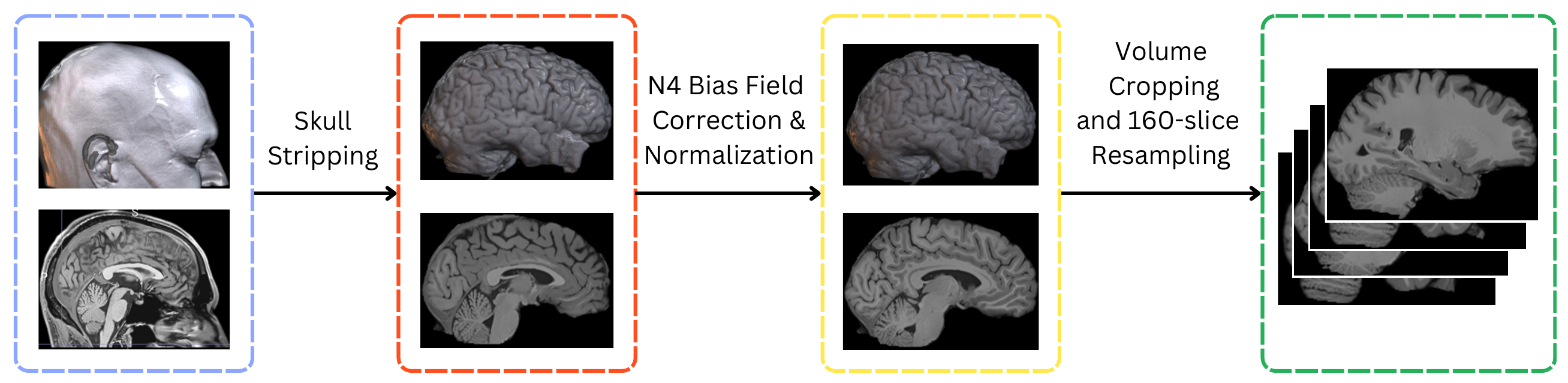}
        \caption{}
        \label{fig:preproc}
    \end{subfigure}

    \vspace{2pt} 
    % Second row
    \begin{subfigure}{0.81\textwidth}
        \includegraphics[width=\linewidth]{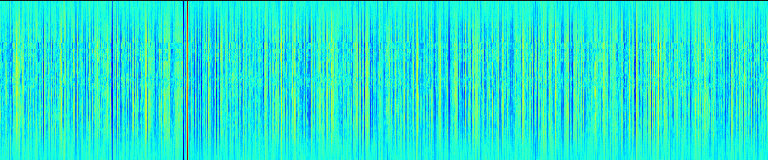}
        \caption{}
         \label{fig:features_heatmap}
    \end{subfigure}\hfill
    \begin{subfigure}{0.165\textwidth}
        \includegraphics[width=\linewidth]{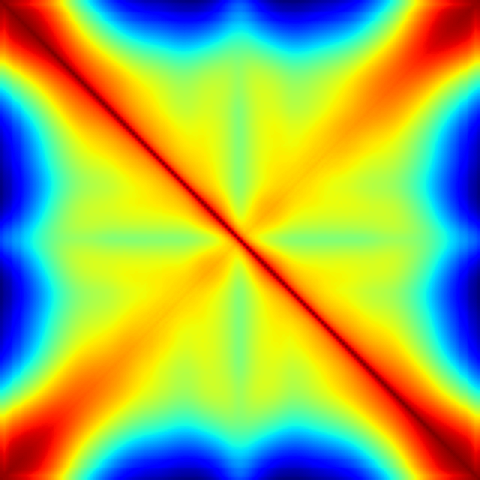}
        \caption{}
        \label{fig:embed_similarity_heatmap}
    \end{subfigure}\hfill
      
    \caption{\textbf{Implementation details of methodology.} \textbf{(a):} Chronological age distribution of the 4086 training samples. \textbf{(b):} Chronological age distribution of the 1022 validation samples. \textbf{(c):} Pre-processing routine converting NIfTI-format 3D T1w MRI images to 2D sagittal slices used as model inputs. \textbf{(d):} Heatmap of a 160 pixel $\times$ 768 pixel 2D feature map (ViT embedding vectors of 160 slices concatenated in order; log-scaled, min-max scaled, and rotated for better visualization) — the input the residual CNN receives for one sample. Slices change along the vertical axis; the horizontal axis is the embedding dimension. Vertical symmetry appears across slices, and adjacent slices show similar embeddings, producing vertical lines of similarly intense positions. \textbf{(e):} Heatmap of a 160$\times$160 cosine similarity matrix between slices of 3D samples, averaged across all samples. The grid axes both index slices 1–160. The principal diagonal represents each slice's similarity with itself (1, most intense red). Intensity spreading from the diagonal shows the ViT produces similar embeddings for neighboring slices, while intensity along the anti-diagonal shows symmetrically antipodal slices are very similar — i.e., the transformer effectively captures sagittal brain symmetry.}
    \label{fig:methodological details}
\end{figure*}

% ===================== Preprocessing ================================ %

\subsection{Data Acquisition and Preprocessing Pipeline}\label{subsec:preprocessing}
To harmonize heterogeneous MRI datasets, we applied a preprocessing pipeline standardizing orientation and image quality. Each scan underwent skull stripping, bias correction, and reorientation to the canonical RAS orientation, followed by cropping, resampling into 160 sagittal slices, and intensity normalization, producing consistent, high-quality inputs for transformer-based feature extraction and CNN regression.

\subsubsection{Input Data Specifications}
We use T1-weighted structural MRI scans acquired with the three-dimensional Magnetization-Prepared Rapid Gradient-Echo (3D MPRAGE) sequence, providing high-resolution anatomical images with excellent gray–white matter contrast essential for accurate brain morphometry \cite{brant1992mp}.

\subsubsection{Skull Stripping and Brain Extraction}

Non-brain tissues (skull, scalp, eyes, etc.) are removed using \textit{DeepBrain}\footnote{\href{https://github.com/iitzco/deepbrain}{https://github.com/iitzco/deepbrain}}, a lightweight, U-Net-based \cite{ronneberger2015unetconvolutionalnetworksbiomedical} skull stripping tool trained on diverse multi-site data. Unlike traditional intensity- or atlas-based methods, it uses convolutional neural networks to accurately delineate brain boundaries across varied acquisition protocols and field strengths.

\subsubsection{Intensity Inhomogeneity Correction}

To correct spatial intensity variations from RF field and coil sensitivity artifacts, we apply the N4 bias field correction algorithm \cite{tustison2010n4itk}, which iteratively estimates a smooth multiplicative bias field via B-spline approximations, yielding spatially consistent tissue intensities while preserving true anatomical contrast.

\subsubsection{Volume Resampling and Slice Generation}

Bias-corrected volumes, reoriented to RAS, are tightly cropped to intracranial boundaries to remove background and reduce computation, then resampled into 160 sagittal slices via cubic-spline interpolation. Sagittal orientation leverages bilateral brain symmetry and captures key structures (corpus callosum, hippocampus, cortex) relevant to age-related changes, facilitating model interpretability.

\subsubsection{Intensity Normalization}

Each slice undergoes min--max normalization to $[0,1]$, followed by standardization with the ViT feature-extractor mean and standard deviation, to reduce global intensity differences across scans and sites, stabilizing input distributions and mitigating site-specific intensity biases to improve model robustness.\\

Further details on preprocessing are provided in Supplementary Section 5.

% ===================== ViT Feature Learning ================================ %

\subsection{Stage 1: Vision Transformer-Based Feature Learning}

This stage uses a Vision Transformer (ViT) to learn discriminative features from individual 2D sagittal brain slices. Training the ViT on an auxiliary task of classifying combined age-decade and sex categories encourages the model to extract representations sensitive to anatomical variations related to aging and sex. The output is a 2D feature map in which each row corresponds to a slice and each column to a learned embedding dimension, i.e., a compact representation of the entire 3D brain volume.

\subsubsection{Classification Task Formulation for Representation Learning}

To train the ViT for discriminative feature extraction, we formulate an auxiliary classification task encouraging age- and sex-sensitive representations. Rather than directly regressing continuous age values, which could cause overfitting, we discretize age into decade-based bins (20-29, 30-39, 40-49, etc.) and further subdivide each by biological sex, creating composite classes such as ``50-59 years, male'' or ``30-39 years, female'' — approximately 16-20 classes depending on the dataset's age distribution.

This formulation serves multiple purposes: it transforms continuous age prediction into classification, which typically yields more robust representations during initial training; it explicitly encourages learning sex-specific aging patterns, given well-documented differences in brain aging trajectories between males and females; and decade-based grouping provides sufficient granularity to capture major developmental and aging milestones while maintaining adequate sample sizes per class for stable training.

\subsubsection{Vision Transformer Architecture and Training Protocol}

The ViT processes input slices through an attention-based architecture. Each slice is divided into non-overlapping $16 \times 16$ pixel patches, linearly projected into a 768-dimensional embedding space, forming a sequence of patch embeddings. Learnable positional encodings are added to preserve spatial information about original patch locations.

The embedding sequence passes through multiple transformer encoder layers, each with multi-head self-attention (MSA) and position-wise feed-forward network (FFN) sub-layers:
\begin{align}
\mathbf{z}^{(l+1)} &= \mathbf{z}^{(l)} + \text{MSA}(\text{LayerNorm}(\mathbf{z}^{(l)})) \\
\mathbf{z}^{(l+1)} &= \mathbf{z}^{(l+1)} + \text{FFN}(\text{LayerNorm}(\mathbf{z}^{(l+1)}))
\end{align}
where $\mathbf{z}^{(l)}$ is the layer-$\ell$ representation and LayerNorm denotes layer normalization for training stability.

During training, 32 evenly spaced sagittal slices are sampled from each subject's 160 slices and processed independently through the ViT encoder:
\begin{equation}
f_{\text{ViT}}: S'_i \rightarrow \mathbb{R}^{768}
\end{equation}

The encoder generates a 768-dimensional feature vector per slice, capturing local anatomical patterns and global contextual information via self-attention.

The classification head is a single linear layer with softmax, producing class probabilities over the age-sex classes, optimized via categorical cross-entropy loss:
\begin{equation}
L_{\text{CE}} = -\sum_{c=1}^C y_c \log(\hat{y}_c)
\end{equation}
where $y_c$ is the one-hot ground truth and $\hat{y}_c$ the predicted probability for class $c$.\\

To improve training efficiency, we employ Weight-Decomposed Low-Rank Adaptation (DoRA) \cite{liu2024dora} for parameter-efficient fine-tuning of the pretrained ViT. Rather than updating every backbone parameter, DoRA decomposes each pretrained weight into a learnable magnitude and normalized directional component. For a pretrained weight matrix $\mathbf{W}_0$, the adapted weight is:

\begin{equation}
\mathbf{W}
=
m
\frac{\mathbf{W}_0 + \Delta\mathbf{W}}
{\left\|\mathbf{W}_0 + \Delta\mathbf{W}\right\|_F},
\qquad
\Delta\mathbf{W}
=
\mathbf{B}\mathbf{A},
\end{equation}

where $m$ is a learnable scalar controlling weight magnitude, $\Delta\mathbf{W}$ is a low-rank update parameterized by $\mathbf{A}\in\mathbb{R}^{r\times d}$ and $\mathbf{B}\in\mathbb{R}^{k\times r}$ with rank $r \ll \min(k,d)$, and $\|\cdot\|_F$ is the Frobenius norm. Learning only $\mathbf{A}$, $\mathbf{B}$, and $m$ greatly reduces trainable parameters while preserving the pretrained backbone's representational capacity, substantially reducing GPU memory and training time versus full fine-tuning. Only the DoRA parameters and task-specific classification head were updated; the remaining ViT backbone stayed frozen.

\subsubsection{Dense Feature Map Generation}

After ViT training on the classification task, the model becomes a feature extractor for the downstream regression task: the classification head is discarded and the encoder frozen. This frozen encoder is applied to all 160 sagittal slices per brain volume, generating a comprehensive feature representation.

Processing all slices, rather than just the sampled training subset, ensures that the loss of anatomical information introduced by slice subsampling is minimized. Each slice yields a 768-dimensional embedding, arranged sequentially by spatial position along the sagittal axis into a 2D feature map:
\begin{equation}
\mathbf{Z} = [\mathbf{z}_1; \mathbf{z}_2; \ldots; \mathbf{z}_{160}]^T \in \mathbb{R}^{160 \times 768}
\end{equation}

This feature map is a compact, learned representation of the 3D brain volume, where each row is a sagittal slice and each column a learned feature dimension, preserving anatomical continuity and enabling the CNN to exploit spatial relationships between adjacent slices.

% ===================== CNN-RESNET ================================ %

\subsection{Stage 2: Residual Convolutional Neural Network for Brain Age Regression}

In the second stage, the $160 \times 768$ feature map from the frozen ViT encoder is repurposed as a ``pseudo-image,'' processed by a custom residual Convolutional Neural Network (CNN) designed to learn spatial relationships between slice-level features along the sagittal axis. This stage integrates these learned spatial patterns with biological sex information to perform the final regression, predicting a scalar brain age.

\subsubsection{Feature Map as Pseudo-Image Representation}

The ViT-stage feature map is treated as a single-channel pseudo-image of dimensions $160 \times 768$, with rows ordered by position along the sagittal axis and columns corresponding to embedding dimensions, so that standard 2D convolutional operations can be applied. Along the slice axis, the convolutional filters can capture gradual anatomical transitions between adjacent slices. Along the embedding axis, however, the dimensions have no inherent spatial or semantic ordering (Section~\ref{sec:discussion}), so the 2D kernels there act as learned local combinations of embedding features rather than as detectors of spatial structure. The use of 2D rather than 1D convolution is therefore an empirical design choice: it gave a small but consistent improvement over a 1D CNN that convolves only along the slice axis (Section~\ref{sec:ablation}), which we do not interpret as evidence of intrinsic spatial structure among the embedding dimensions.

\subsubsection{Residual Convolutional Block Architecture}
The CNN comprises three hierarchical residual convolutional blocks progressively extracting and refining features while maintaining gradient flow via ResNet-inspired skip connections. Each block uses a dual-pathway design: the main path applies a convolution, \(\text{SiLU}(x)=x\cdot\text{sigmoid}(x)\) activation, and max pooling (kernel size 2, stride 1) for local translation invariance, while the residual path aligns dimensions using \(1\times1\) convolutions (for channel mismatches) and nearest-neighbor interpolation (for spatial mismatches). Both paths merge via element-wise addition to preserve information and support gradient propagation.\\

ConvBlock 1 uses 8 filters of size \((10,60)\) to capture broad spatial patterns; ConvBlock 2 refines these into 4 channels with \((5,15)\) kernels for localized features; ConvBlock 3 compresses into 1 channel with \((2,6)\) kernels, producing a compact age-relevant representation.

\subsubsection{Fully Connected Layers and Feature Integration}
After convolutional extraction, the output tensor is flattened into a 1D vector. Since its size depends on post-pooling spatial dimensions, the first fully connected layer (FC1) is initialized dynamically via a dummy forward pass. The flattened features pass through two FC layers: FC1 maps to 512 dimensions, followed by layer normalization (more stable than batch normalization for small batches \cite{ba2016layer}), \(\text{SiLU}\) activation, and dropout (\(p=0.3\)); FC2 reduces to 128 dimensions with the same normalization, activation, and dropout. This progressive reduction enforces increasingly abstract age-related representations. Biological sex (binary: 0 male, 1 female) is then concatenated to the 128-dimensional vector, forming 129 dimensions. This late fusion lets earlier layers learn sex-independent features while the final layer incorporates sex-specific aging differences to improve accuracy.

\subsubsection{Final Regression Layer}

The final layer is a linear transformation mapping the 129-dimensional combined vector to a scalar predicted brain age:
\begin{equation}
\hat{y} = \mathbf{W}_3 \mathbf{h}_{\text{combined}} + b_3
\end{equation}
where $\hat{y}$ is predicted chronological age, $\mathbf{W}_3$ a $1 \times 129$ weight matrix, $\mathbf{h}_{\text{combined}}$ the concatenated feature vector including sex, and $b_3$ a scalar bias.

\subsubsection{Loss Function}

The model is primarily trained using Mean Squared Error (MSE) loss ($L_{\text{MSE}}$), penalizing predictions by squared difference from actual ages, adopted for its use in prior imaging-based age regression work \cite{ueda20193dcnn} \cite{He2022}
\cite{Zhang2024}.
\begin{equation}
L_{\text{MSE}} = \frac{1}{N} \sum_{i=1}^N (a_i - \hat{y}_i)^2
\end{equation}
where $a_i$ is ground-truth age for subject $i$, $\hat{y}_i$ the prediction, and $N$ the batch size.\\

\subsubsection{Optimization and Regularization}

The gradient of the MSE loss guides optimization via the Adam optimizer, with learning rates of $5 \times 10^{-4}$ for the CNN and $1 \times 10^{-4}$ for the ViT backbone, balancing convergence speed and stability; Adam's adaptive mechanisms help navigate the deep architecture's complex loss landscape.

Regularization uses multiple mechanisms: \textbf{Dropout} (p=0.3) after each fully connected layer randomly deactivates neurons to encourage robust feature learning; \textbf{layer normalization} stabilizes training and acts as an implicit regularizer; the \textbf{residual connections} in convolutional blocks provide alternative gradient paths and prevent overly complex learned transformations. Algorithm \ref{alg:brainrot} of Supplementary Section~\ref{supp:algos} shows the full training pipeline of BrainRotVit.

\subsection{Model Interpretability via Guided Backpropagation and ViT Patch Mapping}

To interpret age-relevant features learned by our two-stage model, we implement a 3D interpretability pipeline combining guided backpropagation of the CNN regression model with patch-wise relevance mapping of the pre-trained ViT. This generates slice-level saliency maps, projected onto ViT patch embeddings and aggregated across subjects into a volumetric attention representation of the brain, as shown in Algorithm \ref{alg:interpretability} of Supplementary Section~\ref{supp:algos}.

\subsubsection{Guided Backpropagation of CNN Features}

We generate guided gradients with respect to the CNN input feature map. The CNN, trained on ViT-derived embeddings, is set to evaluation mode. Guided backpropagation registers hooks on all ReLU and SiLU activations: forward hooks record activations, backward hooks allow only positive gradients through neurons active during the forward pass. Given an input feature map $(1,160,768)$ and the subject's binary sex encoding, the module produces a normalized gradient map per slice, highlighting input dimensions that positively influence predicted brain age.

\subsubsection{Projection onto ViT Patch Tokens}

To relate CNN saliency to anatomical locations in the original MRI slices, we map guided gradients onto ViT patch embeddings. Each preprocessed slice passes through the ViT backbone to obtain patch embeddings; the guided CNN vector is projected onto these via dot product, yielding patch-wise importance scores. CLS-token attention weights to each patch in the final transformer layer are also extracted. Both patch scores and CLS attention are normalized to $[0,1]$.

\subsubsection{Slice-Level Heatmap Generation}

Patch scores and CLS attention maps are multiplicatively fused into a slice-level heatmap, upsampled to original slice resolution via bilinear interpolation and normalized, producing a per-slice saliency image reflecting combined CNN and ViT importance.

\subsubsection{Sanity Checks for Saliency Maps}
To verify saliency maps reflect learned model parameters rather than input or architectural artifacts, we conduct model-parameter randomization sanity checks following Adebayo et al.~\cite{adebayo2018sanity}. Using a held-out set of $N=100$ validation subjects, we progressively randomize the trained ViT's weights, cascading from the top layer to full randomization, comparing resulting unfused, unmasked central-slice attention maps against the fully trained model's via per-subject Spearman rank correlation, averaged across subjects. The original model trivially yields $\rho = 1.00$. Randomizing only the top ViT layer causes immediate collapse to $\rho = 0.21$; full backbone randomization drives correlation to near zero or negative ($\rho = -0.25 \pm 0.14$ for raw attention, $\rho = -0.38$ for fused saliency), indicating the maps depend strongly on learned weights rather than architecture or input structure alone. Randomizing only the CNN regression head introduces a relatively small fluctuation to the fused map ($\rho = 0.997$), indicating the fused saliency signal is dominated by ViT attention, with the CNN projection contributing a smaller independent signal.\\

We also assess stability and reproducibility of attention maps across subjects. Per-subject unfused central-slice attention maps show substantial inter-subject variability at the voxel level (median coefficient of variation $\approx 1.29$), consistent with individual anatomical differences. Despite this, split-half reliability analysis — randomly partitioning the validation cohort into two independent groups of 50 — shows the mean attention pattern is highly reproducible across halves (Spearman $\rho = 0.942$, Pearson $r = 0.943$), indicating that while individual attention maps vary, the group-level aging-associated pattern is stable and reproducible. Full randomization cascade curves, example maps at increasing randomization, and inter-subject/split-half stability figures are in Supplementary Section 2.3.\\

\subsubsection{Aggregation into 3D Attention Volume}

To construct a volumetric representation, slice-level heatmaps are averaged across validation subjects for each sagittal slice. The resulting per-slice maps are stacked sequentially along the sagittal axis into a 3D attention volume $(S, H, W)$, where $S$ is slice count and $H$, $W$ are slice height and width. The volume is normalized, preserving anatomical continuity while highlighting regions consistently associated with brain aging.\\

\subsubsection{Identification of Regions of Interest}

To identify regions of interest (ROI) receiving the most attention, the 3D attention volume is cropped, centered, and aligned via RAS affine registration to fit a bounding box matching the Automated Anatomical Labelling atlas 3 (AAL3), achieving complete overlap between volume and atlas. We validate registration by visually inspecting alignment of major anatomical landmarks between attention volume and atlas/template (figures in Supplementary Section 2.1). We iterate over all AAL regions, calculating average voxel intensity in the attention map, then group regions and their voxel-count-weighted intensities into larger regions (e.g., grouping cerebellum sub-regions into a single ROI); the complete list and groupings are in Supplementary Section 2.2.
Note that the slice extraction, resampling, and affine alignment pipeline used to register the attention volume to AAL3 is inherently approximate; ROI intensity scores should thus be interpreted as coarse regional indicators of attention rather than voxel-precise localisation, and claims about specific substructures or boundaries within an ROI should be considered qualitative rather than exact.

% ===================== EXPERIMENTAL SETUP ================================ %

\subsection{Code, Environment, and Availability}\label{subsec:environment}

Unless otherwise stated, all experiments were carried out on two Debian 12 workstations with NVIDIA GeForce RTX~3060 (12\,GB VRAM) GPUs and AMD Ryzen~7 processors (3700X and 5700G), with 16 GB system memory. The end-to-end training experiment discussed in Section~\ref{sec:discussion} exceeded the memory of these workstations and was therefore run separately on a server with an NVIDIA H100 NVL (94\,GB VRAM) GPU, an AMD EPYC 9124 processor, and 756\,GB system memory. Implementation used PyTorch \cite{paszke2019pytorchimperativestylehighperformance} with CUDA acceleration; the ViT component was adapted from the official Hugging~Face Transformers \cite{wolf2020huggingfacestransformersstateoftheartnatural} implementation. Brain MRI preprocessing (loading, cropping, resampling, normalization) used NiBabel \cite{brett2024nibabel} and SciPy \cite{virtanen2020scipy}. The ViT encoder trained for 5 epochs on the auxiliary age-sex classification task; the residual CNN regression head trained for up to 200 epochs, using the same optimization settings (Adam, MSE loss), with early stopping on validation MAE to prevent overfitting. The pipeline remained stable throughout, with the ViT stage converging within 12 hours and the CNN stage completing in about 4 hours on the described hardware.

The code is available through a GitHub repository at \href{https://github.com/wjalal/BrainRotViT/}{https://github.com/wjalal/BrainRotViT/}.

\section{Results}
\label{sec:results}
\begin{figure*}
    \centering
    \begin{subfigure}[b]{0.48\textwidth}
        \centering
        \includegraphics[width=\linewidth]{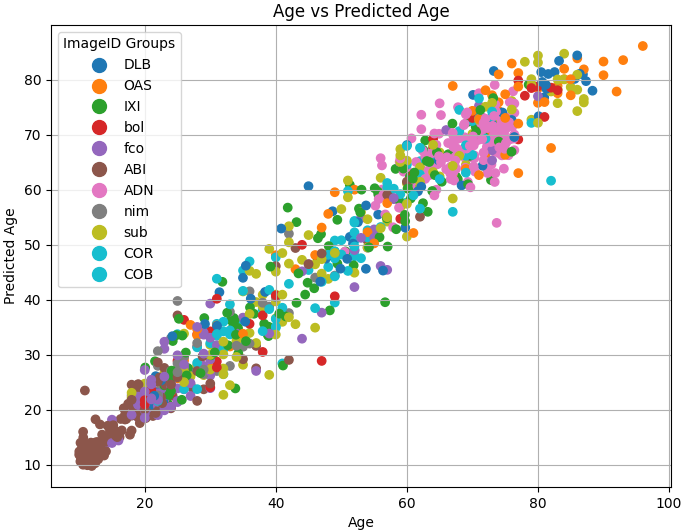}
        \caption{}
        \label{fig:full_val}
    \end{subfigure}
    \begin{subfigure}[b]{0.475\textwidth}
        \centering
        \includegraphics[width=\linewidth]{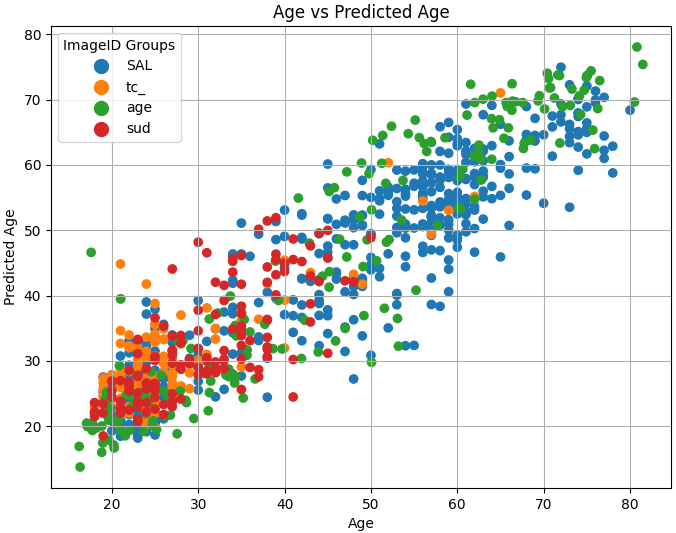}
        \caption{}
        \label{fig:full_test}
    \end{subfigure}
    \vspace{2pt} 

    \begin{subfigure}[b]{0.49\textwidth}
        \centering
        \includegraphics[width=\linewidth]{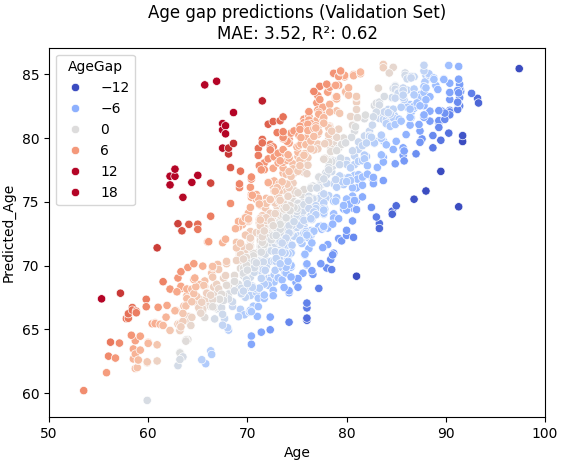}
        \caption{}
        \label{fig:adni_val}
    \end{subfigure}
    \begin{subfigure}[b]{0.49\textwidth}
        \centering
        \includegraphics[width=\linewidth]{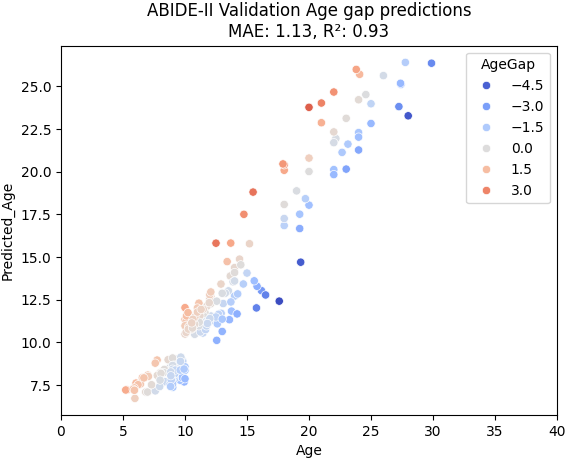}
        \caption{}
        \label{fig:abide_val}
    \end{subfigure}

    \vspace{2pt} 
    % Second row: as before
    \begin{subfigure}[b]{0.35\textwidth}
        \centering
        \includegraphics[width=\linewidth]{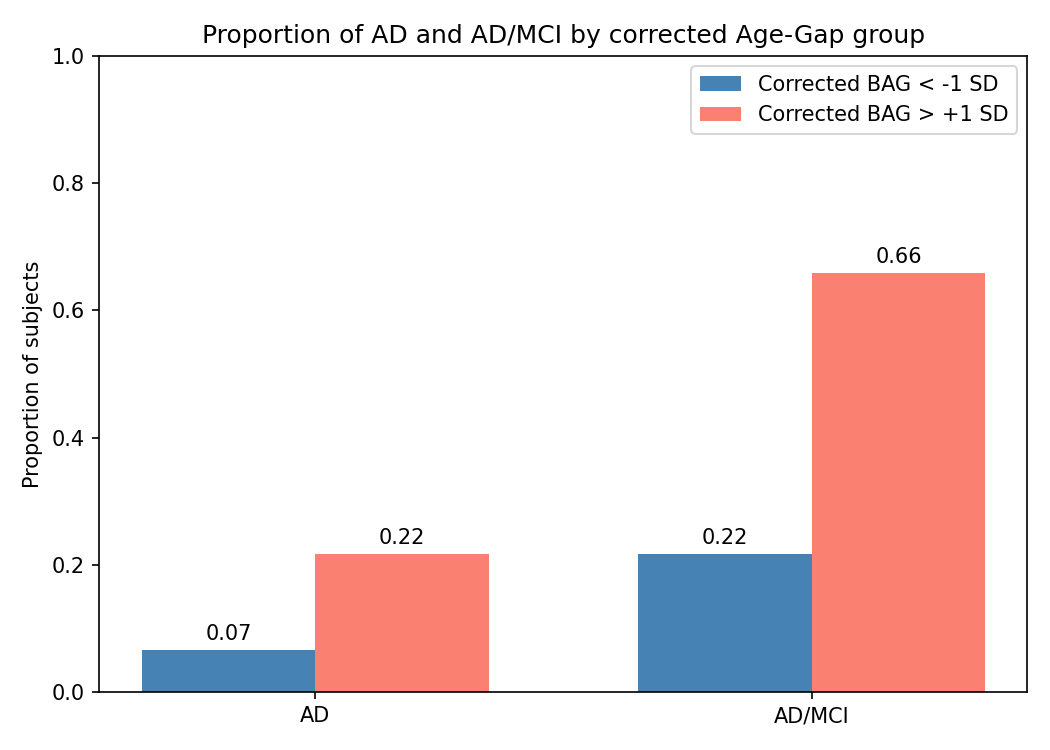}
        \caption{}
        \label{fig:ad_mci_pos_neg}
    \end{subfigure}%
        \begin{subfigure}[b]{0.3125\textwidth}
        \centering
        \includegraphics[width=\linewidth]{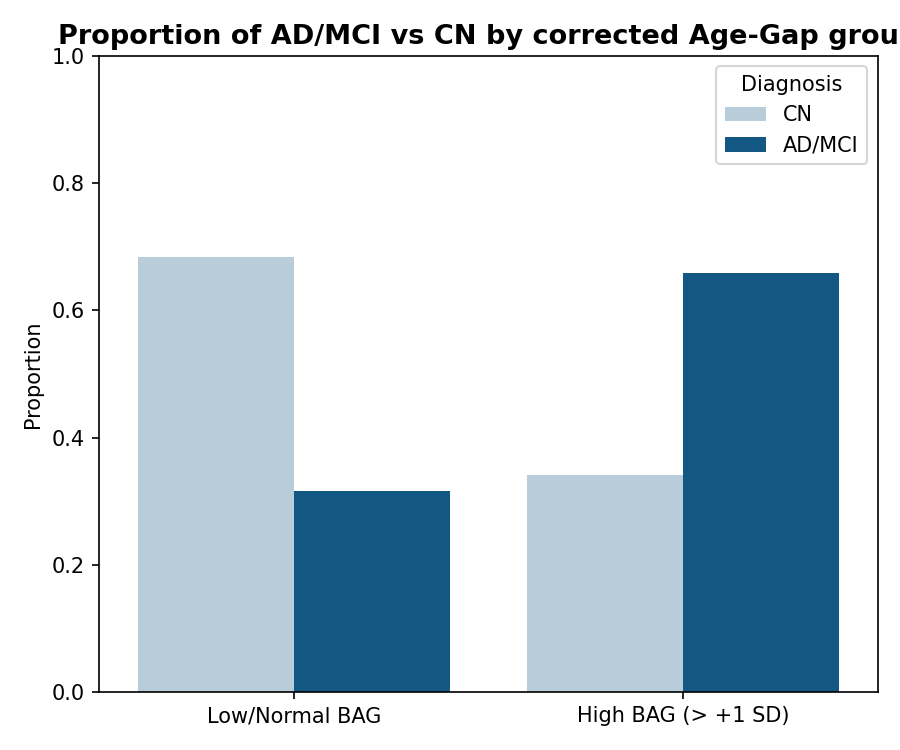}
        \caption{}
        \label{fig:ad_mci_high}
    \end{subfigure}
    \begin{subfigure}[b]{0.3125\textwidth}
        \centering
        \includegraphics[width=\linewidth]{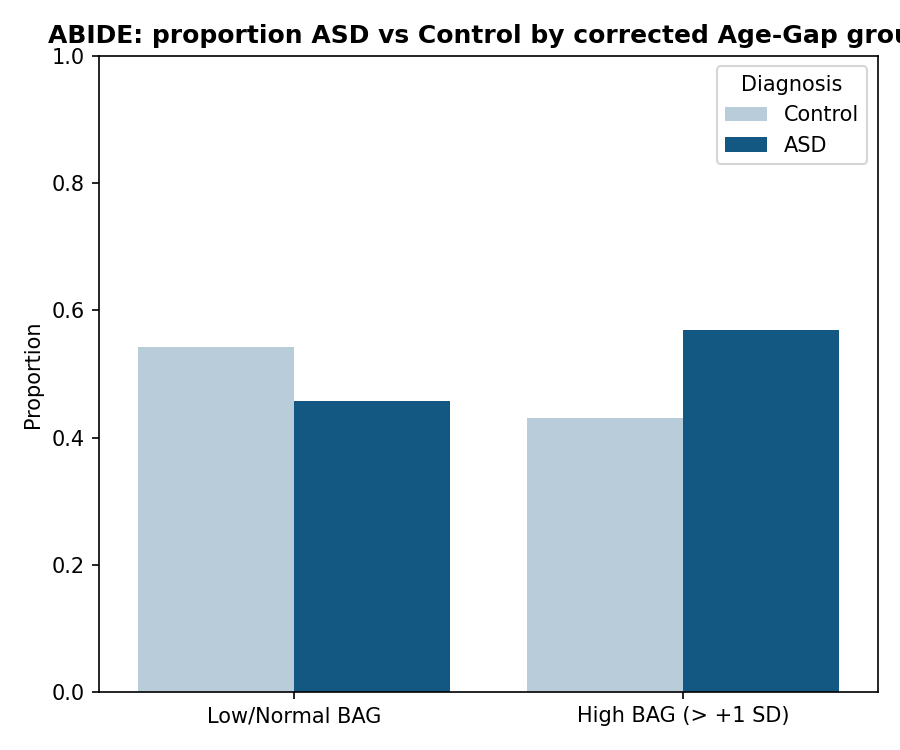}
        \caption{}
        \label{fig:asd_high}
    \end{subfigure}%

    \caption{\textbf{(a):} Scatter plot of predicted age of validation samples from 11 datasets (MAE = 3.43 years, Pearson $r = 0.98$, Spearman $\rho = 0.97$, $R^2 = 0.96$). \textbf{(b):} Scatter plot of predicted age of independent test samples from 4 datasets (MAE = 5.15 years, Pearson $r = 0.925$, Spearman $\rho = 0.901$, $R^2 = 0.854$). \textbf{(c):} Scatter plot of validation (1,119 scans from 276 subjects) on ADNI only (5,881 scans from 1,378 total subjects). \textbf{(d):} Scatter plot of validation (223 out of 1,113 subjects) on ABIDE-II only. \textbf{(e):} Bias-corrected proportion of AD and AD/MCI subjects with extreme positive aging ($\textit{BAG}>1 \textit{std.dev.}$) and extreme negative aging ($\textit{BAG}<-1 \textit{std.dev.}$) visualized. \textbf{(f)} Bias-corrected proportion of AD/MCI subjects with extreme positive aging ($\textit{BAG}>1 \textit{std.dev.}$) visualized against other AD/MCI subjects in ADNI. \textbf{(g)} Bias-corrected proportion of ASD subjects with extreme positive aging ($\textit{BAG}>1 \textit{std.dev.}$) visualized against other ASD subjects in ABIDE-II.  }
    \label{fig:methodological_details}
\end{figure*}

\subsection{Validation Performance and Comparison with State-of-the-Art}
\label{subsec:validation-performance}
The proposed hybrid Vision Transformer--CNN framework achieved strong predictive performance on the validation set, reaching a mean absolute error (MAE) of \textbf{3.43 years} and a Pearson correlation coefficient ($r$) of \textbf{0.98} with chronological age. The coefficient of determination ($R^2$) of \textbf{0.96} further indicates high predictive fidelity and stability across subjects. All architectural and hyperparameter selection was performed on the validation split, so the 3.43-year figure is a development-set estimate and is optimistic; the four fully held-out cohorts (Section~\ref{subsec:independent-cohort}) provide the unbiased estimate of generalization.

Table~\ref{tab:validation-comparison} compares BrainRotViT with representative baseline and state-of-the-art methods using a common upstream preprocessing pipeline, with model-specific input adaptation where required. The comparison methods were chosen to represent the main types of brain age estimation models while ensuring that all of them could be reproduced fairly in our study. We included a CNN-based baseline (3D-ResNet), a Transformer-based baseline (3D-ViT), and several strong hybrid and multi-stage models with publicly available implementations: SFCN~\cite{PENG2021101871}, the Two-Stage Age Network (TSAN)~\cite{cheng2021brain}, the Global-Local Transformer~\cite{He2022}, and Triamese-ViT~\cite{Zhang2024}. We only selected models that could be retrained from their original source code using the same dataset and common upstream preprocessing pipeline, with model-specific input resizing and intensity handling where required by the original implementation. This ensures a fair comparison, as results reported on different datasets or with different preprocessing procedures are often not directly comparable. See Section~\ref{train_and_computational_efficiency} for implementation details.

Baseline models show substantially higher errors, with MAE values of 5.55 years for 3D-ResNet and 6.37 years for 3D-ViT. Among the advanced approaches, the full TSAN model achieves the lowest MAE among competing methods at 3.72 years. BrainRotViT attains a 0.29-year lower MAE than TSAN (3.43 vs.\ 3.72 years; approximately 7.8\%), together with the highest Pearson correlation ($r=0.98$) and coefficient of determination ($R^2=0.96$), whereas TSAN attains a slightly higher Spearman correlation ($\rho=0.98$ vs.\ 0.97). Because this margin is modest and was measured on the validation set used for model selection (five-seed mean MAE of BrainRotViT: $3.50 \pm 0.05$ years; Supplementary Section~\ref{app:ablations}), we regard the validation accuracy of the two methods as broadly comparable; the differences are more consistent on the external cohorts (Section~\ref{subsec:independent-cohort}) and in training cost (Section~\ref{train_and_computational_efficiency}).

\begin{table}[h]
    \centering
    \caption{Comparison of the proposed BrainRotViT model with existing methods on validation data. See Supplementary Section~\ref{app:stratified} for further details.}
    \label{tab:validation-comparison}
    \resizebox{\columnwidth}{!}{%
    \begin{tabular}{lcccc}
        \toprule
        \textbf{Method} & \textbf{MAE} & \textbf{Pearson $r$} & \textbf{Spearman $\rho$} & \textbf{$R^2$} \\
        \midrule
        3D-ResNet (baseline) & 5.55 & 0.96 & 0.92 & 0.88 \\
        3D-ViT (baseline) & 6.37 & 0.81 & 0.87 & 0.78 \\
        SFCN \cite{PENG2021101871} & 5.03 & 0.96 & 0.95 & 0.91 \\
        Two-Stage Age \\ Network (TSAN) \cite{cheng2021brain} & 3.72 & 0.92 & \textbf{0.98} & 0.92 \\
        TSAN 1\textsuperscript{st}-stage & 4.23 & 0.92 & 0.97 & 0.91 \\
        Global-Local \\ Transformer \cite{He2022} & 5.10 & 0.96 & 0.92 & 0.90 \\
        Triamese-ViT \cite{Zhang2024} & 5.95 & 0.89 & 0.90 & 0.87 \\
        BrainRotViT (ours) & \textbf{3.43} & \textbf{0.98} & 0.97 & \textbf{0.96} \\
        \bottomrule
    \end{tabular}%
    }
\end{table}

% \begin{figure}[h]
% \centering
% \includegraphics[width=\linewidth]{figures/age_vs_predicted.png}
% \caption{Scatter plot of chronological age versus predicted brain age for the validation cohort.}
% \label{fig:age-vs-predicted}
% \end{figure}

% Compared to traditional regression-based approaches such as XGBoost and Deep Neural Networks (DNN), the proposed model reduces prediction error by over 50\%. When benchmarked against deep-learning baselines like 3D CNN and CNN--MLP, BrainRotViT shows a consistent 20--30\% improvement in MAE. The high correlation ($r = 0.98$) and $R^2 = 0.93$ indicate that the model captures both global and local age-related patterns effectively. Figure~\ref{fig:age-vs-predicted} shows a clear linear relationship between predicted and chronological ages, with most data points closely following the identity line. These results confirm that adding transformer-based long-range feature extraction helps improve the model’s accuracy and stability in brain age prediction.

\subsection{Independent Cohort Testing}
\label{subsec:independent-cohort}

To assess the generalization capability of the proposed BrainRotViT model, we conducted independent testing on four external cohorts that were excluded from model training. As summarized in Table~\ref{tab:independent-testing}, BrainRotViT achieved consistently low MAE across the evaluated external datasets, with MAE values between 4.79 and 5.35 years. Across all external cohorts combined ($n=948$), the model achieved an overall MAE of 5.15 years, with generally favorable agreement between predicted and chronological age ($R^2=0.854$, Pearson $r=0.925$, Spearman $\rho=0.901$) and some variability across cohorts. Cohort-specific Pearson correlations ranged from 0.740 to 0.940 and Spearman correlations from 0.512 to 0.936, with the weakest agreement in SUDMEX ($R^2=0.35$) and TrueCrime ($R^2=0.50$, $\rho=0.512$), the two cohorts with the narrowest age ranges (Section~\ref{sec:discussion}). All correlation coefficients were statistically significant (each cohort $p<10^{-9}$), indicating a positive association between predicted and chronological age in every independent dataset, although its strength was cohort-dependent.

\begin{table}[!htbp]
\centering
\footnotesize
\setlength{\tabcolsep}{5pt}
\renewcommand{\arraystretch}{1.1}
\caption{Independent cohort evaluation of BrainRotViT. Each external dataset was excluded from training and used only for testing.}
\label{tab:independent-testing}
\begin{tabular}{lccccc}
\toprule
\textbf{Cohort} & \textbf{$n$} & \textbf{MAE} & \textbf{$R^2$} & \textbf{Pearson $r$} & \textbf{Spearman $\rho$} \\
\midrule
SALD & 494 & 5.35 & 0.850 & 0.925 & 0.906 \\
AgeRisk & 186 & 5.12 & 0.880 & 0.940 & 0.936 \\
SUDMEX & 135 & 4.87 & 0.350 & 0.740 & 0.760 \\
TrueCrime & 133 & 4.79 & 0.500 & 0.810 & 0.512 \\
\hline
Overall & 948 & 5.15 & 0.854 & 0.925 & 0.901 \\
\bottomrule
\end{tabular}
\end{table}

To compare the generalizability of our method against existing approaches, we evaluated BrainRotViT alongside representative CNN- and Transformer-based models which had the best validation performance, on four independent external cohorts. As shown in Table~\ref{tab:independent-testing-comparison}, BrainRotViT consistently achieved the lowest MAE across all four cohorts. Compared with TSAN, which was the strongest competitor on the internal validation set, BrainRotViT reduced the MAE on every external dataset, indicating lower error under cross-dataset shift. Similar improvements were observed relative to SFCN and 3D-ResNet, with BrainRotViT also achieving the lowest overall MAE across all external cohorts.

\begin{table}[!htbp]
\centering
\footnotesize
\setlength{\tabcolsep}{6pt}
\renewcommand{\arraystretch}{1.15}
\caption{Independent cohort performance comparison (MAE, years) between BrainRotViT and representative baseline and state-of-the-art methods. Lower values indicate better performance.}
\label{tab:independent-testing-comparison}
\begin{tabular}{lcccc}
\toprule
\textbf{Dataset} & \textbf{Ours} & \textbf{TSAN \cite{cheng2021brain}} & \textbf{SFCN \cite{PENG2021101871}} & \textbf{3DResNet} \\
\midrule
SALD       & \textbf{5.35} & 5.91 & 5.48 & 6.94 \\
TrueCrime  & \textbf{4.79} & 5.18 & 6.09 & 12.71 \\
AgeRisk    & \textbf{5.12} & 5.57 & 7.20 & 8.85 \\
SUDMEX     & \textbf{4.87} & 5.29 & 7.12 & 10.26 \\
\midrule
Overall    & \textbf{5.15} & 5.65 & 6.14 & 8.59 \\
\bottomrule
\end{tabular}
\end{table}

\subsection{Training Time and Computational Efficiency}
\label{train_and_computational_efficiency}
To evaluate computational efficiency, we compared the end-to-end training durations of our proposed BrainRotViT framework with several baseline and state-of-the-art (SOTA) architectures. 

\begin{table}[h]
\centering
\caption{Comparison of approximate training times, model size, and computational cost across different models. All models were trained on identical hardware settings (RTX~3060, Ryzen~3700X CPU, 16\,GB RAM).}
\vspace{3pt}
\setlength{\tabcolsep}{4pt}
\renewcommand{\arraystretch}{1.1}
\footnotesize
\begin{tabular}{lcccc}
\hline
\textbf{Model} & \thead{\textbf{Training}\\\textbf{Time (h)}} & \thead{\textbf{Trainable}\\\textbf{Params}} & \thead{\textbf{GFLOPs}\\\textbf{/sample}} & \thead{\textbf{Peak}\\\textbf{VRAM}}\\
\hline
\textbf{BrainRotViT} \\\textbf{(CNN head)} & \textbf{4} & 50.4M & 1.6 & 0.23\,GB\\
\textbf{BrainRotViT} \\\textbf{(Transformer)} & \textbf{12} & 0.33M & $\approx$1,098\textsuperscript{$\dagger$} & 1.38\,GB\textsuperscript{\ddag} \\
SFCN & 20 & \textbf{2.96M} & 131.7 & 8.03\,GB\textsuperscript{\ddag}\\
3D-ResNet & 24 & 64.2M & 253.4 & 3.98\,GB\textsuperscript{\ddag}\\
3D-ViT & 30 & 28.1M & \textbf{112.5} & 4.99\,GB\textsuperscript{\ddag}\\
TSAN & 36 & 204.8M & 227.9 & 10.12\,GB\textsuperscript{\ddag}\\
Triamese-ViT & 52 & 283.1M & 277.2 & 6.07\,GB\textsuperscript{\ddag}\\
Global–Local \\Transformer & 60 & 20.4M & 114.3\textsuperscript{*} & \textbf{1.26}\,\textbf{GB}\textsuperscript{\ddag}\\
\hline
\end{tabular}
\begin{flushleft}
\footnotesize
\textsuperscript{$\dagger$}Transformer trained with DoRA over 32 training slices.\\
\textsuperscript{*}Reported as 38.1 GFLOPs $\times$ 3 for ensembling models for sagittal, coronal, and axial slices.\\
\textsuperscript{\ddag}Measured at each model's configured training batch size (weights + gradients + Adam optimizer state + activations): BrainRotViT ViT bs=8, 3D-ResNet bs=4, SFCN bs=4, 3D-ViT bs=4, TSAN bs=8, Triamese-ViT bs=2, Global--Local bs=4. Batch size selection for BrainRotViT was done by empirically optimizing for best accuracy. Batch sizes for other models were assigned as suggested in their original implementations, unless it exceeded the GPU memory of our setup, in which case it was reduced to the nearest multiple that fit. Patience with 20 epochs was applied as the stopping strategy for all models. See Supplementary Section~\ref{app:baselines} for further details.
\end{flushleft}
\label{tab:train_time}
\end{table}

As shown in Table~\ref{tab:train_time}, BrainRotViT achieves a favorable balance between accuracy and computational cost. Despite its hybrid design, it completes training in significantly less time than transformer-based models such as Triamese-ViT, which is the heaviest model based on parameter count, and Global–Local Transformer, which despite its lightweight structure is very slow to reach convergence. Our approach also outperforms both pure CNN and pure transformer baselines in predictive accuracy. This efficiency underscores the scalability of our approach for large, multi-site MRI datasets.

% The results indicate that the model generalizes effectively to unseen data distributions and acquisition protocols. Notably, datasets such as ADNI and Cam-CAN achieved the lowest MAEs (3.5–3.7 years), reflecting the model’s adaptability to high-quality, well-curated MRI cohorts. In contrast, slightly higher errors in COBRE and OASIS are attributed to smaller sample sizes and site-specific variability in scanning parameters. Nonetheless, all results remain within a tight error margin, confirming that the multi-cohort training strategy successfully mitigates dataset bias and promotes cross-site generalization. The consistent performance across diverse cohorts highlights the robustness of the proposed architecture and its suitability for large-scale, heterogeneous neuroimaging applications.

\subsection{Impact of Brain Aging on Alzheimer's Disease and Neurodegeneration}
\label{subsec:adni_analysis}
For the ADNI dataset, we split the 5,881 samples (drawn from 1,378 unique subjects) at the \emph{subject} level, assigning 80\% of subjects to training and 20\% to validation so that no subject's scans appeared in both sets (a per-scan random split would leak subject identity across repeat visits and inflate performance). This yielded 4,762 training and 1,119 validation samples, with subjects grouped into AD (Alzheimer’s Disease), MCI (Mild Cognitive Impairment), and CN (Cognitively Normal). Trained separately on this split, our architecture achieved an MAE of 3.52 years and an $R^2$ of 0.62 on the held-out validation set (Figure \ref{fig:adni_val}). Plotting the brain age-gaps (BAG) of each sample against their cognitive status group (Figures \ref{fig:ad_mci_high}, \ref{fig:ad_mci_pos_neg}), we found that subjects who exhibited extreme aging (subjects with a positive BAG greater than one standard deviation) were 5.49 times as likely to belong to the AD/MCI group rather than the CN group, compared to those who did not (odds ratio = 5.494, 95\% CI [3.744–8.061]; relative risk = 2.228, 95\% CI [1.950–2.547]; $p < 0.001$). Moreover, comparing extreme positive agers with extreme negative agers (BAG $<$ -1 std. dev.), we found that the former were 6.59 times as likely to belong to the AD/MCI group rather than CN (odds ratio = 6.591, 95\% CI [4.033–10.771]; relative risk = 2.528, 95\% CI [1.953–3.273]; $p < 0.001$). \\

Because brain-age predictions exhibit a systematic bias with chronological age (over-estimating age in younger and under-estimating it in older individuals), we applied the age-bias correction of de Lange and Cole \cite{delange2020correction}, i.e., the brain-age gap was regressed on chronological age in the training-set cognitively-normal controls, and the fitted trend was subtracted from every subject's gap to yield an age-independent corrected BAG. Associations between corrected BAG and diagnosis were assessed with multivariable logistic regression adjusting for age, sex, and acquisition site, reporting adjusted odds ratios with 95\% confidence intervals (See Figures \ref{fig:ad_mci_high}, \ref{fig:ad_mci_pos_neg}, and Supplementary Section~\ref{adni-age-gap-analyses} for details). After age-bias correction, each one–standard-deviation increase in corrected BAG was associated with 1.72-fold higher odds of belonging to AD/MCI groups (adjusted OR 1.72, 95\% CI [1.43–2.07]; $p < 10^{-8}$) and 2.28-fold higher odds of belonging to the AD group (adjusted OR 2.28, 95\% CI [1.73–3.00]; $p < 10^{-8}$), adjusting for age, sex, and acquisition site. These findings suggest that individuals whose brains undergo accelerated aging are more likely to belong to the Alzheimer's disease or mild cognitive impairment groups than individuals with normally-aged or relatively slow-aged brains, which is widely supported by the existing body of research \cite{franke2019ten, millar2023multimodal, biondo2022brain, wittens2024brain}.

\subsection{Impact of Autism Spectrum Disorder on Brain Aging}
\label{subsec:abide_analysis}
The Autism Brain Imaging Data Exchange II (ABIDE II) is a case-control study that
consists of roughly equal numbers of autistic and control subjects. For the ABIDE
dataset (Table \ref{tab:dataset_info}), we randomly split the 1,114 subject-disjont samples into
80\% for training and 20\% for validation. Trained separately on this split, our architecture achieved an
MAE of 1.13 years and an $R^2$ of 0.93 on the held-out validation set (Figure
\ref{fig:abide_val}). Plotting the brain age-gaps of each sample against their
diagnosis status (Figure \ref{fig:asd_high}), we found that Autism Spectrum
Disorder (ASD) subjects were 1.67 times as likely to exhibit accelerated brain
aging (BAG $>$ 1 std.\ dev.) compared to control subjects (odds ratio = 1.6736,
95\% CI [1.100--2.546]; $\chi^2$ $p = 0.0204$). This threshold-based association
remained significant after applying the age-bias correction of de Lange and
Cole~\cite{delange2020correction}, estimated only on the training-set controls to avoid
circularity (corrected-BAG odds ratio = 1.563, 95\% CI [1.059--2.307];
$p = 0.031$). Our findings suggest that individuals with ASD are more likely to
exhibit structural patterns of accelerated brain aging, which coincides with
studies that have found age-related changes to be more pronounced in autistic
subjects \cite{braden2019thinning, dickinson2022electrophysiological}.

\subsection{Attention Map Analysis}
\label{subsec:attention-map}

\begin{figure*}
    \centering
    % --- First composite row ---
    \begin{minipage}{0.52\textwidth}
        \begin{subfigure}[b]{\textwidth}
            \centering
            \includegraphics[width=\linewidth]{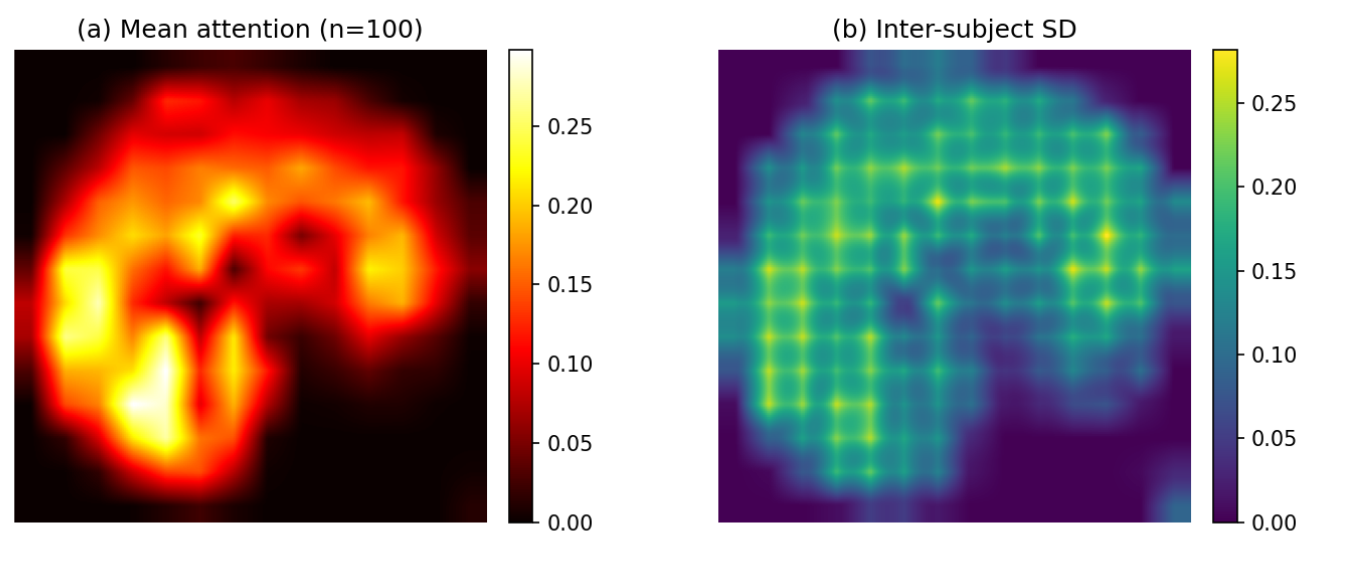}
            \caption{}
            \label{fig:attention_slice_heatmap}
        \end{subfigure}
        \vspace{2pt}
        \begin{subfigure}[b]{\textwidth}
            \centering
            \includegraphics[width=\linewidth]{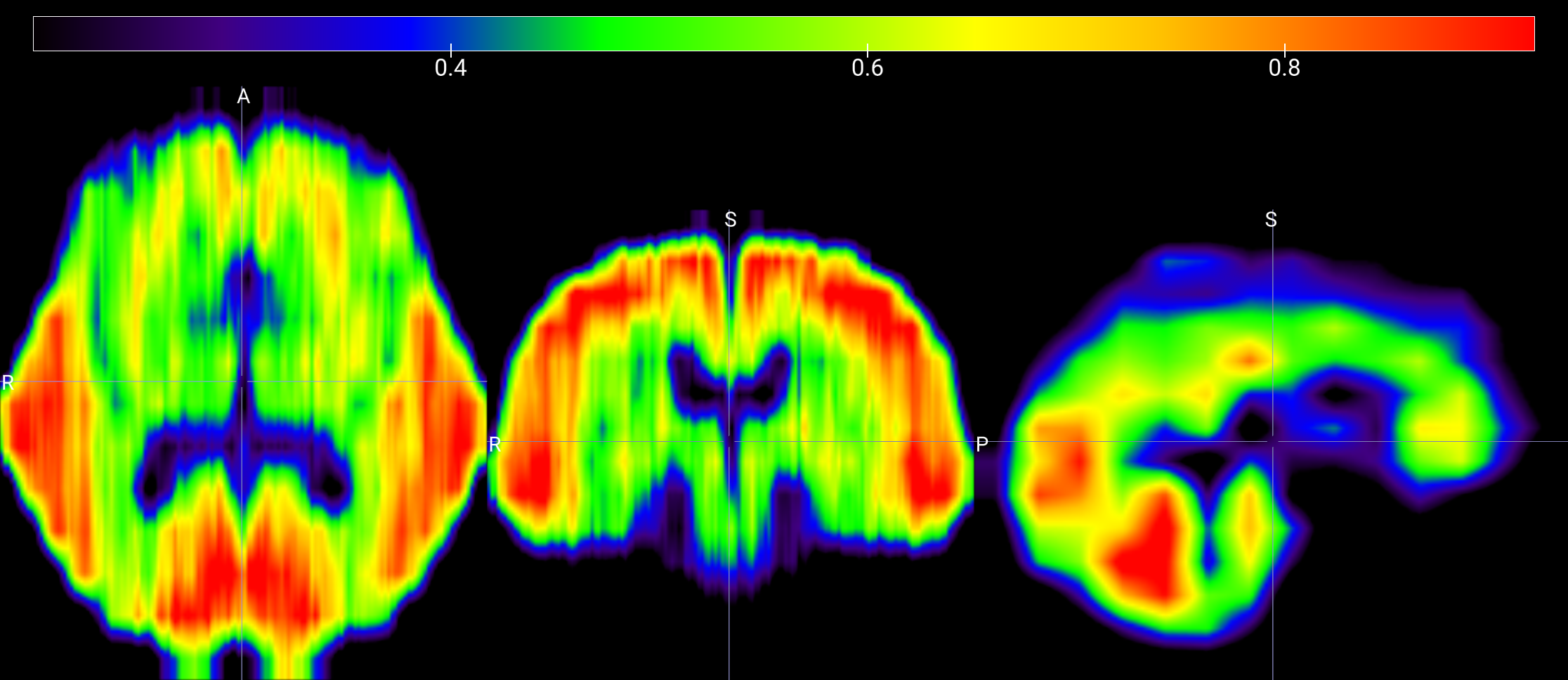}
            \caption{}
            \label{fig:att_3d_merged}
        \end{subfigure}
    \end{minipage}
    \begin{minipage}{0.28\textwidth}
        \begin{subfigure}[b]{\textwidth}
            \centering
            \includegraphics[width=\linewidth]{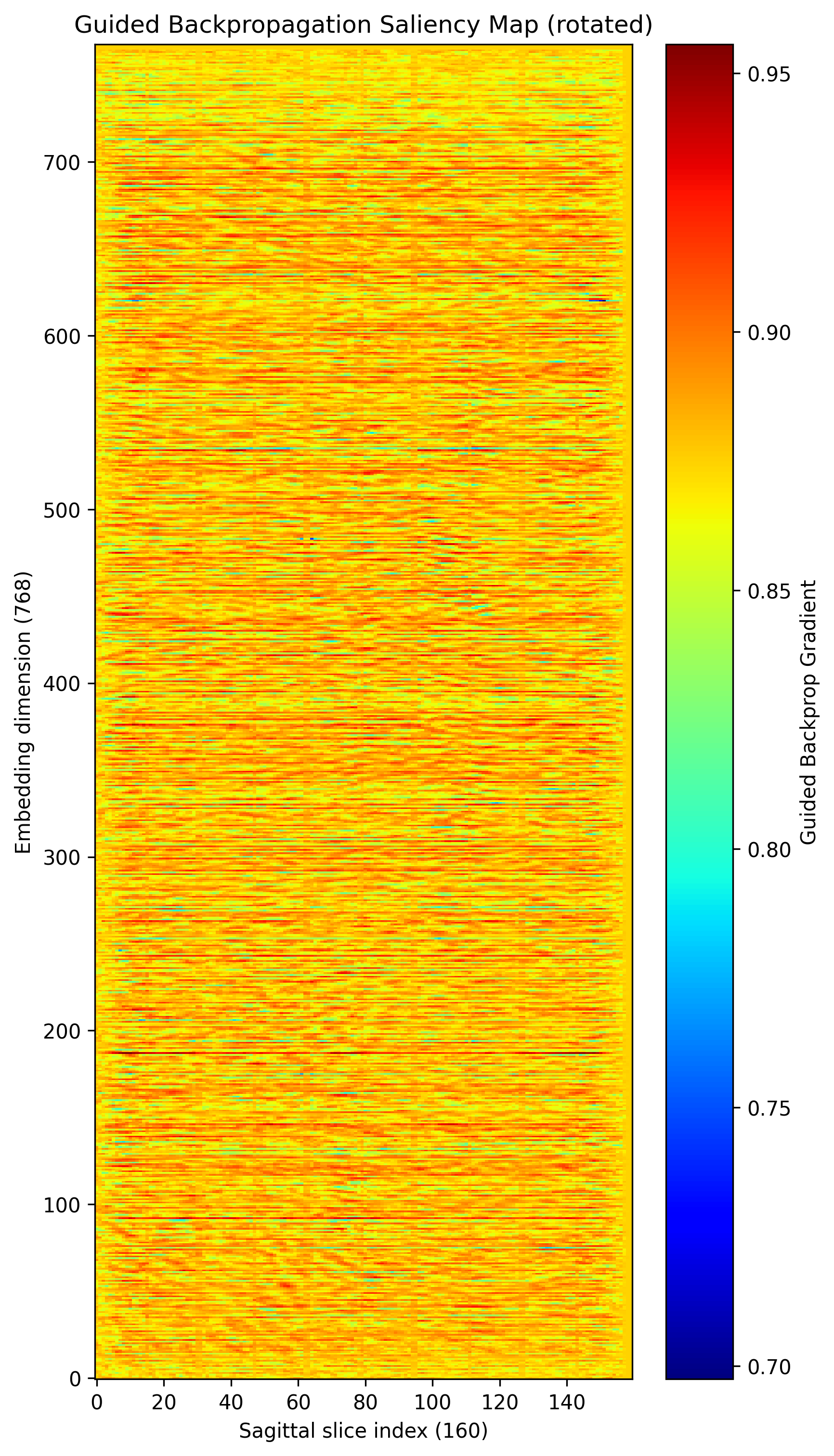}
            \caption{}
            \label{fig:gbp}
        \end{subfigure}
    \end{minipage}

    \vspace{4pt}

    % --- Row 1 ---
    \begin{subfigure}[b]{0.40\textwidth}
        \centering
        \includegraphics[width=\linewidth]{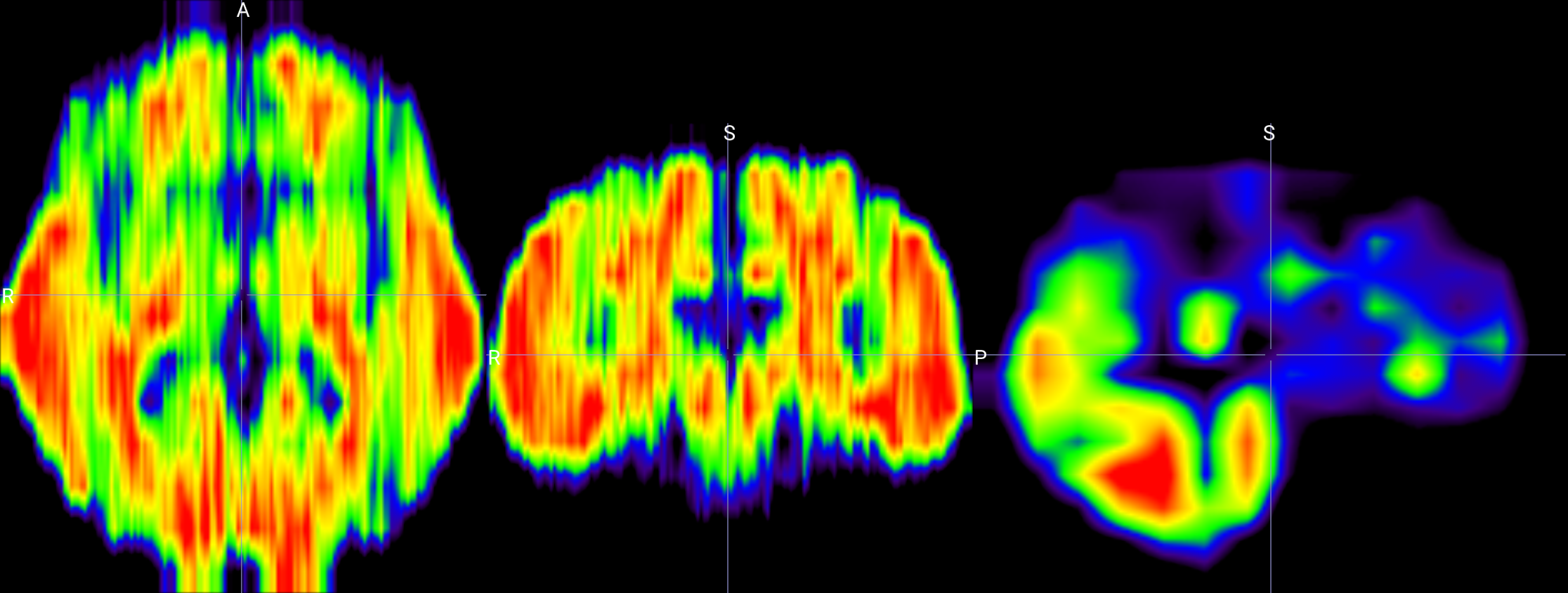}
        \caption{10–20 years}
        \label{fig:att10-20}
    \end{subfigure}
    \begin{subfigure}[b]{0.40\textwidth}
        \centering
        \includegraphics[width=\linewidth]{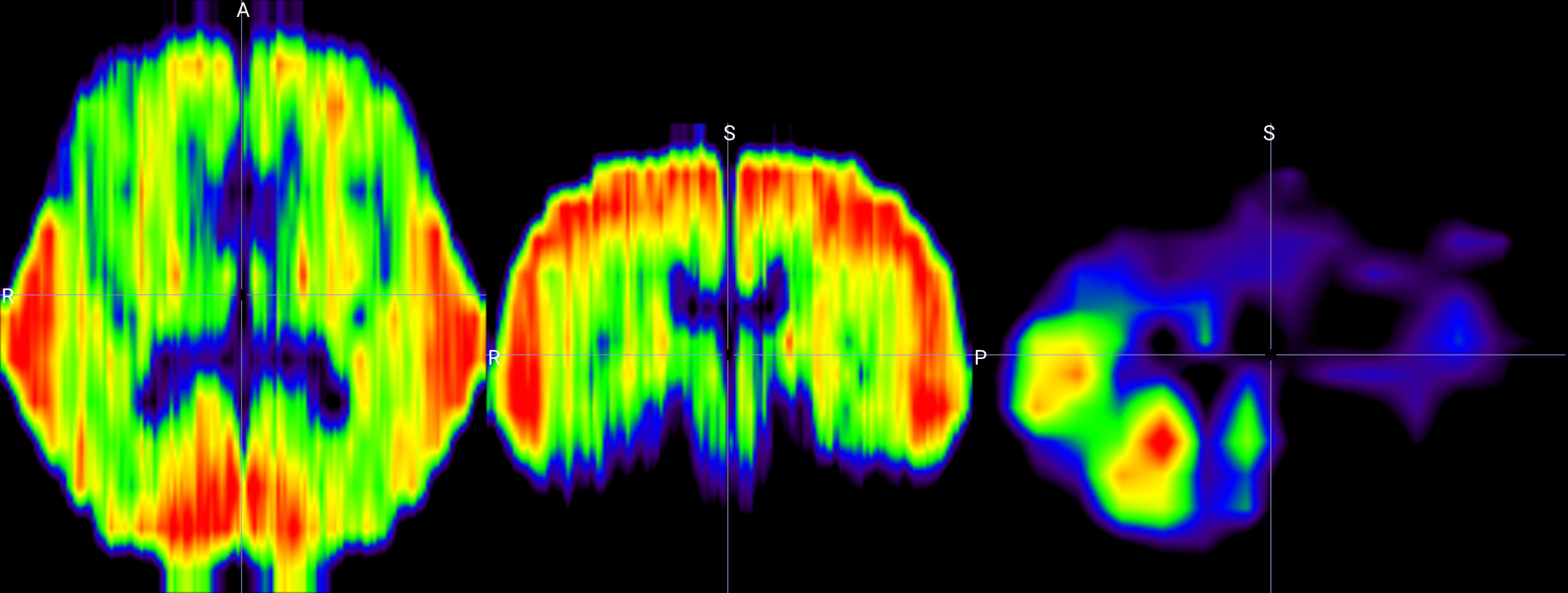}
        \caption{20–30 years}
        \label{fig:att20-30}
    \end{subfigure}

    \vspace{2pt}

    % --- Row 2 ---
    \begin{subfigure}[b]{0.40\textwidth}
        \centering
        \includegraphics[width=\linewidth]{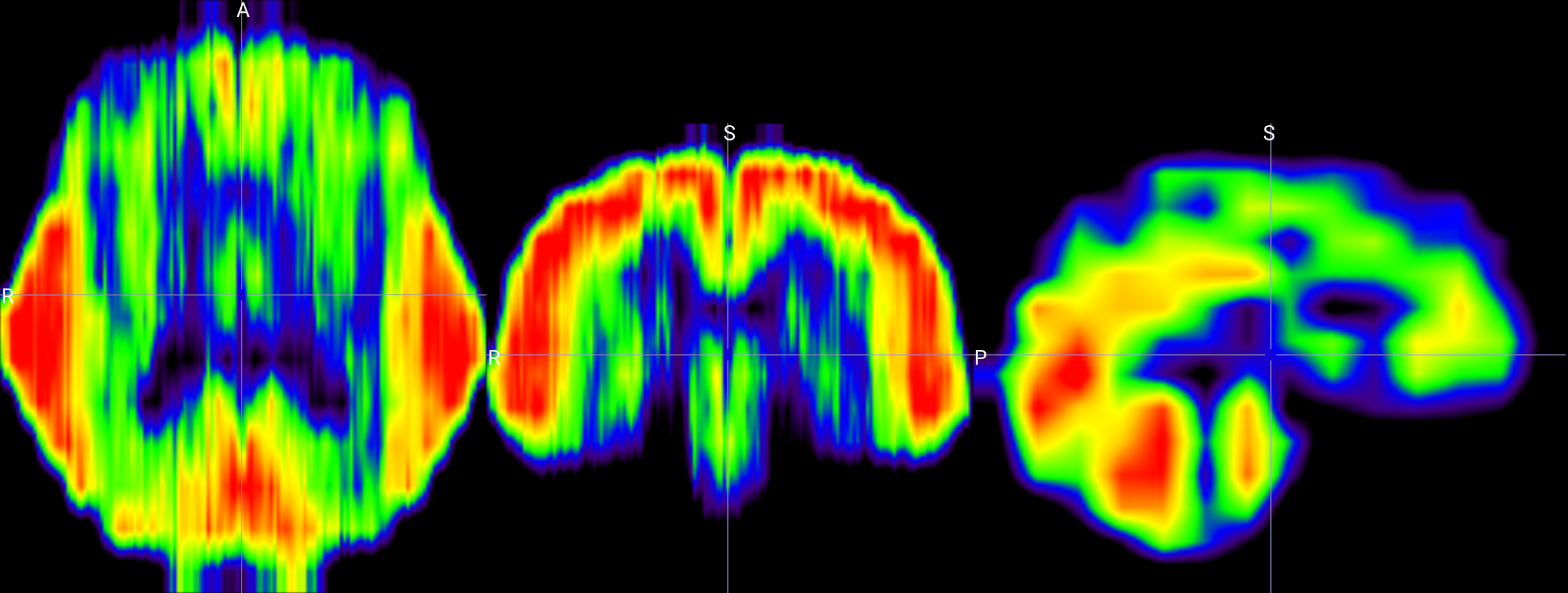}
        \caption{30–40 years}
        \label{fig:att30-40}
    \end{subfigure}
    \begin{subfigure}[b]{0.40\textwidth}
        \centering
        \includegraphics[width=\linewidth]{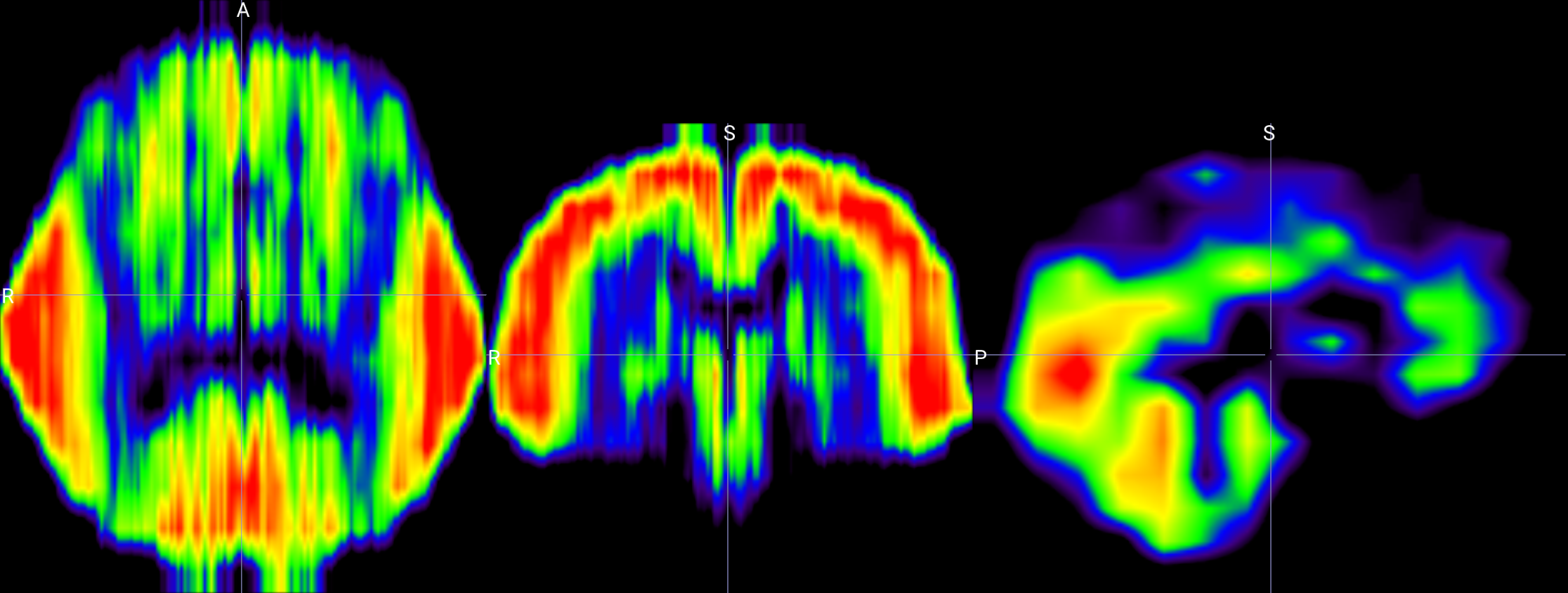}
        \caption{40–50 years}
        \label{fig:att40-50}
    \end{subfigure}

    \vspace{2pt}

    % --- Row 3 ---
    \begin{subfigure}[b]{0.40\textwidth}
        \centering
        \includegraphics[width=\linewidth]{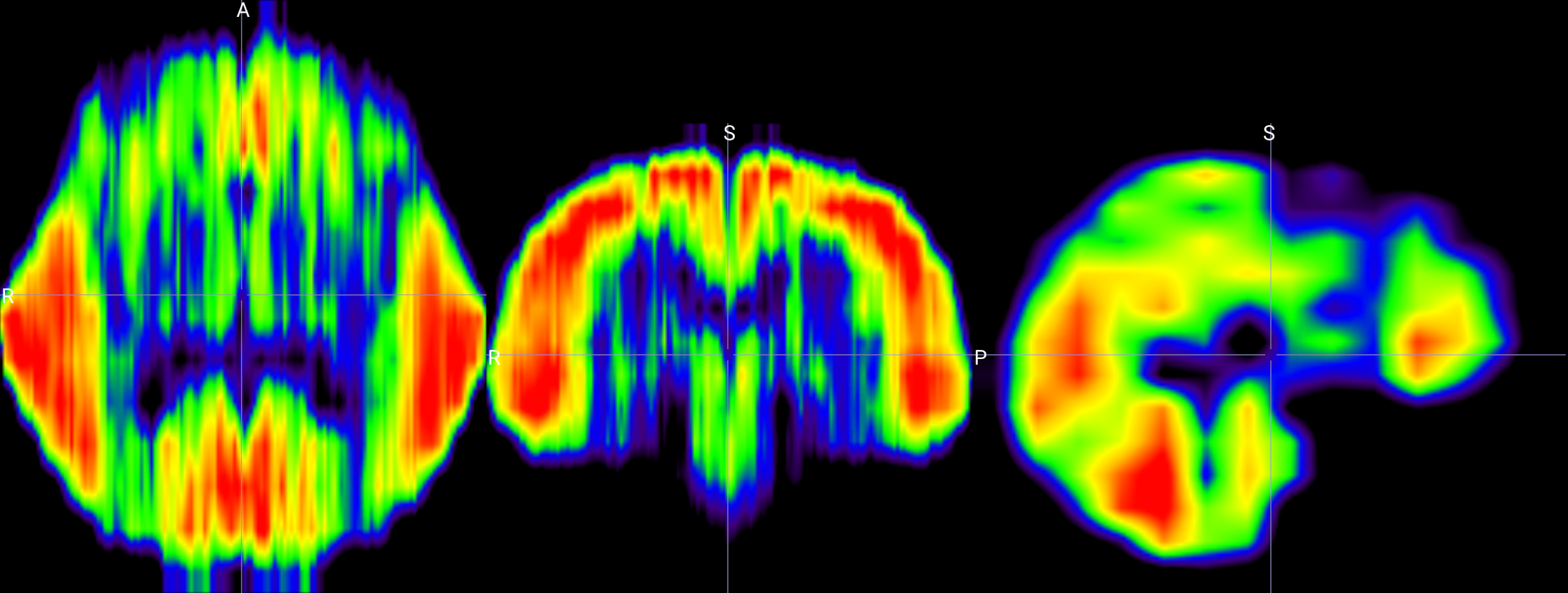}
        \caption{50–60 years}
        \label{fig:att50-60}
    \end{subfigure}
    \begin{subfigure}[b]{0.40\textwidth}
        \centering
        \includegraphics[width=\linewidth]{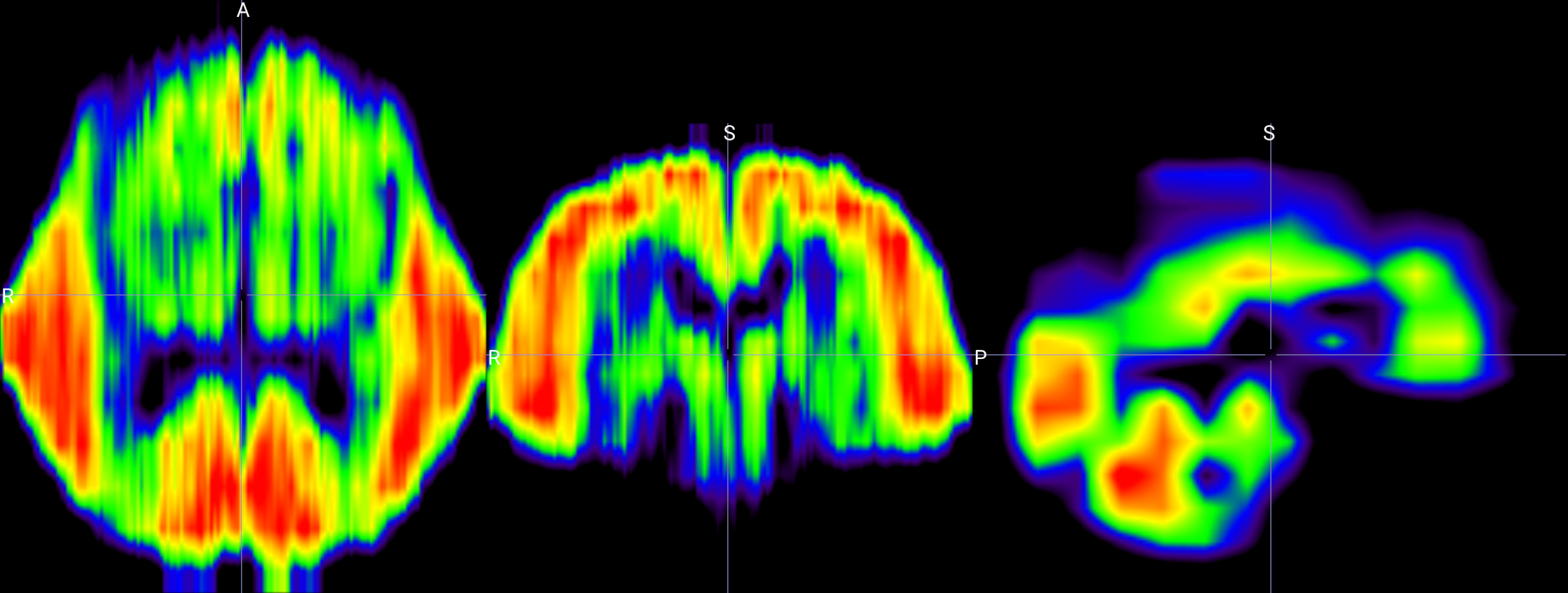}
        \caption{60–70 years}
        \label{fig:att60-70}
    \end{subfigure}

    \vspace{2pt}

    % --- Row 4 ---
    \begin{subfigure}[b]{0.40\textwidth}
        \centering
        \includegraphics[width=\linewidth]{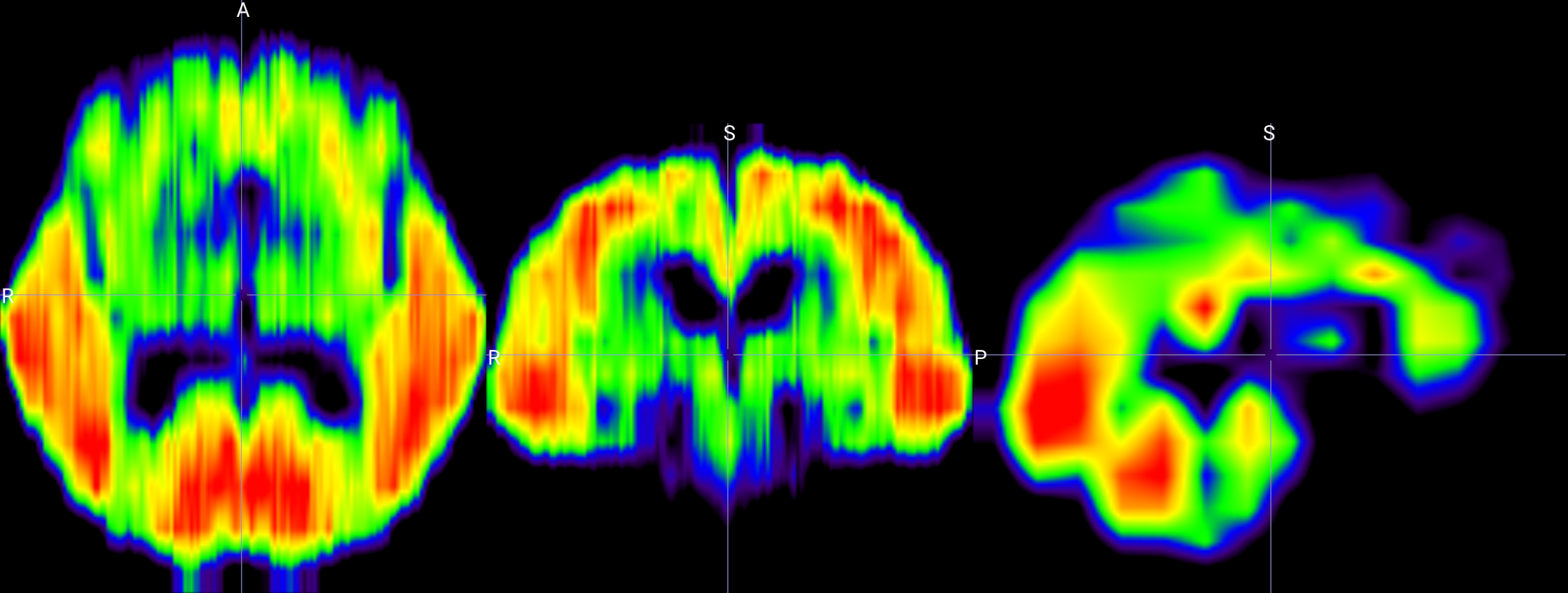}
        \caption{70–80 years}
        \label{fig:att70-80}
    \end{subfigure}
    \begin{subfigure}[b]{0.40\textwidth}
        \centering
        \includegraphics[width=\linewidth]{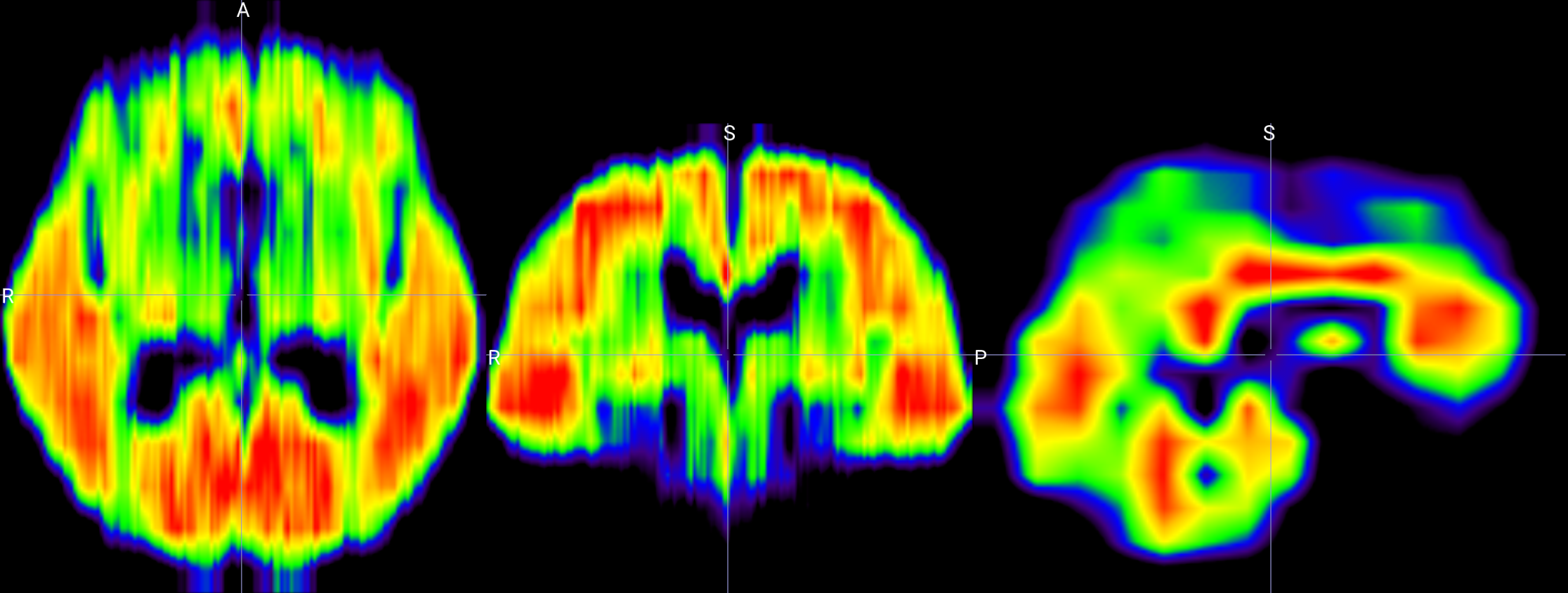}
        \caption{80–90 years}
        \label{fig:att80-90}
    \end{subfigure}

    \caption{\textbf{Model Interpretation.} 
    \textbf{(a):} Example of per-slice attention map extracted from vision transformer. (b) Mid-slice section views of fused 3D attention map along axial, coronal, and sagittal planes (from left to right).\textbf{(c):} Saliency map of trained residual CNN extracted through guided back propagation. 
    \textbf{(d)–(k):} 3D attention map previews by age range.}
    \label{fig:att_map_comp}
\end{figure*}

To further interpret the spatial focus of the Vision Transformer within the proposed BrainRotViT framework, we visualize the mean attention distribution across subjects in the validation set. Figure~\ref{fig:attention_slice_heatmap} illustrates the aggregated attention heatmap derived from the middle sagittal slice (slice index 80) averaged over randomly-selected 100 validation samples from Table \ref{tab:dataset_info}. The 3D attention visualization in Figure \ref{fig:att_3d_merged} reveals several neuroanatomical regions exhibiting strong activation from the model, corresponding to areas known to undergo age-related structural and functional changes. \\

The visual pathway of the brain emerges as a region of strong structural relationship with age in our interpretability pipeline. High attention is consistently assigned to the occipital lobe, particularly the region surrounding the calcarine sulcus on the medial surface of the occipital lobe, alongside the occipital gyrus. This aligns with prior work showing that the calcarine cortex thins with age \cite{salat2004thinning}, and that the calcarine fissure is among the regions most prominently affected by aging, with comparatively little additional atrophy attributable to Alzheimer's disease \cite{bakkour2013effects}. Argiris et al. similarly report that changes in the calcarine fissure are strongly associated with aging \cite{argiris2021quantifying}. Considering the occipital lobe as a whole, findings are more mixed: while Argiris et al. also note age-related changes in occipital cortex activation \cite{argiris2021quantifying}, Fujita et al. find that the occipital lobe undergoes comparatively modest structural change with age \cite{fujita2023characterization}, suggesting that the aging signal in this lobe may be concentrated around the calcarine region rather than distributed uniformly across it. Our model also assigns substantial attention to the fusiform gyrus, which shows major intensity changes across age groups. This is consistent with reports that the fusiform gyrus decreases in size with age \cite{shah2021impact}, and with transcriptomic evidence linking the fusiform gyrus to neurodegeneration and Alzheimer's disease through gene co-expression and differential co-expression network analyses using post-mortem tissue from cognitively healthy and AD-affected individuals \cite{ribeiro2023fusiform}, further supporting a structural relationship between this region and aging.\\

The lower and medial posterior regions of the cerebellum show high attention concentration, reflecting their involvement in age-associated decline in motor coordination and memory consolidation, as has been supported by several studies \cite{arleo2024consensus, woodruff2010differential, samstag2025neuropathological}. Particularly the cerebellar vermis, whose volume has been strongly associated with aging-related changes \cite{koppelmans2017regional, bernard2014moving, raz1992age}, appears as a region that our aging model assigns a high average intensity invariably across age ranges. Close to the cerebellum, the cuneus is another region exhibiting high intensity in our attention maps. While recent studies using functional MRI data have noted changes in cuneus activation with age \cite{heckner2021aging, knights2023neural}, our study uses structural data only, thus hinting that with aging the cuneal cortex undergoes structural changes as well.\\

The precentral cortex, which exhibited high attention concentration in our model, has been linked with age-associated changes quite commonly in the literature. Prior studies have identified thinning of the precentral cortex and the atrophy of the primary motor cortex as quantitative markers of aging \cite{c2011age, salat2004thinning}. One study compares such changes with the structural changes in the postcentral gyrus \cite{zhou2020diffusion}, which also appears as a high-attention region in our model. Functional differences in the primary motor cortex, which is located within the precentral cortex, have also been attributed to aging \cite{wen2025motor, cheng2018age, cirillo2025modulation}, thereby further substantiating the biological interpretability of our aging model. \\

Across all age ranges, the temporal lobes, particularly the middle temporal gyrus, remained as a prominent region of interest for our aging model. This coincides with several prior studies that have confirmed the role of the temporal lobes in aging and neurodegeneration \cite{bigler2002temporal, bartzokis2001age, hrybouski2023aging}. Moreover, comparing the volume of the region of relative intensity greater than 0.4 near the temporal gyrus between all age groups (Figures \ref{fig:att10-20}–\ref{fig:att80-90}), it was observed that the volume of the high-attention region in the temporal lobe increases with age, suggesting that changes in the temporal lobes accelerate with age, which is also supported by Fujita et al.~\cite{fujita2023characterization}.\\

While slightly higher than average attention concentration was observed in a part of the frontal lobe, in general the frontal lobe exhibited relatively lower attention weight compared to postcentral and cerebellar regions (Figure \ref{fig:att_3d_merged}). However, it was observed that the intensity of attention applied by the model to the frontal lobe varies significantly by age group (Figures \ref{fig:att10-20}–\ref{fig:att80-90}). Sampling the intensity in a region in the medial part of the superior frontal gyrus, relatively high intensity was found among samples in the 10-20 year age group. This suggests that the frontal lobe undergoes the most rapid change specifically in adolescence and early adulthood, as is backed up by numerous prior studies \cite{gogtay2004dynamic, giedd1999brain, lenroot2006brain, blakemore2012imaging}. Elderly subjects, particularly the 80-90 year age group, also exhibited high intensity of attention in the same region, thus substantiating prior studies that suggest that the frontal lobe undergoes faster structural change in the brains of older adults \cite{goh2013frontal, zhu2015reduced, li2013age, fujita2023characterization}.

% The inter-subject standard deviation map complements these observations by showing that attention variability is highest around the sensorimotor and parietal cortices, implying individualized aging trajectories across participants. Together, these results confirm that the Vision Transformer effectively localizes morphometric patterns most predictive of brain aging, aligning with established neurobiological evidence.

\section{Ablation Study}
\label{sec:ablation}
\begin{table*}[t]
\centering
\footnotesize
\renewcommand{\arraystretch}{1.1}
\newcommand{\abgroup}[1]{\multicolumn{4}{c}{\textit{#1}} \\ \cmidrule(lr){1-4}}
\newcommand{\abgap}{\addlinespace[3pt]}
\caption{Summary of ablation experiments. Each row varies one component of the proposed model; the selected configuration reaches a validation MAE of 3.43 years in every row. Alt.\ MAE is the range (or value) of validation MAE, in years, over the alternatives tested.}
\label{tab:ablation-summary}
\begin{tabular}{@{}p{3.3cm}p{6.9cm}p{4.2cm}c@{}}
\toprule
\textbf{Component} & \textbf{Alternatives tested} & \textbf{Selected} & \textbf{Alt.\ MAE} \\
\midrule
\abgroup{ViT pretraining}
ViT backbone & ViT-small, ViT-large, DeiT~\cite{deit}, DeiT-tiny, RoPE ViT~\cite{heo2024ropevit}, MaxViT~\cite{tu2022maxvit}, Swin-tiny~\cite{liu2021Swin} & ViT-base & 3.48--3.71 \\ \abgap
Age--sex group width & 5 and 20 years & 10 years & 4.72--5.26 \\ \abgap
Slices per volume & 16 and 64 & 32 & 3.77--3.95 \\ \abgap
Age/sex supervision & Age classes only; age classes + FiLM sex input; direct age regression (with or without FiLM sex input); no sex input in CNN head & Composite age--sex classes, sex fused in CNN head & 3.56--7.06 \\
\midrule
\abgroup{Regression head}
Front-end architecture & MLP only, ViT, Bi-GRU \cite{cho2014learning} \cite{schuster1997bidirectional}, Bidirectional Mamba \cite{gu2023mamba} \cite{zhu2024vision}, Sequential Transformer \cite{vaswani2017attention} & Residual CNN & 3.76--8.12 \\ \abgap
Convolutional blocks & 1, 2, 4, 5, 6 & 3 & 3.49--4.23 \\ \abgap
Kernel sizes & Three alternative size sets and a dilated variant & (10,60), (5,15), (2,6) & 3.47--3.82 \\ \abgap
Activation function & ReLU, Leaky ReLU, GELU & SiLU & 3.48--3.71 \\ \abgap
Convolution axes & 1D (slice axis only) & 2D & 3.64 \\ \abgap
Residual connections & Removed & Kept & 3.70 \\ \abgap
Late sex fusion & Removed & Kept & 3.56 \\
\midrule
\abgroup{Optimization}
Learning rate & $1\times10^{-4}$, $1\times10^{-3}$, $5\times10^{-3}$, $1\times10^{-2}$ & $5\times10^{-4}$ & 3.83--6.13 \\
\bottomrule
\end{tabular}
\end{table*}

To evaluate the contribution of individual components and hyperparameter choices, we ran twelve ablation experiments on the validation set. Each experiment changes one aspect of the proposed method and keeps all other settings fixed. Table~\ref{tab:ablation-summary} summarizes the alternatives tested and Figure~\ref{fig:ablation-grid} plots each comparison; per-experiment tables and discussion are in Supplementary Section~\ref{supp:ablation_details}, and multi-seed statistics are in Supplementary Section~\ref{app:ablations}.

The proposed configuration achieved the lowest MAE in all twelve experiments. The largest degradations came from how age and sex are used during pretraining (5.07--7.06 for the alternatives to composite age--sex classes), from front ends that do not model relationships across slice embeddings (MLP-only and ViT-only, 8.12 and 6.51), and from the learning rate and age--sex bin width. By contrast, the ViT variant, activation function, kernel configuration and the sequence-model front ends (Bi-GRU, Bi-Mamba, Sequential Transformer) each changed MAE by less than 0.4 years; the sequence models, which come closest, are compared over multiple seeds in Supplementary Section~\ref{sup:subsec:frontend}. Residual connections, late sex fusion and 2D rather than 1D convolution each gave a smaller but consistent gain (0.13--0.27 years).

\begin{figure*}[h]
  \centering
  % Row 1
  \begin{subfigure}[t]{0.32\textwidth}
    \centering
    \begin{tikzpicture}
      \begin{axis}[
        width=\linewidth,
        height=3.0cm,
        xlabel={Backbone}, ylabel={MAE},
        xtick=data,
        symbolic x coords={Base,DeiT,RoPE,MaxViT,Swin},
        xticklabel style={rotate=20,anchor=east,font=\small},
        ymin=3.2, ymax=3.7
      ]
        \addplot+[mark=*, thick] coordinates {
          (Base,3.43) (DeiT,3.59) (RoPE,3.62) (MaxViT,3.48) (Swin,3.56)
        };
      \end{axis}
    \end{tikzpicture}
    \caption{ViT variants}
  \end{subfigure}\hfill
  \begin{subfigure}[t]{0.32\textwidth}
    \centering
    \begin{tikzpicture}
      \begin{axis}[
        width=\linewidth, height=3.0cm,
        title={},
        xlabel={Group Width (years)}, ylabel={MAE},
        xtick=data, symbolic x coords={5,10,20},
        ymin=3, ymax=6
      ]
        \addplot+[mark=o] coordinates {(5,4.72) (10,3.43) (20,5.26)};
      \end{axis}
    \end{tikzpicture}
    \caption{Age–sex granularity}
  \end{subfigure}\hfill
  \begin{subfigure}[t]{0.32\textwidth}
    \centering
    \begin{tikzpicture}
      \begin{axis}[
        width=\linewidth, height=3.0cm,
        title={},
        xlabel={Slices}, ylabel={MAE},
        xtick=data, symbolic x coords={16,32,64},
        ymin=3.2, ymax=4.2
      ]
        \addplot+[mark=*, thick] coordinates {(16,3.95) (32,3.43) (64,3.77)};
      \end{axis}
    \end{tikzpicture}
    \caption{Slices per volume}
  \end{subfigure}

  \vspace{6pt}

  % Row 2
  \begin{subfigure}[t]{0.32\textwidth}
  \centering
  \begin{tikzpicture}
    \begin{axis}[
      width=\linewidth, height=3.0cm,
      title={},
      xlabel={}, ylabel={MAE},
      xtick=data,
      symbolic x coords={CNN,bi-GRU,bi-Mamba,Seq.-Trans.},
      xticklabel style={rotate=25,anchor=east,font=\footnotesize},
      ymin=3.4, ymax=4
    ]
      \addplot+[mark=o] coordinates {
        (CNN,3.43)
        (bi-GRU,3.76)
        (bi-Mamba,3.81)
        (Seq.-Trans.,3.76)
      };
    \end{axis}
  \end{tikzpicture}
  \caption{Front-end architectures}
\end{subfigure}
  \begin{subfigure}[t]{0.32\textwidth}
    \centering
    \begin{tikzpicture}
      \begin{axis}[
        width=\linewidth, height=3.0cm,
        title={},
        xlabel={}, ylabel={MAE},
        xtick=data,
        symbolic x coords={
          Composite,
          Age only,
          Age+FiLM,
          Regr.,
          Regr.+FiLM,
          Enc. only
        },
        xticklabel style={rotate=30,anchor=east,font=\footnotesize},
        ymin=3, ymax=7.5
      ]
        \addplot+[mark=*, thick] coordinates {
          (Composite,3.43)
          (Age only,5.07)
          (Age+FiLM,5.76)
          (Regr.,5.35)
          (Regr.+FiLM,7.06)
          (Enc. only,3.56)
        };
      \end{axis}
    \end{tikzpicture}
    \caption{ViT + CNN variants}
  \end{subfigure}\hfill
  \begin{subfigure}[t]{0.32\textwidth}
    \centering
    \begin{tikzpicture}
      \begin{axis}[
        width=\linewidth, height=3.0cm,
        title={},
        xlabel={\# conv blocks}, ylabel={MAE},
        xtick=data, symbolic x coords={1,2,3,4,5,6},
        ymin=3.2, ymax=4.5
      ]
        \addplot+[mark=*, thick] coordinates {
          (1,4.23) (2,3.76) (3,3.43) (4,3.49) (5,3.54) (6,3.73)
        };
      \end{axis}
    \end{tikzpicture}
    \caption{Number of conv blocks}
  \end{subfigure}

  \vspace{6pt}

  % Row 3
  \begin{subfigure}[t]{0.32\textwidth}
    \centering
    \begin{tikzpicture}
      \begin{axis}[
        width=\linewidth, height=3.0cm,
        title={},
        xlabel={}, ylabel={MAE},
        xtick=data,
        symbolic x coords={
          K1,
          K2,
          K3,
          K4,
          Dilated
        },
        xticklabel style={font=\small},
        ymin=3.2, ymax=4.0
      ]
        \addplot+[mark=o] coordinates {
          (K1,3.43)
          (K2,3.47)
          (K3,3.82)
          (K4,3.54)
          (Dilated,3.50)
        };
      \end{axis}
    \end{tikzpicture}
    \caption{Kernel sizes: K1=(10,60,5,15,2,6), K2=(12,80,6,20,3,8), K3=(6,40,4,12,2,6), K4=(8,50,5,15,1,4)}
  \end{subfigure}\hfill
  \begin{subfigure}[t]{0.32\textwidth}
    \centering
    \begin{tikzpicture}
      \begin{axis}[
        width=\linewidth, height=3.0cm,
        title={},
        xlabel={}, ylabel={MAE},
        xtick=data, symbolic x coords={SiLU,ReLU,L-ReLU,GELU},
        xticklabel style={font=\small},
        ymin=3.2, ymax=3.8
      ]
        \addplot+[mark=*, thick] coordinates {(SiLU,3.43) (ReLU,3.48) (L-ReLU,3.52) (GELU,3.71)};
      \end{axis}
    \end{tikzpicture}
    \caption{Activation functions}
  \end{subfigure}\hfill
  \begin{subfigure}[t]{0.32\textwidth}
    \centering
    \begin{tikzpicture}
      \begin{axis}[
        width=\linewidth, height=3.0cm,
        ybar,
        bar width=25pt,        % wider bars
        yticklabel style={font=\small},
        title={},
        xlabel={}, ylabel={Val. MAE},
        symbolic x coords={With,Without},
        xtick=data,
        ymin=3.3, ymax=3.6,
        xticklabel style={font=\small},
        enlarge x limits=0.6   % reduces the gap between bars
      ]
        \addplot coordinates {(With,3.43) (Without,3.56)};
      \end{axis}
    \end{tikzpicture}
    \caption{Sex fusion}
\end{subfigure}

  \vspace{6pt}

    % Row 4
    \begin{subfigure}[t]{0.32\textwidth}
      \centering
      \begin{tikzpicture}
        \begin{axis}[
          width=\linewidth, height=3.0cm,
          title={},
          xlabel={LR}, ylabel={MAE},
          xtick=data, symbolic x coords={1e-4,5e-4,1e-3,5e-3,1e-2},
          xticklabel style={rotate=20,anchor=east,font=\footnotesize}, 
          ymin=3, ymax=7
        ]
          \addplot+[mark=*, thick] coordinates {
            (1e-4,4.52) (5e-4,3.43) (1e-3,3.83) (5e-3,5.10) (1e-2,6.13)
          };
        \end{axis}
      \end{tikzpicture}
      \caption{Learning rate (optimal: 5e-4)}
    \end{subfigure}
\begin{subfigure}[t]{0.32\textwidth}
    \centering
    \begin{tikzpicture}
      \begin{axis}[
        width=\linewidth, height=3.0cm,
        ybar,
        bar width=25pt,          % wider bars
        title={},
        xlabel={}, ylabel={MAE},
        symbolic x coords={With,Without},
        xtick=data,
        ymin=3, ymax=4,
        xticklabel style={font=\small},
        enlarge x limits=0.6     % reduced gap between bars
      ]
        \addplot coordinates {(With,3.43) (Without,3.70)};
      \end{axis}
    \end{tikzpicture}
    \caption{Residual connections}
\end{subfigure}\hfill 
\hfill
\begin{subfigure}[t]{0.32\textwidth}
  \centering
  \begin{tikzpicture}
    \begin{axis}[
      width=\linewidth, height=3.0cm,
      ybar,
      bar width=25pt,
      title={},
      xlabel={}, ylabel={MAE},
      symbolic x coords={2D CNN,1D CNN},
      xtick=data,
      ymin=3.2, ymax=3.8,
      xticklabel style={font=\small},
      enlarge x limits=0.6
    ]
      \addplot coordinates {
        (2D CNN,3.43)
        (1D CNN,3.64)
      };
    \end{axis}
  \end{tikzpicture}
  \caption{2D vs.\ 1D convolution}
\end{subfigure}

  \caption{\textbf{Ablation study results.}
\textbf{(a)} Comparison of ViT backbone architectures;
\textbf{(b)} Impact of age--sex group granularity;
\textbf{(c)} Effect of slice count per volume;
\textbf{(d)} Front-end architecture comparison;
\textbf{(e)} Sex information integration methods;
\textbf{(f)} Optimal depth of convolutional blocks;
\textbf{(g)} Kernel size configurations;
\textbf{(h)} Activation function comparison;
\textbf{(i)} Sex fusion effectiveness;
\textbf{(j)} Learning rate sensitivity;
\textbf{(k)} Impact of residual connections;
\textbf{(l)} Comparison of 2D and 1D convolutional front-ends.}
  \label{fig:ablation-grid}
\end{figure*}

\section{Discussion}
\label{sec:discussion}
In this study, we propose a framework that demonstrates how combining the global contextual modeling capacity of Vision Transformers with the spatial precision of residual CNNs yields a powerful and interpretable architecture for brain-age estimation. Our BrainRotViT model achieved the lowest MAE among the re-trained CNN- and transformer-based comparison methods, by a modest margin over the strongest competitor (TSAN) on the validation set, while requiring less training time, lower peak memory, and fewer trainable parameters than TSAN. Although we strive to make fair comparisons with methods proposed in prior studies, it remains an inherently complicated task. It is common in this domain for studies to compare against others by directly quoting the reported metrics \cite{Aghaei2024, rajabli, KIANIAN2024127974}; however, this does not accurately represent the true learning capabilities of those models. Each study uses its own dataset collection with a unique distribution, and to ensure a fair comparison, all methods should be trained and validated on the same data distribution. In our work, we made an effort to achieve fairness by re-running the training, validation, and testing pipelines of prior methods on an identical distribution of MRI images using a common upstream preprocessing pipeline, with model-specific adaptations where required.\\

To support evaluation on external datasets, we curated an extensive collection of open-access sMRI images from 15 different global datasets. Such a diverse data distribution is particularly vulnerable to site-specific artifacts, differences in scanner protocols, and demographic variability. Although the heterogeneity of this data somewhat affects the validation MAE, our model achieves competitive predictive performance on individual datasets (Figures \ref{fig:adni_val}, \ref{fig:abide_val}), while also showing consistent MAE-based performance across the external datasets (Tables \ref{tab:independent-testing}, \ref{tab:independent-testing-comparison}) and training faster than the comparison models. These results are encouraging regarding the model's performance under cross-cohort/site variation, which is an important property for translational neuroimaging applications, although they do not establish invariance to site effects.\\

The comparison with the baseline architectures in Table \ref{tab:validation-comparison} demonstrates that the two canonical paradigms fail in complementary ways, motivating the proposed hybrid design. The pure 3D Vision Transformer records the highest error (MAE = 6.37 years) and weakest correlation ($r = 0.81$), consistent with the large data requirements of transformer architectures. With only 4,086 training volumes, the model is unlikely to learn robust spatial priors, causing attention to capture dataset-specific noise rather than anatomically meaningful patterns. In contrast, the 3D-ResNet performs better (MAE = 5.55 years) owing to the strong inductive bias of convolutions, but its local receptive fields limit whole-brain context modeling. BrainRotViT addresses this trade-off by combining globally informed representations from a frozen, classification-pretrained ViT with efficient spatial aggregation by a lightweight residual CNN. Consequently, it achieves not only the lowest MAE but also the highest $R^2$ (0.96). Although TSAN attains a slightly higher Spearman correlation ($\rho = 0.98$), its lower $R^2$ (0.92) indicates greater scatter around the identity line, suggesting that BrainRotViT's predictions lie somewhat closer to chronological age while maintaining strong rank ordering.\\

The external validation results suggest that the apparent discrepancy between MAE and $R^2$ across cohorts is likely explained largely by differences in age distributions rather than inconsistent predictive performance. Although the SUDMEX and TrueCrime cohorts achieve the lowest MAEs (4.87 and 4.79 years), they also exhibit the lowest $R^2$ values (0.35 and 0.50). This is consistent with a statistical property of $R^2$ and does not by itself indicate poorer absolute prediction accuracy. Both cohorts have comparatively narrow age distributions (SD = 7.70 and 8.80 years), whereas SALD and AgeRisk span much broader ranges (SD = 17.44 and 19.27 years). Because $R^2$ measures explained variance relative to the total variance in the data, a compressed age range causes even modest residual errors to account for a larger proportion of the variance, lowering $R^2$. In contrast, MAE is independent of the target variance and therefore provides a more reliable measure of absolute prediction accuracy across heterogeneous cohorts. Accordingly, we regard MAE as the primary metric for assessing cross-cohort generalisation, with cohort-specific $R^2$ serving as a secondary, range-dependent indicator.\\

The comparison across external cohorts is best interpreted as an evaluation of performance under distribution shift rather than a simple ranking of competing methods. Although all models exhibit some performance degradation on unseen datasets, the extent of this degradation varies substantially across architectures. The 3D-ResNet degrades most markedly, with its MAE increasing from 5.55 years on the validation set to 12.71 years on the TrueCrime cohort, suggesting that this model may rely on site-specific image characteristics that do not generalise well across acquisition sites. BrainRotViT retains the lowest MAE on every external cohort, and its margin over TSAN widens from 0.29 years on the validation set to 0.50 years on the external cohorts; SFCN, whose validation error is higher, shows a smaller absolute increase (5.03 to 6.14 years, compared with 3.43 to 5.15 years for BrainRotViT). We hypothesise that the classification-pretrained, frozen Vision Transformer learns representations that are less dependent on site-specific characteristics, whereas the lightweight CNN focuses solely on spatial aggregation for age regression. Performance under distribution shift is particularly important for clinical deployment, where scanners, populations, and imaging protocols often differ from those seen during training.\\

Another factor influencing cross-cohort performance may be the demographic composition of the training data. Although the training pool combines multiple international datasets, it is dominated by cohorts of predominantly European ancestry, with only limited East Asian representation through resources such as FCON1000 and CORR. In this context, SALD, an entirely Chinese cohort, records the highest external MAE (5.35 years) and, as the largest external dataset (494 of 948 samples), contributes substantially to the overall error. This observation is consistent with previous reports that brain-age models trained primarily on European-ancestry populations exhibit higher prediction errors on East Asian cohorts \cite{GRAVINA2026246}, as well as broader evidence of reduced cross-ethnicity generalisation in neuroimaging prediction models \cite{Li2022CrossEthnicity}. The lower correlation metrics observed for SUDMEX are more likely explained by its narrow age distribution, although an additional contribution from distribution shift cannot be excluded. Because individual-level ancestry information was unavailable, these observations should be interpreted as motivation for future fairness-aware evaluation rather than as direct evidence of ancestry-specific model bias.\\

% Although our model effectively learns the regions of the brain associated with aging, it exhibited relatively low attention to the frontal cortex compared to the higher attention observed in post-frontal cortical regions and the cerebellum. We found that substantial attention to the frontal lobe occurred primarily among subjects aged 10–20 years, suggesting that frontal cortical changes are less prominent in older individuals. This finding contrasts with prior literature suggesting that the frontal cortex is, in general, the region most susceptible to age-related change \cite{peters2006ageing, blinkouskaya2021brain}, but nonetheless offers a new perspective on studying brain aging across the human lifespan.\\

Our analysis of brain age gaps among autistic and control subjects in the case–control ABIDE-II study suggests that autistic subjects are more likely to exhibit higher age gaps, i.e., to show accelerated aging. While prior studies have reported that aging-related structural changes appear more pronounced among autistic individuals, there is no direct evidence in the literature to suggest that their brains age faster than those of controls. In fact, one study reported lower age gaps among ASD subjects compared to controls \cite{wang2021predicting}, supporting the notion of delayed development in autism. However, we did not observe any such statistically significant pattern in our brain age-gap data.\\

The design of our learning architecture is novel in its combination of a transformer encoder with a convolutional regression module that applies convolutions over a 2D matrix of concatenated embeddings. While sagittal slices clearly follow a spatial axis, treating the embedding dimension of transformer outputs as spatial may seem semantically inappropriate. Embedding vectors represent abstract features, and there is no inherent reason to expect a direct correspondence between the positional values of two embeddings produced by the same transformer. However, we argue that neighboring sagittal slices of a 3D MRI are so anatomically similar that a vision transformer will tend to embed them in similar ways, producing a structured slice-by-embedding representation that the residual CNN can effectively exploit. This should not be interpreted as implying an inherent spatial or semantic ordering of the embedding dimensions themselves; rather, it reflects the empirical tendency of anatomically similar slices to be embedded similarly. To confirm this empirically, we calculated the cosine similarity among all slice embeddings for each 3D MRI and found that neighboring slice embeddings are indeed highly similar, as are embeddings of slices that are antipodal in sagittal symmetry (Figure \ref{fig:embed_similarity_heatmap}). We also replaced the CNN in the architecture's second stage with sequence models that are more semantically aligned with embedding sequences (a Bi-GRU, a sequential Transformer, and a bidirectional Mamba model), but none of them matched the residual CNN (Table~\ref{tab:ablation-summary}; Supplementary Section~\ref{sup:subsec:frontend}). Finally, a 1D CNN that convolves only along the slice axis performed slightly but consistently worse than the 2D CNN across five random seeds ($3.593 \pm 0.039$ vs.\ $3.496 \pm 0.047$ years; Supplementary Section~\ref{sup:subsec:2dv1d}). We take this as empirical support for the 2D formulation, not as evidence that the embedding dimensions possess a spatial ordering; the gain may instead reflect the additional weight sharing and local feature mixing introduced by 2D kernels.

We also investigated whether the proposed architecture could instead be trained end-to-end using a single regression loss, rather than through the two-stage process described above. Specifically, we connected the embeddings generated for each of the 160 slices by the ViT directly to the CNN and trained the resulting network jointly via backpropagation from a single MSE loss, using DoRA together with gradient checkpointing so that the trainable parameter count (50.73M) matched that of our two-stage pipeline (Table~\ref{tab:train_time}). With the same 20-epoch early-stopping patience applied to all other models, training ran for 42 epochs and reached a validation MAE of 4.17 years, compared to 3.43 years for BrainRotViT. Joint training was also substantially more demanding, requiring approximately 33 TFLOPs per sample, 26\,GB of peak VRAM, and approximately 56 hours of training on an NVIDIA H100 GPU (Section~\ref{subsec:environment}), whereas the two stages of BrainRotViT together require approximately 1.1 TFLOPs per sample, under 1.4\,GB of peak VRAM, and approximately 16 hours on a consumer RTX~3060. Decoupling the two stages therefore yields better accuracy at a fraction of the computational cost, likely because end-to-end optimization must propagate gradients through a considerably deeper joint parameter space.\\

This resource-driven design philosophy extends to the architecture as a whole. Wider access to more powerful GPUs and larger memory would enable the exploration of more sophisticated architectures, such as employing 3D attention or multiple vision transformers coupled to a CNN optimized under a shared loss function. Yet these very constraints encouraged us to design a model that is not only accurate but also lightweight and fast to train compared to those proposed in previous studies. Beyond improving brain-age prediction, we believe the proposed architecture itself holds promise for broader applications. By modifying the regression or classification heads, it could be adapted to classify cognitive status, predict conversion from mild impairment to Alzheimer's disease, or diagnose other neurological conditions from structural MRI patterns, given the availability of sufficiently large imaging datasets. More broadly, because our framework does not depend on prior anatomical knowledge to manually select specific regions or slices, but instead learns directly from the whole 3D image through representation learning, it is not inherently tied to brain anatomy. In principle, the same architecture could be used in other 3D medical imaging domains, thus applying the method in a domain-agnostic manner.

% \section{Conclusion}
% \label{sec:conclusion}

\section{Acknowledgments}
\label{sec:ack}
Data used in the preparation of this article were obtained from the Alzheimer's Disease Neuroimaging Initiative (ADNI) database. Data collection and sharing for the Alzheimer's Disease Neuroimaging Initiative (ADNI) is funded by the National  Institute on Aging (National Institutes of Health Grant U19AG024904). The grantee organization is the Northern California Institute for Research and Education. In the past, ADNI has also received funding from the National Institute of Biomedical Imaging and Bioengineering, the Canadian Institutes of Health Research, and generous private sector contributions through the Foundation for the National Institutes of Health (FNIH) from several companies and non-profits (See https://adni.loni.usc.edu for full up-to-date information). Data collection and sharing for this project was also partly provided by the Cambridge Centre for Ageing and Neuroscience (CamCAN). CamCAN funding was provided by the UK Biotechnology and Biological Sciences Research Council (grant number BB/H008217/1), together with support from the UK Medical Research Council and University of Cambridge, UK. We thank Annette Weekes-Holder, Garry Detzler, and Baycrest Imaging Centre staff for their help with data collection for the BOLD variability dataset. In the context of the DLBS dataset, we thank the former lab members and associates who have made this research possible, and who are listed in the Keys to the Kingdom. In addition, we thank the participants who generously contributed their time and effort to the DLBS project. This study made use of data from the IXI dataset (https://brain-development.org/ixi-dataset/), which are made available under the Creative Commons CC BY-SA 3.0 license. Data used in the study were provided in part by OASIS (OASIS-1: Cross-Sectional: Principal Investigators: D. Marcus, R, Buckner, J, Csernansky J. Morris; P50 AG05681, P01 AG03991, P01 AG026276, R01 AG021910, P20 MH071616, U24 RR021382). For all the other open-access datasets used in our study, we cordially thank the Neuroimaging Tools and Resources Collaboratory (NITRC) and the OpenNeuro (https://openneuro.org/) initiative for providing the data through their image repositories in easily accessible formats. We sincerely thank Chris Rorden for developing the open-source tool MRIcroGL \cite{rorden2025mricrogl}, which was thoroughly used in our whole workflow to view and understand 3D MRI Images, as well as to generate the attention map images for model interpretation.

\bibliographystyle{ieeetr}
\begingroup
\small
\bibliography{bibliography}
\endgroup

\onecolumn   % Switch to single-column format (forces page break)
\pagenumbering{arabic} 
\setcounter{figure}{0}
\renewcommand{\thefigure}{S\arabic{figure}}
\setcounter{table}{0}
\renewcommand{\thetable}{S\arabic{table}}
\setcounter{section}{0}
% % Options for packages loaded elsewhere

% \PassOptionsToPackage{unicode}{hyperref}
% \PassOptionsToPackage{hyphens}{url}
% \documentclass[
% ]{article}
% \usepackage{xcolor}
% \usepackage[a4paper,top=2.5cm,bottom=2.5cm,left=1.5cm,right=1.5cm,marginparwidth=1.25cm,columnsep=0.75cm]{geometry}
% \usepackage{amsmath,amssymb}
% \usepackage{iftex}
% \usepackage{graphicx}
% \usepackage{float}
% \usepackage{subcaption}
% \usepackage{makecell}
% \usepackage{booktabs}
% \usepackage{multirow}
% \ifPDFTeX
%   \usepackage[T1]{fontenc}
%   \usepackage[utf8]{inputenc}
%   \usepackage{textcomp} % provide euro and other symbols
% \else % if luatex or xetex
%   \usepackage{unicode-math} % this also loads fontspec
%   \defaultfontfeatures{Scale=MatchLowercase}
%   \defaultfontfeatures[\rmfamily]{Ligatures=TeX,Scale=1}
% \fi
% \usepackage{lmodern}
% \ifPDFTeX\else
%   % xetex/luatex font selection
% \fi
% % Use upquote if available, for straight quotes in verbatim environments
% \IfFileExists{upquote.sty}{\usepackage{upquote}}{}
% \IfFileExists{microtype.sty}{% use microtype if available
%   \usepackage[]{microtype}
%   \UseMicrotypeSet[protrusion]{basicmath} % disable protrusion for tt fonts
% }{}
% \makeatletter
% \@ifundefined{KOMAClassName}{% if non-KOMA class
%   \IfFileExists{parskip.sty}{%
%     \usepackage{parskip}
%   }{% else
%     \setlength{\parindent}{0pt}
%     \setlength{\parskip}{6pt plus 2pt minus 1pt}}
% }{% if KOMA class
%   \KOMAoptions{parskip=half}}
% \makeatother
% \usepackage{longtable,booktabs,array}
\newcounter{none} % for unnumbered tables
% \usepackage{calc} % for calculating minipage widths
% % Correct order of tables after \paragraph or \subparagraph
% \usepackage{etoolbox}
% \makeatletter
% \patchcmd\longtable{\par}{\if@noskipsec\mbox{}\fi\par}{}{}
% \makeatother
% % Allow footnotes in longtable head/foot
% \IfFileExists{footnotehyper.sty}{\usepackage{footnotehyper}}{\usepackage{footnote}}
% \makesavenoteenv{longtable}
% \setlength{\emergencystretch}{3em} % prevent overfull lines
% \providecommand{\tightlist}{%
%   \setlength{\itemsep}{0pt}\setlength{\parskip}{0pt}}
% \usepackage{bookmark}
% \IfFileExists{xurl.sty}{\usepackage{xurl}}{} % add URL line breaks if available
% \urlstyle{same}
% \hypersetup{
%   hidelinks,
%   pdfcreator={LaTeX via pandoc}}

% \author{}
% \date{}

\begin{center}
    {\LARGE\bfseries\itshape
    BrainRotViT: Transformer-ResNet Hybrid for Explainable Modeling of Brain Aging from 3D sMRI
    \par}
    \vspace{0.5em}
    {\Large\bfseries
    Supplementary Materials
    \par}
\end{center}

\vspace{1em}

\section{ADNI Age-Gap Analyses}\label{adni-age-gap-analyses}

\subsection{Age-Bias Analysis}

Upon analysis of performance metrics stratified by age, sex, and site, it is observed that although the overall MAE is 3.52 (Pearson 0.79), precision is strongly age-dependent. The model appears to over-predict the young and under-predict the old, which is a common regression-to-the-mean bias. The residual-vs-age slope of $-0.38$/yr (corr $-0.62$) quantifies it; Bland--Altman bias is $\approx 0$ but limits of agreement are wide ($\pm 8.8$ yr). MAE F 3.30 $<$ M 3.78; CN 3.09 $<$ MCI 4.23 $\approx$ AD 4.17; site MAE ranges 1.48--4.87.\\
%------------------------------------------------------------
% Stratified performance (2x2 panel)
%------------------------------------------------------------
\begin{figure}[h]
    \centering
    \includegraphics[width=0.92\linewidth]{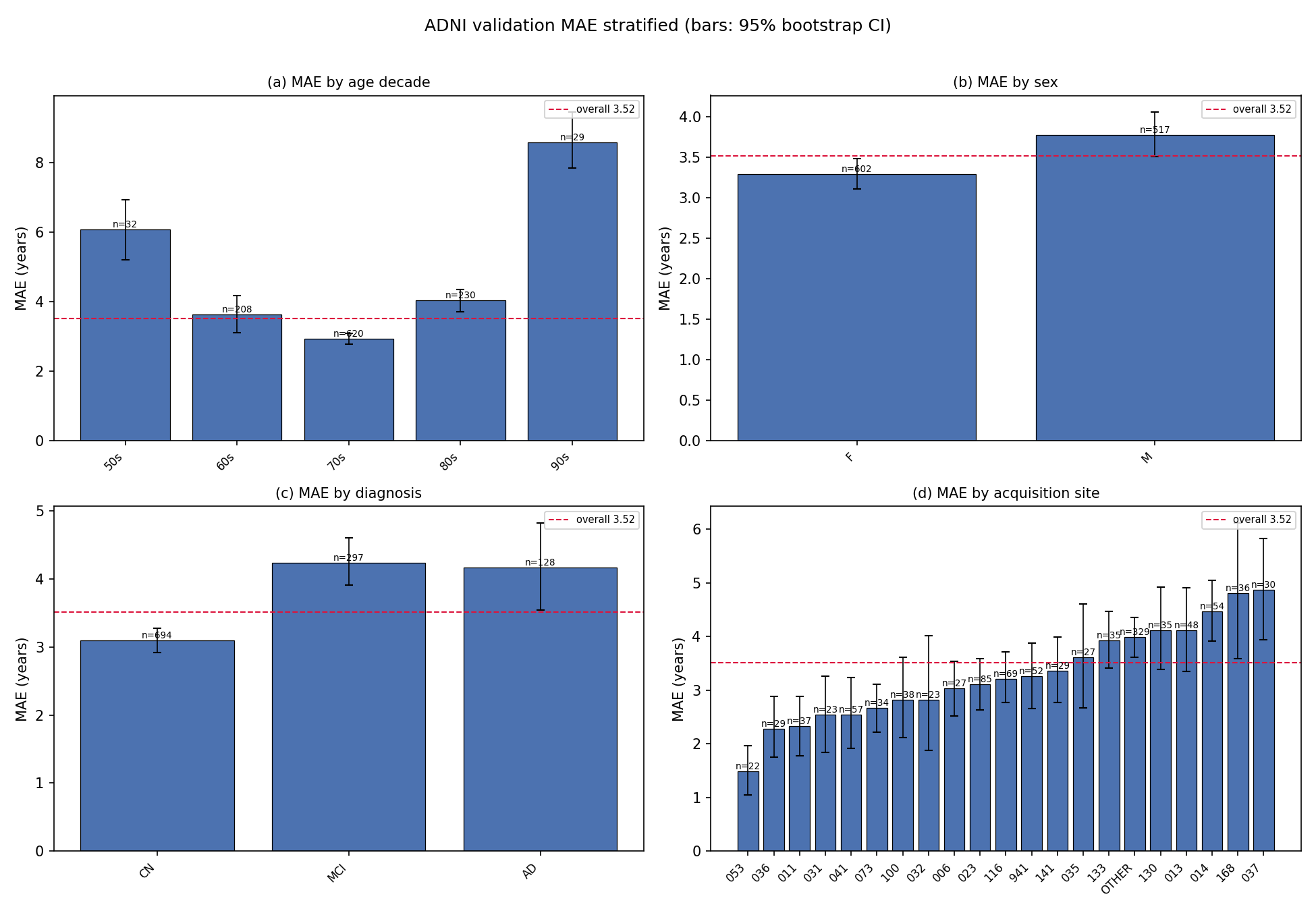}
    \caption{Validation performance stratified by demographic and clinical subgroups. Bars show MAE with 95\% bootstrap confidence intervals for (A) age decade, (B) sex, (C) diagnosis, and (D) acquisition site. Prediction error is lowest for cognitively normal participants and increases in the youngest and oldest age groups, consistent with regression-to-the-mean effects. Sites with fewer than 10 validation subjects were pooled into ``OTHER''.}
    \label{fig:adni_val_mae_stratified}
\end{figure}

%------------------------------------------------------------
% Bland-Altman
%------------------------------------------------------------
\begin{figure}[h]
    \centering
    \includegraphics[width=0.8\linewidth]{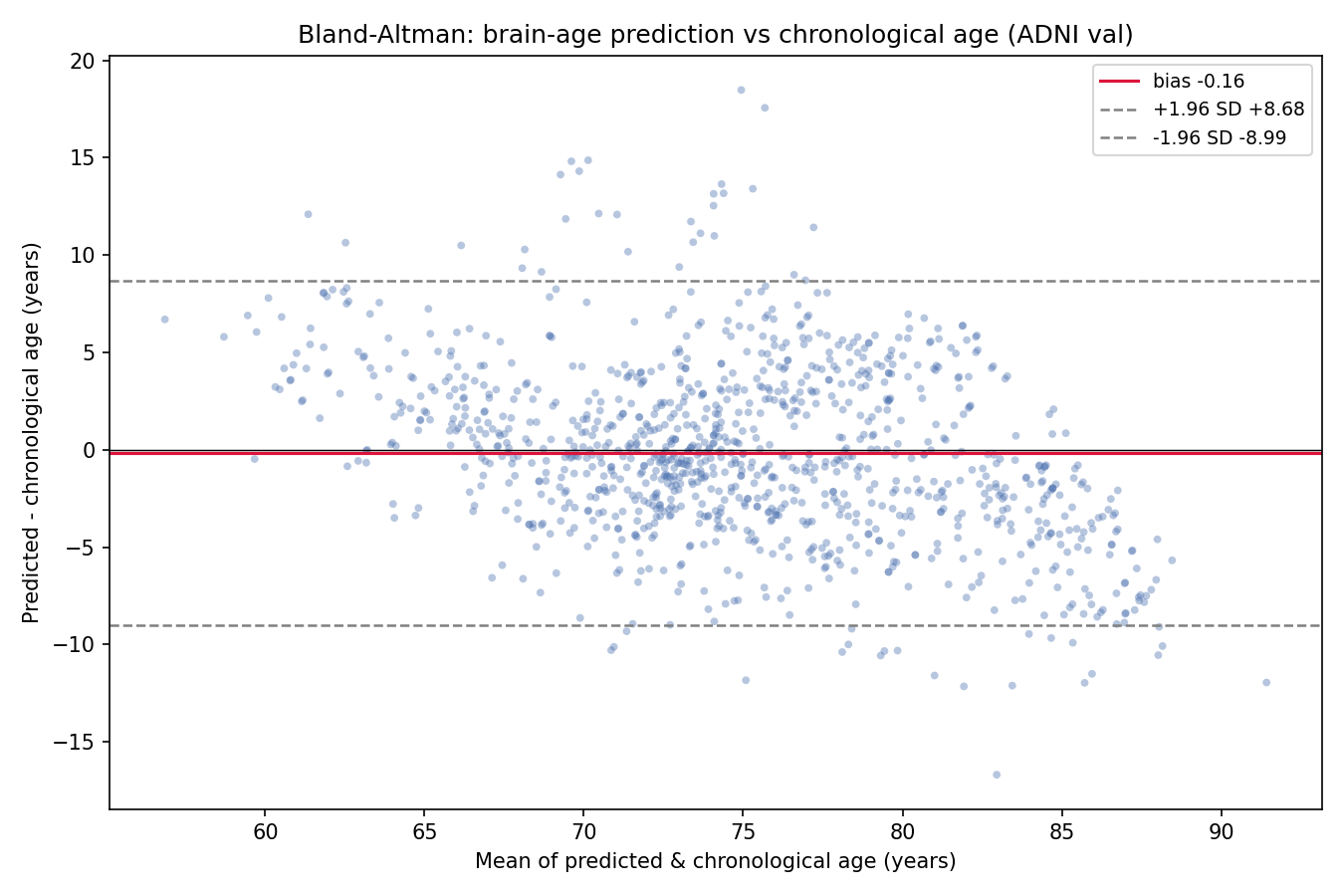}
    \caption{Bland--Altman analysis of predicted versus chronological age on the ADNI validation cohort. The mean signed bias is close to zero, indicating little overall systematic error, while the 95\% limits of agreement (approximately $\pm8.8$ years) illustrate the spread of individual prediction errors.}
    \label{fig:adni_val_bland_altman}
\end{figure}

%------------------------------------------------------------
% Residual vs age
%------------------------------------------------------------
\begin{figure}[H]
    \centering
    \includegraphics[width=0.8\linewidth]{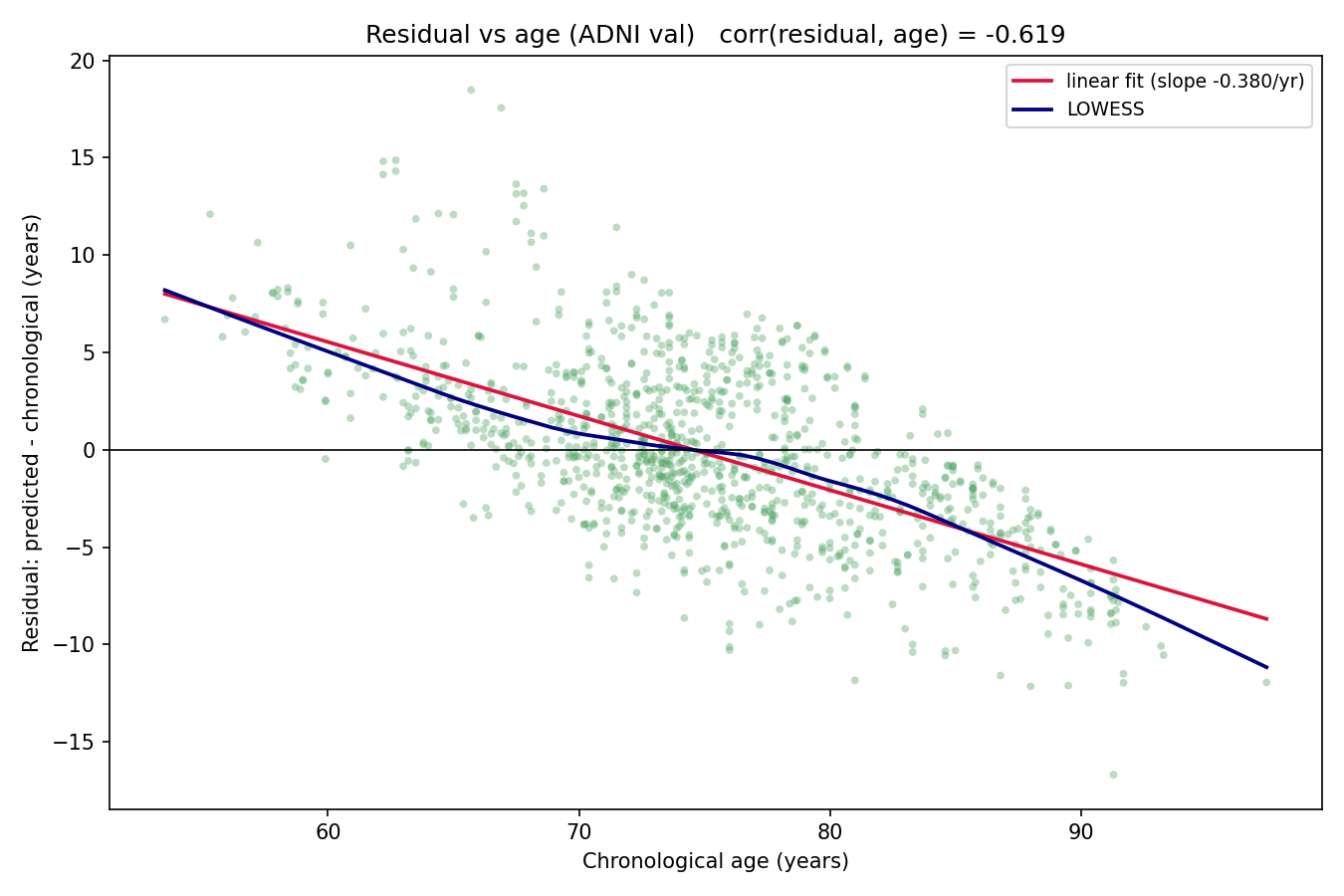}
    \caption{Residual age (predicted age $-$ chronological age) plotted against chronological age. A negative slope ($-0.38$ years/year; correlation $r=-0.62$) indicates regression-to-the-mean behavior, with overestimation in younger participants and underestimation in older participants. This age-dependent bias motivates the age-adjusted analyses reported in the main text.}
    \label{fig:adni_val_residual_vs_age}
\end{figure}

\begin{table}[h]
\centering
\small
\begin{tabular}{llrrrrr}
\toprule
Stratum Type & Stratum & $n$ & MAE (95\% CI) & RMSE & Bias \\
\midrule
Overall & All & 1119 & 3.517 (3.356--3.683) & 4.509 & -0.158 \\
\midrule
Age decade & 50s & 32 & 6.085 (5.193--6.927) & 6.561 & 6.056 \\
Age decade & 60s & 208 & 3.629 (3.108--4.164) & 5.220 & 3.098 \\
Age decade & 70s & 620 & 2.921 (2.762--3.090) & 3.623 & 0.123 \\
Age decade & 80s & 230 & 4.027 (3.699--4.353) & 4.787 & -3.664 \\
Age decade & 90s & 29 & 8.587 (7.839--9.462) & 8.891 & -8.587 \\
\midrule
Sex & F & 602 & 3.296 (3.109--3.485) & 4.075 & -0.268 \\
Sex & M & 517 & 3.776 (3.506--4.062) & 4.966 & -0.030 \\
\midrule
Diagnosis & AD & 128 & 4.165 (3.544--4.825) & 5.663 & 2.688 \\
Diagnosis & CN & 694 & 3.092 (2.922--3.274) & 3.881 & -1.001 \\
Diagnosis & MCI & 297 & 4.233 (3.909--4.607) & 5.251 & 0.583 \\
\midrule
Site & 006 & 27 & 3.034 (2.524--3.544) & 3.326 & -1.179 \\
Site & 011 & 37 & 2.333 (1.778--2.884) & 2.948 & -0.974 \\
Site & 013 & 48 & 4.122 (3.352--4.905) & 4.933 & -0.980 \\
Site & 014 & 54 & 4.470 (3.914--5.052) & 4.946 & -0.933 \\
Site & 023 & 85 & 3.114 (2.638--3.588) & 3.881 & 1.742 \\
Site & 031 & 23 & 2.542 (1.834--3.265) & 3.114 & -1.564 \\
Site & 032 & 23 & 2.827 (1.873--4.019) & 3.906 & 0.671 \\
Site & 035 & 27 & 3.607 (2.675--4.607) & 4.463 & -2.129 \\
Site & 036 & 29 & 2.275 (1.746--2.882) & 2.787 & -0.983 \\
Site & 037 & 30 & 4.871 (3.939--5.831) & 5.567 & -1.749 \\
Site & 041 & 57 & 2.543 (1.915--3.235) & 3.641 & 1.280 \\
Site & 053 & 22 & 1.483 (1.051--1.960) & 1.845 & 0.971 \\
Site & 073 & 34 & 2.672 (2.215--3.114) & 2.992 & -1.582 \\
Site & 100 & 38 & 2.826 (2.121--3.608) & 3.698 & -1.413 \\
Site & 116 & 69 & 3.217 (2.766--3.708) & 3.764 & -0.950 \\
Site & 130 & 35 & 4.114 (3.382--4.915) & 4.740 & -1.652 \\
Site & 133 & 35 & 3.927 (3.410--4.473) & 4.237 & 2.655 \\
Site & 141 & 29 & 3.360 (2.776--3.996) & 3.765 & 1.489 \\
Site & 168 & 36 & 4.804 (3.584--6.120) & 6.201 & -2.126 \\
Site & 941 & 52 & 3.259 (2.660--3.880) & 3.968 & 0.228 \\
Site & OTHER & 329 & 3.987 (3.617--4.361) & 5.354 & 0.195 \\
\bottomrule
\end{tabular}
\caption{Stratified validation performance on the ADNI cohort. Mean absolute error (MAE) with 95\% bootstrap confidence intervals, root mean squared error (RMSE), and signed bias are reported for the overall cohort and stratified by age decade, sex, diagnosis, and acquisition site. Sites with fewer than 10 validation subjects were pooled into ``OTHER'' to ensure stable estimates.}
\label{tab:stratified_performance}
\end{table}

\vspace{1em}

\subsection{Age-Bias Correction for BAG}

{\def\LTcaptype{none} % do not increment counter
\begin{longtable}[]{@{}
  >{\raggedright\arraybackslash}p{(\linewidth - 6\tabcolsep) * \real{0.2500}}
  >{\raggedright\arraybackslash}p{(\linewidth - 6\tabcolsep) * \real{0.2500}}
  >{\raggedright\arraybackslash}p{(\linewidth - 6\tabcolsep) * \real{0.2500}}
  >{\raggedright\arraybackslash}p{(\linewidth - 6\tabcolsep) * \real{0.2500}}@{}}
\toprule\noalign{}
\begin{minipage}[b]{\linewidth}\raggedright
Analysis
\end{minipage} & \begin{minipage}[b]{\linewidth}\raggedright
Uncorrected (crude)
\end{minipage} & \begin{minipage}[b]{\linewidth}\raggedright
(a) Corrected BAG
\end{minipage} & \begin{minipage}[b]{\linewidth}\raggedright
(b) Corrected + adjusted
\end{minipage} \\
\midrule\noalign{}
\endhead
\bottomrule\noalign{}
\endlastfoot
AD/MCI, High vs rest (threshold OR) & 5.49 {[}3.74--8.06{]} & 4.17
{[}3.04--5.74{]}, $p=9\times10^{-20}$ & \emph{(secondary)} 4.17 \\
AD/MCI, High vs Low (threshold OR) & 6.59 {[}4.03--10.77{]} & 6.96
{[}4.04--11.99{]}, $p=3\times10^{-13}$ & \emph{(secondary)} 6.96 \\
AD/MCI \textasciitilde{} BAG, adjusted (age+sex+site) & --- & --- & OR
1.72 per +1 SD {[}1.43--2.07{]}, $p=9\times10^{-9}$ (1.16/yr) \\
AD \textasciitilde{} BAG, adjusted (age+sex+site) & --- & --- & OR 2.28
per +1 SD {[}1.73--3.00{]}, $p=5\times10^{-9}$ \\
\end{longtable}
}

Age-bias correction attenuated the High-vs-rest OR (5.49→4.17) --- as
expected, some raw signal was age confound --- while the extreme-group
contrast held. The adjusted per-SD OR (1.72)  survives adjustment for age, sex, and site and stays highly significant. Sites (60 in the val set) with \textless10 subjects were pooled
into ``OTHER'' (→36 levels) so the categorical term is estimable; each
model also reports an age+sex-only version as a robustness check.\\

\section{Interpretability Pipeline Implementation Details}

\subsection{Registration and Attention Map Alignment}
\label{supp:reg-align}

\begin{figure}[H]
\centering
\begin{subfigure}[b]{0.48\textwidth}
\centering
\includegraphics[width=\textwidth]{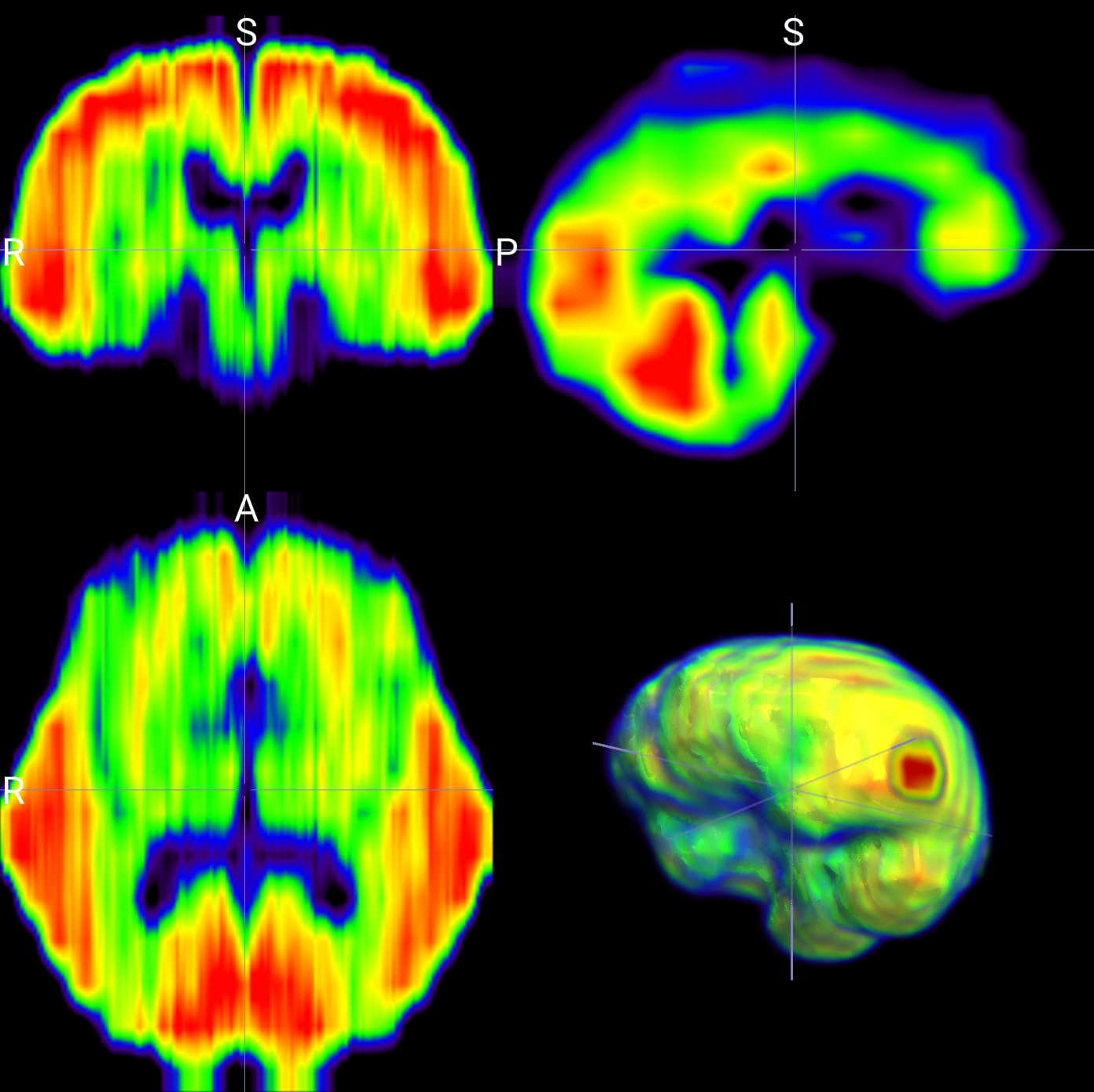}
\caption{Intensity plot of the raw 3D attention map generated by our interpretability pipeline.}
\label{fig:supp-attention-intensity}
\end{subfigure}
\hfill
\begin{subfigure}[b]{0.48\textwidth}
\centering
\includegraphics[width=\textwidth]{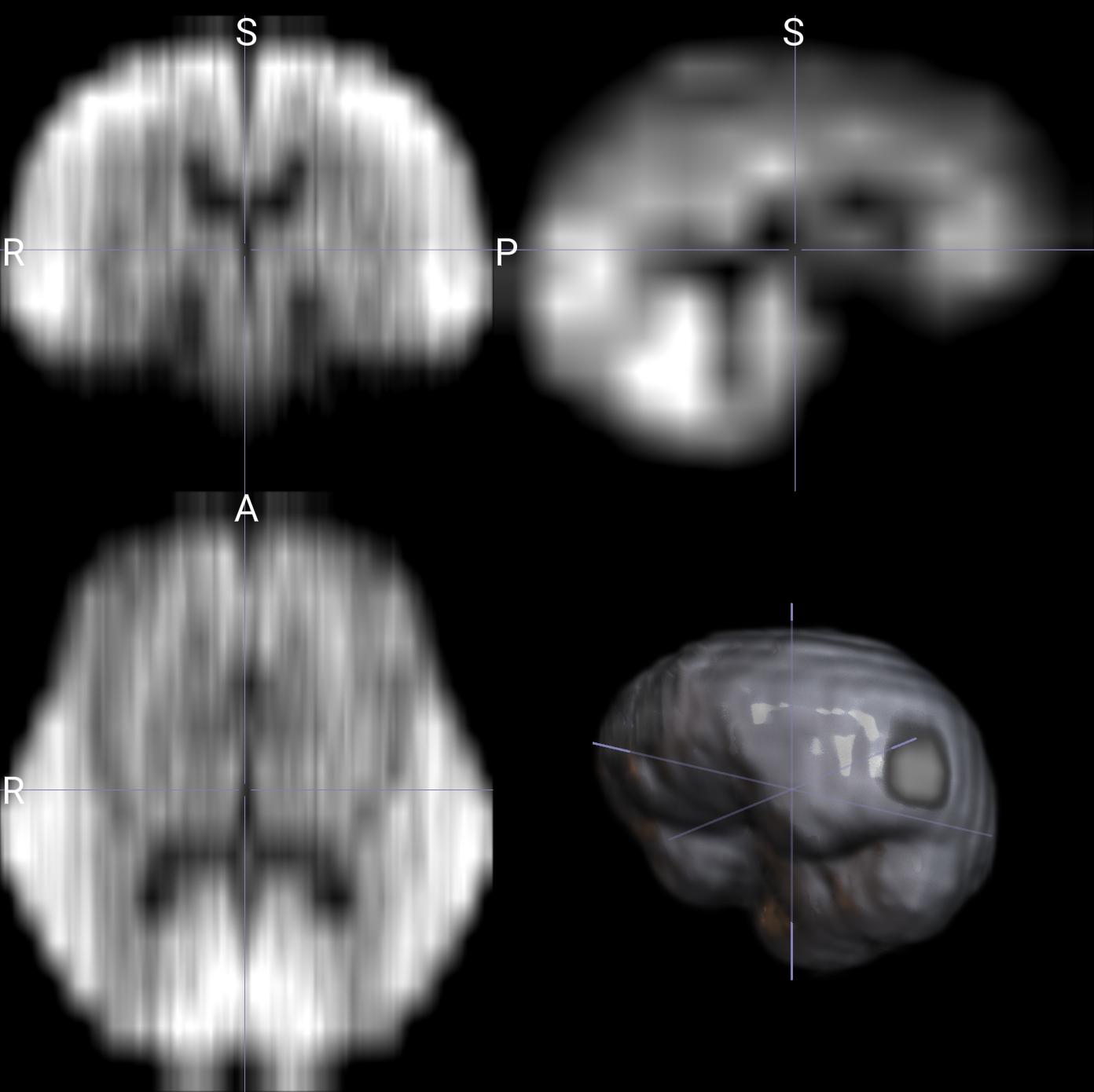}
\caption{3D attention map in grayscale (the deepest black meaning intensity=0 and the brightest white meaning intensity=1).}
\label{fig:supp-attention-grayscale}
\end{subfigure}

\vspace{0.5em}

\begin{subfigure}[b]{0.48\textwidth}
\centering
\includegraphics[width=\textwidth]{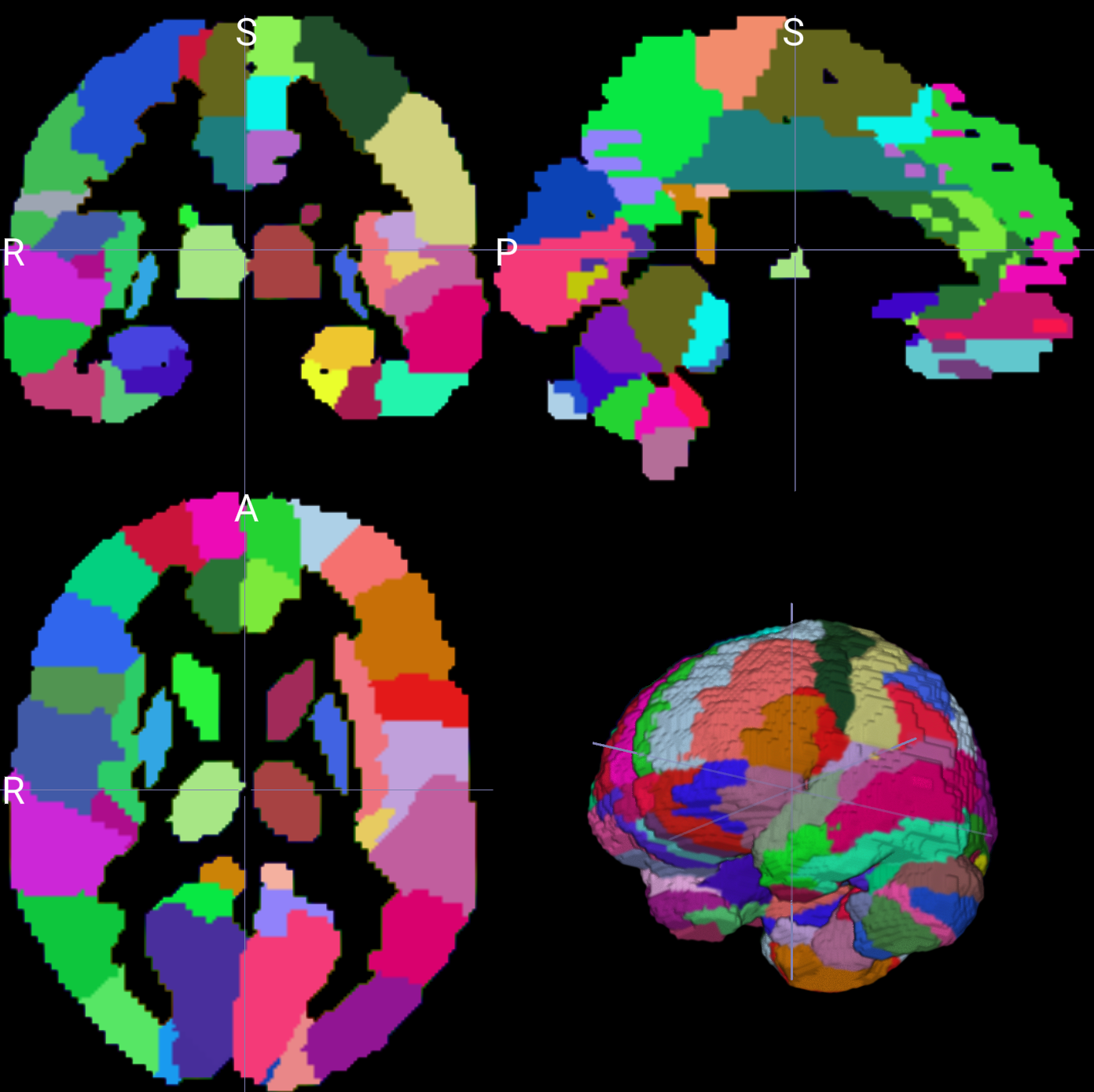}
\caption{NIfTI image of the AAL3 atlas, with each region's volume assigned a distinct solid color.}
\label{fig:supp-aal3-colored}
\end{subfigure}
\hfill
\begin{subfigure}[b]{0.48\textwidth}
\centering
\includegraphics[width=\textwidth]{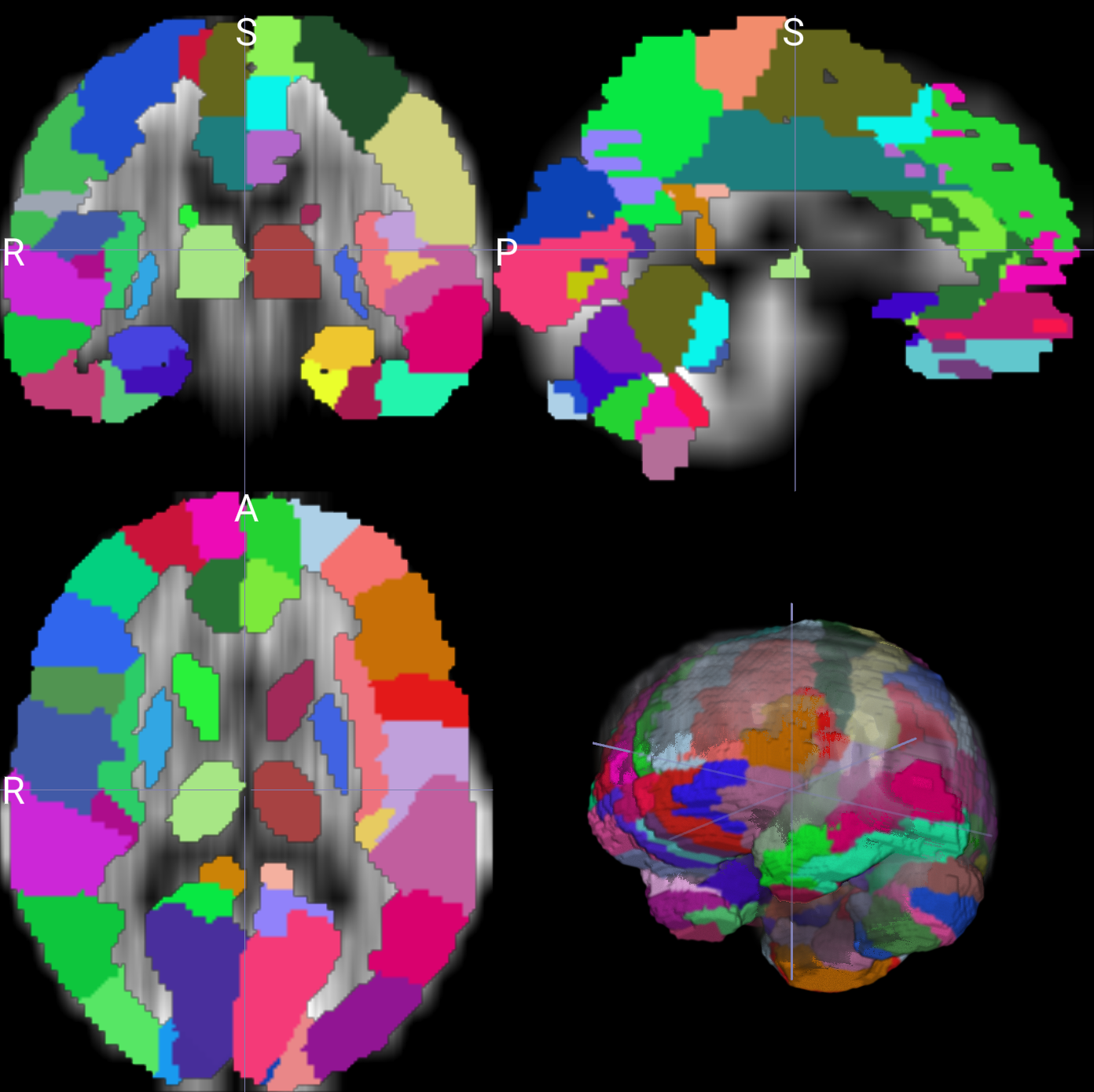}
\caption{AAL3 atlas aligned with the attention volume produced by our interpretability pipeline.}
\label{fig:supp-aal3-aligned}
\end{subfigure}

\caption{Alignment and registration of the 3D attention volume to the AAL3 atlas. (a) and (b) show the raw attention volume prior to registration; (c) shows the reference AAL3 atlas; (d) shows the final aligned overlay used to compute per-region attention scores. All the images were rendered using the open-source NIfTI viewing tool, MRIcroGL}
\label{fig:supp-alignment}
\end{figure}

To validate the registration of the 3D attention volume to the AAL3 atlas, we visually inspect the correspondence between the two volumes at each stage of the alignment pipeline. Figure~\ref{fig:supp-alignment} shows the raw attention intensity distribution, the corresponding grayscale attention volume, the AAL3 atlas with each anatomical region rendered in a distinct solid color, and the final overlay of the aligned attention volume on the atlas. Visual inspection of the overlay confirms that major anatomical landmarks in the attention volume correspond closely to their counterparts in the AAL3 atlas, supporting the validity of the affine registration procedure described in the main text.\\

\subsection{Ranked Regions of Interest}
\label{supp:ranked-roi}

Table~\ref{tab:supp-roi-ranked} lists the complete set of grouped ROIs derived from the AAL3 atlas, ranked in descending order by the maximum attention intensity observed within each region following alignment. Total voxel counts are also reported for each region to provide context on regional size. As noted in the main text, these scores are coarse regional indicators of attention rather than voxel-precise localisation, given the approximate nature of the slice extraction, resampling, and affine alignment steps involved in registration. We consulted the existing literature to contextualize the biological relevance of the most highly attended regions; the corresponding discussion is provided in the main text.\\

\begin{table}[H]
\centering
\footnotesize
\setlength{\tabcolsep}{6pt}
\renewcommand{\arraystretch}{1.1}
\begin{tabular}{clcc}
\hline
\textbf{Rank} & \textbf{Region} & \textbf{Max Intensity} & \textbf{Total Voxels} \\
\hline
1  & Calcarine        & 0.999609 & 32921 \\
2  & Precentral       & 0.999533 & 27058 \\
3  & Occipital        & 0.999030 & 79832 \\
4  & Cuneus           & 0.998887 & 40388 \\
5  & Temporal         & 0.998687 & 211174 \\
6  & Fusiform         & 0.998079 & 69355 \\
7  & Frontal          & 0.997673 & 323096 \\
8  & Cerebellum       & 0.996644 & 156456 \\
9  & Postcentral      & 0.995959 & 24733 \\
10 & Supplementary Motor Area & 0.995709 & 38429 \\
11 & Lingual          & 0.994943 & 18450 \\
12 & Paracentral      & 0.994694 & 30155 \\
13 & Heschl           & 0.991207 & 22047 \\
14 & Vermis           & 0.988397 & 16251 \\
15 & Parietal         & 0.972327 & 74190 \\
16 & Angular          & 0.962445 & 51541 \\
17 & Rolandic         & 0.959995 & 10733 \\
18 & Precuneus        & 0.906631 & 26083 \\
19 & Cingulum         & 0.886694 & 57234 \\
20 & SupraMarginal    & 0.873127 & 15770 \\
21 & Insula           & 0.862784 & 40442 \\
22 & Caudate          & 0.709852 & 15883 \\
23 & ParaHippocampal  & 0.695122 & 18652 \\
24 & Putamen          & 0.684233 & 8510  \\
25 & Thalamus         & 0.676635 & 8399  \\
26 & Pallidum         & 0.663594 & 13173 \\
27 & Hippocampus      & 0.597772 & 7606  \\
28 & Amygdala         & 0.525721 & 1965  \\
29 & Olfactory        & 0.505724 & 2286  \\
30 & Rectus           & 0.367310 & 5930  \\
\hline
\end{tabular}
\caption{Ranked ROIs by maximum attention intensity from the aligned attention volume overlapped with the AAL3 atlas.}
\label{tab:supp-roi-ranked}
\end{table}

\subsection{Saliency Sanity Checks and Stability Analysis}
\label{supp:saliency-sanity-check}

To support the sanity-check and stability results reported in the main text, this section provides the corresponding figures from the model-parameter randomization analysis and the inter-subject and split-half stability evaluations, computed over $N=100$ held-out validation subjects.

\begin{figure}[h]
\centering
\includegraphics[width=0.9\textwidth]{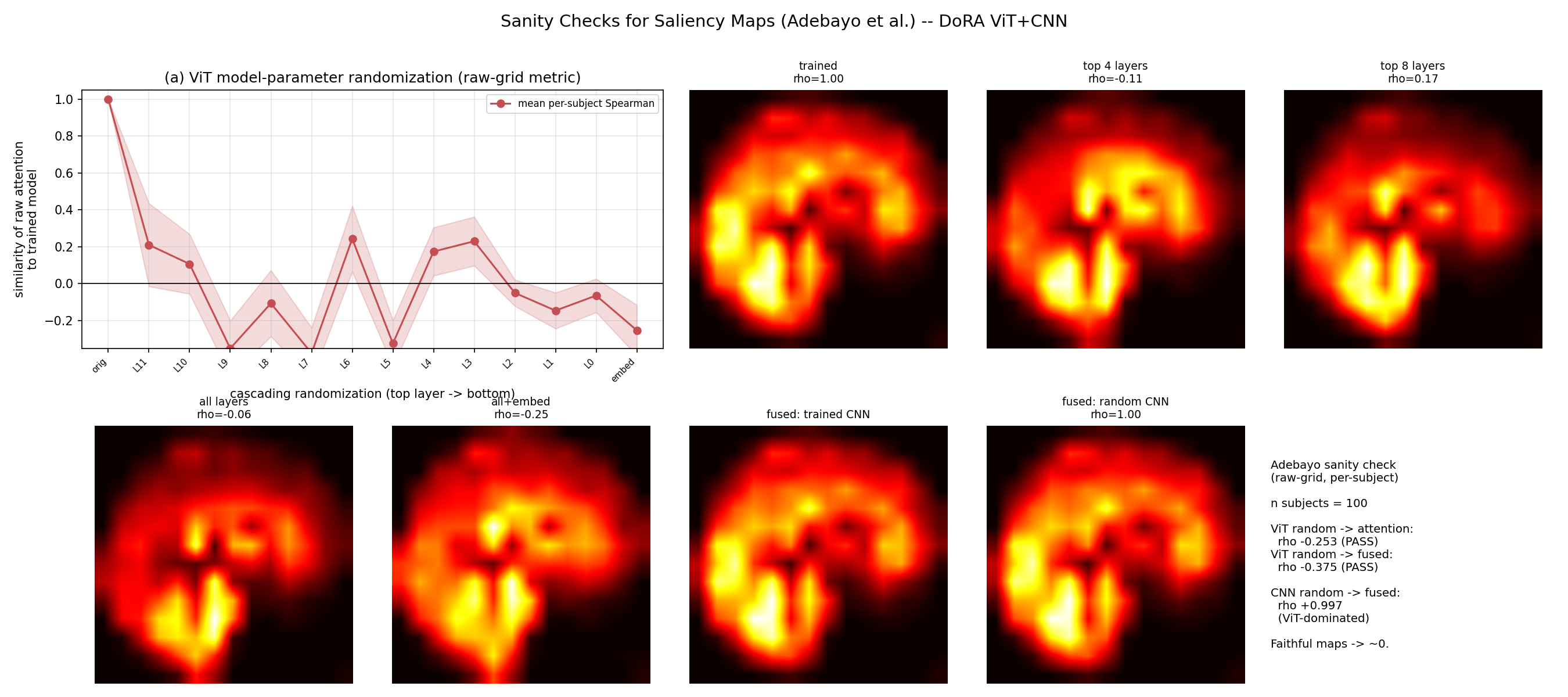}
\caption{Model-parameter randomization sanity check. The cascade curve shows the Spearman correlation between the original and randomized-model attention maps as ViT layers are progressively randomized from the top layer down to full randomization, alongside representative attention maps at each stage. A comparison with CNN-only randomization is also shown, illustrating that the fused saliency map is dominated by the ViT attention component.}
\label{fig:supp-sanity-randomization}
\end{figure}

Figure~\ref{fig:supp-sanity-randomization} shows the cascade of Spearman correlations between the trained model's attention map and attention maps generated after progressively randomizing the ViT backbone's weights, alongside example attention maps at increasing levels of randomization and a comparison against CNN-only randomization.

\begin{figure}[h]
\centering
\includegraphics[width=0.9\textwidth]{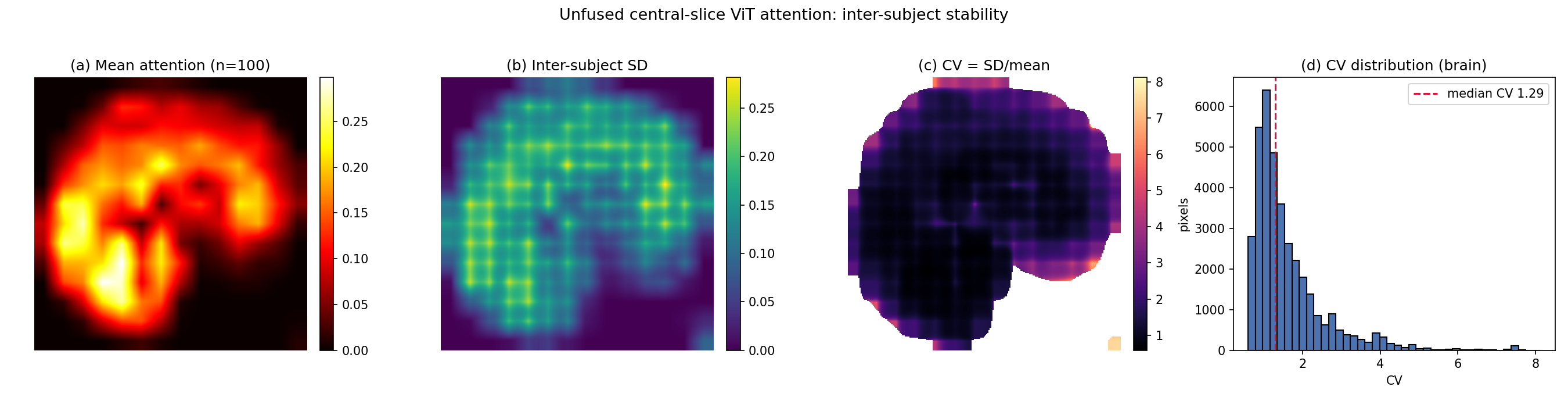}
\caption{Inter-subject variability of attention maps. Panels show the mean, standard deviation, and coefficient of variation (CV) computed across per-subject unfused central-slice attention maps, along with a histogram of voxel-level CV values across the validation cohort.}
\label{fig:supp-stability-intersubject}
\end{figure}

Figure~\ref{fig:supp-stability-intersubject} shows the mean, standard deviation, and coefficient of variation (CV) across per-subject unfused central-slice attention maps, along with the distribution of voxel-level CV values.

\begin{figure}[h]
\centering
\includegraphics[width=0.9\textwidth]{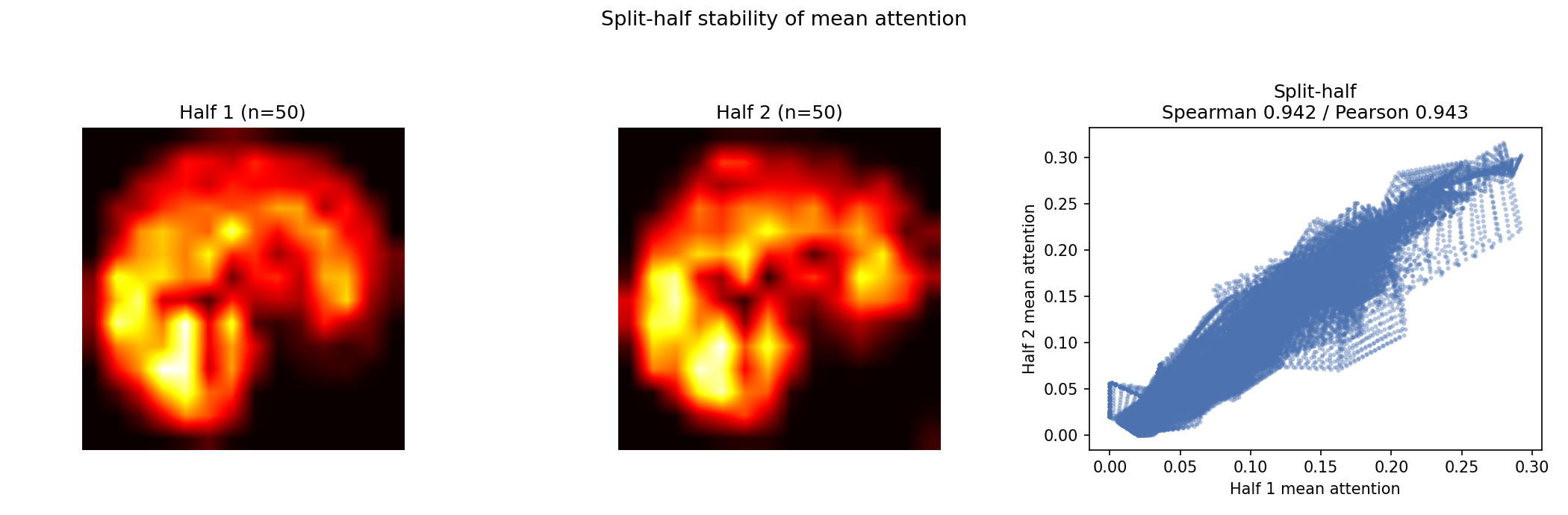}
\caption{Split-half stability analysis. The two panels show mean attention maps computed independently from two randomly partitioned halves of the validation cohort ($n=50$ each), with the corresponding voxel-wise scatter plot showing agreement between the two halves (Spearman $\rho = 0.942$, Pearson $r = 0.943$).}
\label{fig:supp-stability-splithalf}
\end{figure}

Figure~\ref{fig:supp-stability-splithalf} shows the split-half reliability analysis, comparing mean attention maps computed from two independent, randomly partitioned halves of the validation cohort.

\section{Implementation Details of Competing Methods}
\label{app:baselines}

This appendix documents the implementation, hyperparameters, and model-selection
protocol for every method in the state-of-the-art comparison, addressing the need for reproducible baseline
configurations. We distinguish two categories of competing method. The four
published architectures---SFCN, the Global--Local
Transformer, Triamese-ViT, and
TSAN---are \emph{reference implementations} evaluated at the
settings recommended by their respective authors; we did not tune them, so their
tuning budget is nil. The two remaining models, 3D-ResNet and 3D-ViT, are
\emph{baselines we implemented ourselves}; the selection criteria and tuning
budgets in Section~\ref{app:selection} apply to these two models only.

\subsection{Common Training Protocol}
\label{app:common}
All methods share a common experimental protocol to ensure a fair comparison.
Samples are partitioned into an $80\%$ training and $20\%$ validation split using
a fixed random seed ($69420$), yielding identical held-out samples across
methods. Every run retains the checkpoint with the lowest validation mean
absolute error (MAE), and all reported metrics are computed from this
best-validation checkpoint. All models were trained on a single
NVIDIA RTX~3060 (12\,GB) GPU; the available memory constrains the batch sizes
reported in Table~\ref{tab:baseline_hparams}. Volumes are skull-stripped,
bias-corrected, brain-cropped, and resampled prior to training; the volumetric
models (3D-ResNet, 3D-ViT, Triamese-ViT) additionally apply foreground intensity
standardisation to a $160\times192\times160$ volume, SFCN applies
mean-normalisation to the same grid, the Global--Local Transformer operates on
two-dimensional slices of a $130\times170\times120$ volume, and TSAN uses its
native MNI-space preprocessing pipeline.

\subsection{Hyperparameters}
\label{app:hparams}
Table~\ref{tab:baseline_hparams} reports the architectural and optimisation
hyperparameters of each method as used in our experiments.

\begin{table}[h]
\centering
\caption{Architecture and optimisation hyperparameters for all competing
methods. The 3D-ResNet and 3D-ViT baselines were implemented and tuned in this
work; the remaining methods use their authors' reference configurations.}
\label{tab:baseline_hparams}
\small
\setlength{\tabcolsep}{4pt}
\resizebox{\textwidth}{!}{%
\begin{tabular}{@{}lllllll@{}}
\toprule
 & \textbf{3D-ResNet} & \textbf{3D-ViT} & \textbf{SFCN} & \textbf{Global--Local} & \textbf{Triamese-ViT} & \textbf{TSAN} \\
\midrule
Type        & Baseline & Baseline & Reference & Reference & Reference & Reference \\
Backbone    & ResNet-34 (3D) & ViViT tubelet & SFCN & VGG8 + transformer & MultiView ViT ($\times3$) & ScaleDense (2-stage) \\
Key config  & \makecell[l]{blocks [3,4,6,3];\\ch.\ 64--512} & \makecell[l]{patch $16^2$, depth-patch 16;\\dim 512, 8 layers,\\8 heads, MLP 2048} & \makecell[l]{6 conv stages\\{}[32--256,64];\\soft-label head} & \makecell[l]{patch 64, step 32;\\6 blocks, hidden 512,\\8 heads, drop 0.5} & \makecell[l]{emb 768, 10 layers,\\12 heads, MLP 3072;\\patch $8^2$, drop 0.1} & \makecell[l]{$n_f{=}8$, 5 blocks;\\scale attention;\\uses sex} \\
Input       & $160^{}{\times}192{\times}160$ & $160{\times}192{\times}160$ & $160{\times}192{\times}160$ & 2D slices & 3 orthogonal views & MNI vol.\ + sex \\
Loss        & MSE & MSE & \makecell[l]{soft-label KL\\(bin 1\,yr, $\sigma{=}1$)} & L1 (MAE) & L1 (MAE) & \makecell[l]{MSE + ranking\\($\lambda{=}10$, $\beta{=}1$)} \\
Optimiser   & Adam (amsgrad) & AdamW & SGD ($m{=}0.9$) & Adam & Adam (amsgrad) & Adam \\
Learning rate & $1{\times}10^{-3}$ & $1{\times}10^{-4}$ & $1{\times}10^{-2}$ & $1{\times}10^{-4}$ & $1{\times}10^{-4}$ & $10^{-3}\!/\,10^{-5}$ \\
Weight decay & $5{\times}10^{-4}$ & $1{\times}10^{-4}$ & $1{\times}10^{-3}$ & $0$ & $1{\times}10^{-5}$ & $5{\times}10^{-4}$ \\
LR schedule & \makecell[l]{Plateau\\(pat.\ 5, ${\times}0.5$)} & \makecell[l]{Cosine warm-restart\\($T_0{=}10$, $T_{\text{mult}}{=}2$)} & \makecell[l]{Step\\($T/3$, ${\times}0.3$)} & \makecell[l]{Step\\(25, ${\times}0.5$)} & \makecell[l]{Plateau\\(pat.\ 5, ${\times}0.5$)} & \makecell[l]{Plateau\\(val MAE)} \\
Batch size  & 4 & 4 & 4 & 4 & 2 & 8 \\
Epochs      & 100 & 100 & 30 & 80 & 100 & $50{+}50$ \\
\bottomrule
\end{tabular}%
}
\end{table}

\subsection{Model Selection and Tuning Budget (Baselines)}
\label{app:selection}
Model selection and hyperparameter tuning are reported for the two baselines
introduced in this work; the reference implementations were run at their
published settings without further tuning and are therefore excluded from this
subsection.

\paragraph{Selection criterion.}
For both baselines, the selection metric is the \emph{validation-set MAE}. Two
levels of selection apply. First, at the \emph{checkpoint} level, the reported
model is the training epoch attaining the lowest validation MAE, with ties broken
by validation loss and then by the earliest epoch. Second, at the
\emph{configuration} level, the final hyperparameters (Table~\ref{tab:baseline_hparams})
are those of the configuration minimising validation MAE across the search space
in Table~\ref{tab:baseline_tuning}. No test-set information is used at any point
of model selection.

\paragraph{Tuning budget.}
Each baseline was tuned with a small manual search over the ranges in
Table~\ref{tab:baseline_tuning}, comprising approximately ten configurations per
model trained under a single random seed. The search prioritised the factors to
which each architecture is most sensitive---network depth and learning rate for
3D-ResNet, and embedding dimension, transformer depth, and learning rate for
3D-ViT---while other settings were fixed at commonly used values. This budget was
chosen to be commensurate with the single-GPU compute available and is reported
in full for reproducibility.

\begin{table}[htbp]
\centering
\caption{Hyperparameter search space and tuning budget for the two baselines.
Selection minimises validation MAE.}
\label{tab:baseline_tuning}
\small
\begin{tabular}{@{}llll@{}}
\toprule
\textbf{Baseline} & \textbf{Search space} & \textbf{Budget} & \textbf{Selection metric} \\
\midrule
3D-ResNet & \makecell[l]{depth $\in\{$18, 34, 50$\}$;\\
lr $\in\{3{\times}10^{-4}, 1{\times}10^{-3}, 3{\times}10^{-3}\}$;\\
weight decay $\in\{1{\times}10^{-4}, 5{\times}10^{-4}\}$;\\
batch size $\in\{2, 4\}$}
& \makecell[l]{$\sim$10 configs,\\1 seed}
& \makecell[l]{validation MAE\\(best checkpoint)} \\
\addlinespace
3D-ViT & \makecell[l]{dim $\in\{384, 512\}$;\\
depth $\in\{6, 8, 12\}$;\\
lr $\in\{5{\times}10^{-5}, 1{\times}10^{-4}, 3{\times}10^{-4}\}$;\\
weight decay $\in\{1{\times}10^{-4}, 1{\times}10^{-2}\}$}
& \makecell[l]{$\sim$10 configs,\\1 seed}
& \makecell[l]{validation MAE\\(best checkpoint)} \\
\bottomrule
\end{tabular}
\end{table}

\section{Error-Based and Stratified Evaluation on the Pooled Validation Set}
\label{app:stratified}

Because Pearson correlation is inflated by a wide chronological age range and can
obscure clinically relevant error, we complement it with error-based,
calibration, and bias analyses on the held-out pooled mixed-cohort validation set
($n=1{,}022$; age $10$--$96$~years). In the pooled design each source dataset
corresponds to a distinct acquisition/scanner cohort, so the by-dataset breakdown
constitutes the scanner/site analysis.\\

\paragraph{Overall accuracy.}
The model attains an MAE of $3.43$~years ($95\%$ CI $[3.24, 3.63]$) and an RMSE of
$4.66$~years, with a small mean signed error of $-0.27$~years, despite a Pearson
correlation of $0.979$ and a Spearman correlation of $0.974$. The near-unity
correlation coexists with a multi-year absolute error, demonstrating that global
correlation over a wide age span overstates individual-level accuracy.

\begin{figure}[htbp]
\centering
\includegraphics[width=\textwidth]{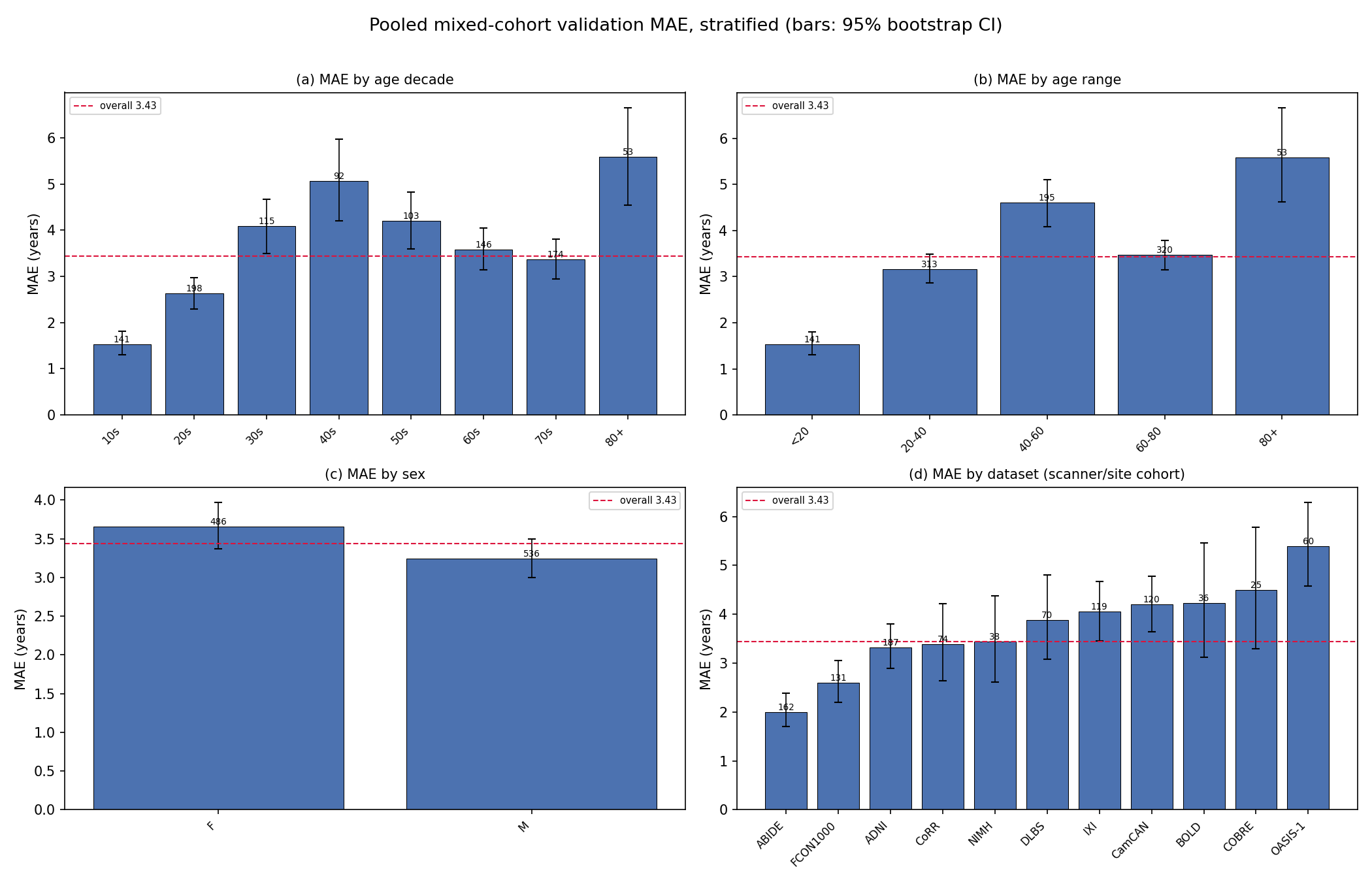}
\caption{Validation MAE stratified by (a) age decade, (b) age range, (c) sex, and
(d) dataset/scanner-site cohort. Error bars are $95\%$ bootstrap CIs; the dashed
line marks the pooled MAE and per-bar labels give the sample count.}
\label{fig:stratified_mae}
\end{figure}

\begin{figure}[htbp]
\centering
\includegraphics[width=\textwidth]{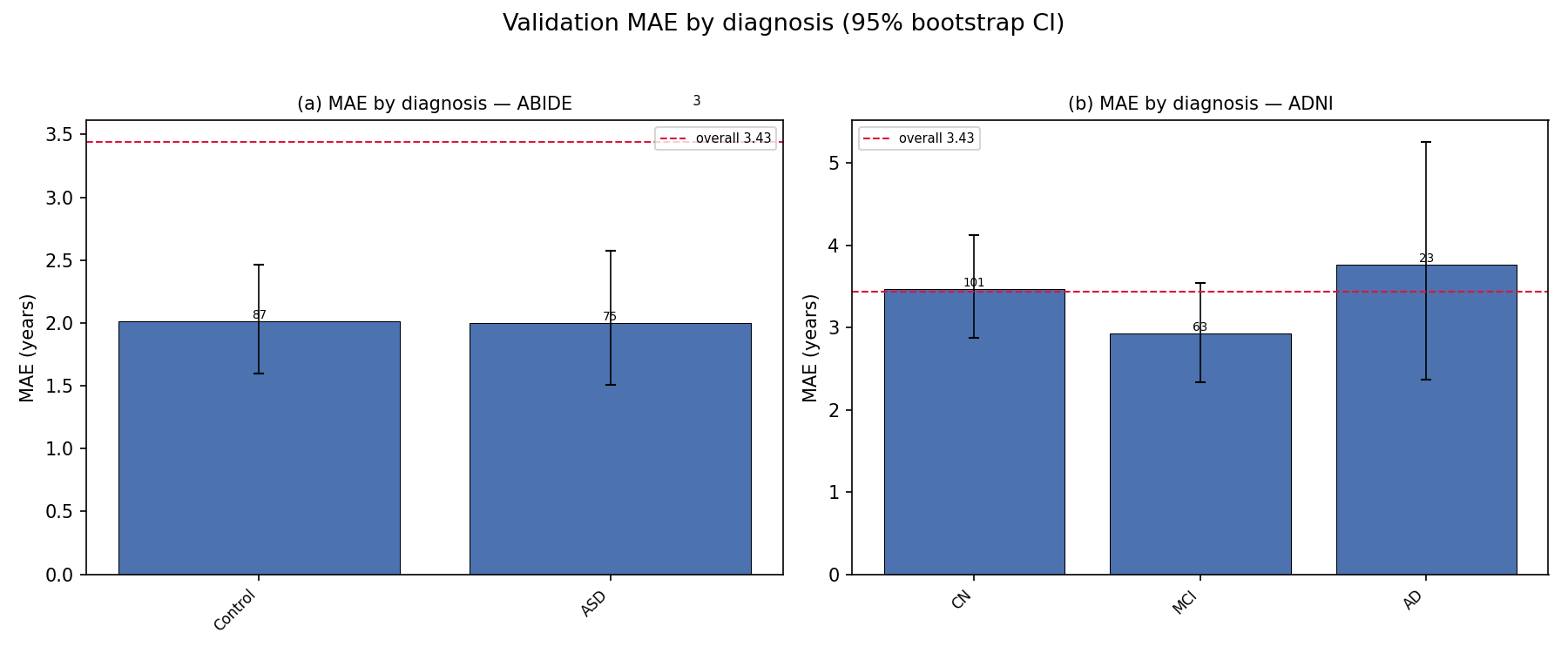}
\caption{Validation MAE by diagnosis for the two labelled cohorts: (a) ABIDE
(ASD vs.\ Control) and (b) ADNI (CN, MCI, AD). Error bars are $95\%$ bootstrap CIs;
the dashed line marks the pooled MAE.}
\label{fig:diagnosis_mae}
\end{figure}

\paragraph{Stratified MAE.}
Table~\ref{tab:stratified_mae} reports MAE (with $95\%$ bootstrap confidence
intervals), RMSE, signed bias, and correlations stratified by age decade
(with $90+$ merged into $80+$ because of small sample size), coarse age range, sex, and source dataset
(scanner/site cohort), together with a diagnosis breakdown for the two pooled
cohorts carrying clinical labels (ABIDE and ADNI). Figures~\ref{fig:stratified_mae}
and~\ref{fig:diagnosis_mae} show the corresponding bar charts. Error is strongly
age-dependent: MAE rises from $1.53$~years in adolescents ($<20$) to $4.61$~years
in mid-adulthood ($40$--$60$) and $5.59$~years at $80+$, with a pronounced
negative bias at older ages ($-5.15$~years).  Accuracy is comparable across sex (F $3.65$ vs.\ M $3.24$~years). Within ABIDE, MAE is essentially identical for ASD and controls ($2.00$ vs.\ $2.01$~years), indicating that prediction accuracy is not confounded by diagnosis when the model is trained on a large, diverse distribution. Within ADNI however,
MAE is $3.47$ (CN), $2.92$ (MCI), and $3.76$~years (AD); notably the AD group shows
a positive bias of $+2.18$~years (brains predicted older than chronological age),
consistent with the accelerated-ageing signal analysed in
Section~\ref{adni-age-gap-analyses}.

\begin{table}[htbp]
\centering
\caption{Validation-set accuracy stratified by age decade ($90+$ merged into
$80+$), age range, sex, dataset (scanner/site cohort), and diagnosis (ABIDE, ADNI).
MAE and its $95\%$ bootstrap CI are in years; bias is the mean signed error;
$r$ is Pearson and $\rho$ Spearman.}
\label{tab:stratified_mae}
\footnotesize
\setlength{\tabcolsep}{5pt}
\begin{tabular}{@{}llrrrrr@{}}
\toprule
\textbf{Stratum} & & $n$ & \textbf{MAE [95\% CI]} & \textbf{RMSE} & \textbf{bias} & $r/\rho$ \\
\midrule
\textbf{Overall} & all & 1022 & 3.43\,[3.24, 3.63] & 4.66 & $-0.27$ & 0.98/0.97 \\
\midrule
\multirow{8}{*}{\textit{Age decade}}
 & 10s & 141 & 1.53\,[1.30, 1.81] & 2.14 & $+0.37$ & 0.86/0.79 \\
 & 20s & 198 & 2.63\,[2.29, 2.97] & 3.58 & $+0.74$ & 0.64/0.66 \\
 & 30s & 115 & 4.09\,[3.49, 4.67] & 5.18 & $-0.26$ & 0.41/0.42 \\
 & 40s & 92  & 5.07\,[4.21, 5.97] & 6.61 & $-0.26$ & 0.41/0.42 \\
 & 50s & 103 & 4.20\,[3.60, 4.83] & 5.21 & $+0.20$ & 0.48/0.49 \\
 & 60s & 146 & 3.59\,[3.15, 4.05] & 4.61 & $+0.62$ & 0.44/0.45 \\
 & 70s & 174 & 3.38\,[2.95, 3.81] & 4.52 & $-1.50$ & 0.30/0.26 \\
 & 80+ & 53  & 5.59\,[4.55, 6.65] & 6.81 & $-5.15$ & 0.38/0.38 \\
\midrule
\multirow{5}{*}{\textit{Age range}}
 & $<$20  & 141 & 1.53\,[1.30, 1.80] & 2.14 & $+0.37$ & 0.86/0.79 \\
 & 20--40 & 313 & 3.17\,[2.86, 3.48] & 4.24 & $+0.38$ & 0.78/0.81 \\
 & 40--60 & 195 & 4.61\,[4.08, 5.11] & 5.91 & $-0.02$ & 0.72/0.73 \\
 & 60--80 & 320 & 3.47\,[3.14, 3.79] & 4.56 & $-0.53$ & 0.65/0.63 \\
 & 80+    & 53  & 5.59\,[4.62, 6.67] & 6.81 & $-5.15$ & 0.38/0.38 \\
\midrule
\multirow{2}{*}{\textit{Sex}}
 & F & 486 & 3.65\,[3.37, 3.97] & 4.93 & $+0.18$ & 0.98/0.97 \\
 & M & 536 & 3.24\,[3.00, 3.50] & 4.40 & $-0.68$ & 0.98/0.98 \\
\midrule
\multirow{11}{*}{\textit{Dataset (site)}}
 & ABIDE    & 162 & 2.00\,[1.71, 2.38] & 2.97 & $-0.27$ & 0.95/0.87 \\
 & FCON1000 & 131 & 2.60\,[2.19, 3.05] & 3.61 & $-0.02$ & 0.97/0.85 \\
 & ADNI     & 187 & 3.32\,[2.90, 3.80] & 4.46 & $-0.31$ & 0.62/0.54 \\
 & CoRR     & 74  & 3.39\,[2.65, 4.22] & 4.78 & $+0.70$ & 0.94/0.96 \\
 & NIMH     & 38  & 3.44\,[2.61, 4.37] & 4.44 & $+1.14$ & 0.89/0.85 \\
 & DLBS     & 70  & 3.89\,[3.09, 4.80] & 5.31 & $+0.39$ & 0.97/0.96 \\
 & IXI      & 119 & 4.05\,[3.46, 4.67] & 5.26 & $-0.47$ & 0.95/0.94 \\
 & CamCAN   & 120 & 4.20\,[3.64, 4.78] & 5.27 & $-0.94$ & 0.96/0.96 \\
 & BOLD     & 36  & 4.22\,[3.13, 5.46] & 5.55 & $-0.58$ & 0.97/0.93 \\
 & COBRE    & 25  & 4.49\,[3.29, 5.78] & 5.50 & $-1.99$ & 0.90/0.92 \\
 & OASIS-1  & 60  & 5.39\,[4.58, 6.28] & 6.35 & $-0.94$ & 0.95/0.89 \\
\midrule
\multirow{2}{*}{\shortstack[l]{\textit{Diagnosis}\\\textit{(ABIDE)}}}
 & ASD     & 75 & 2.00\,[1.51, 2.57] & 3.02 & $-0.25$ & 0.96/0.92 \\
 & Control & 87 & 2.01\,[1.60, 2.46] & 2.93 & $-0.29$ & 0.95/0.79 \\
\midrule
\multirow{3}{*}{\shortstack[l]{\textit{Diagnosis}\\\textit{(ADNI)}}}
 & CN  & 101 & 3.47\,[2.88, 4.12] & 4.67 & $-0.85$ & 0.65/0.60 \\
 & MCI & 63  & 2.92\,[2.34, 3.54] & 3.82 & $-0.37$ & 0.60/0.52 \\
 & AD  & 23  & 3.76\,[2.37, 5.25] & 5.12 & $+2.18$ & 0.45/0.34 \\
\bottomrule
\end{tabular}
\end{table}

\begin{figure}[htbp]
\centering
\includegraphics[width=\textwidth]{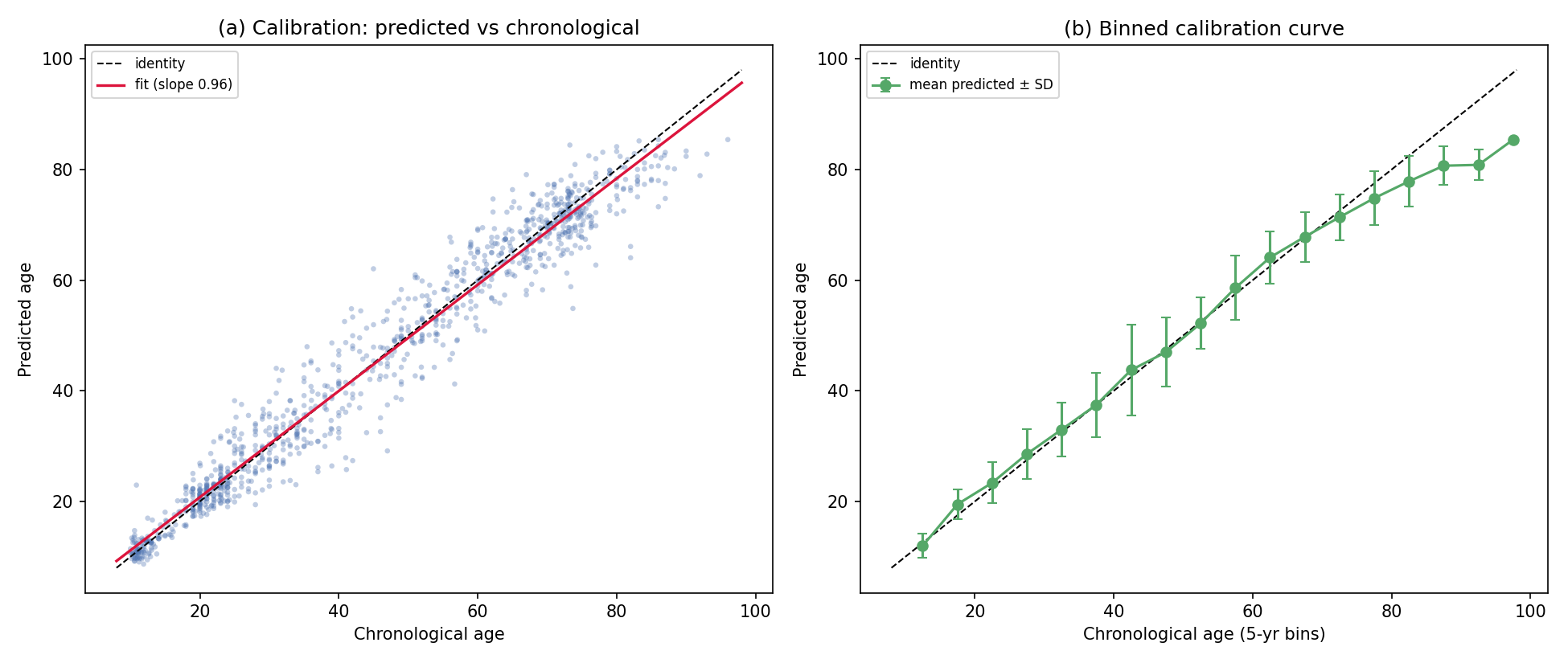}
\caption{(a) Predicted vs chronological age with identity and fitted line.
(b) Binned calibration curve (mean predicted age $\pm$ SD per $5$-year bin).}
\label{fig:calibration}
\end{figure}

\paragraph{Calibration.}
Figure~\ref{fig:calibration} shows predicted versus chronological age with the
identity line and a linear fit, together with a binned calibration curve (mean
predicted age $\pm$ SD in $5$-year windows). The fitted slope below one reflects
regression-to-the-mean, with mild over-prediction of younger subjects and
progressive under-prediction beyond $\sim$70~years.

\begin{figure}[htbp]
\centering
\includegraphics[width=\textwidth]{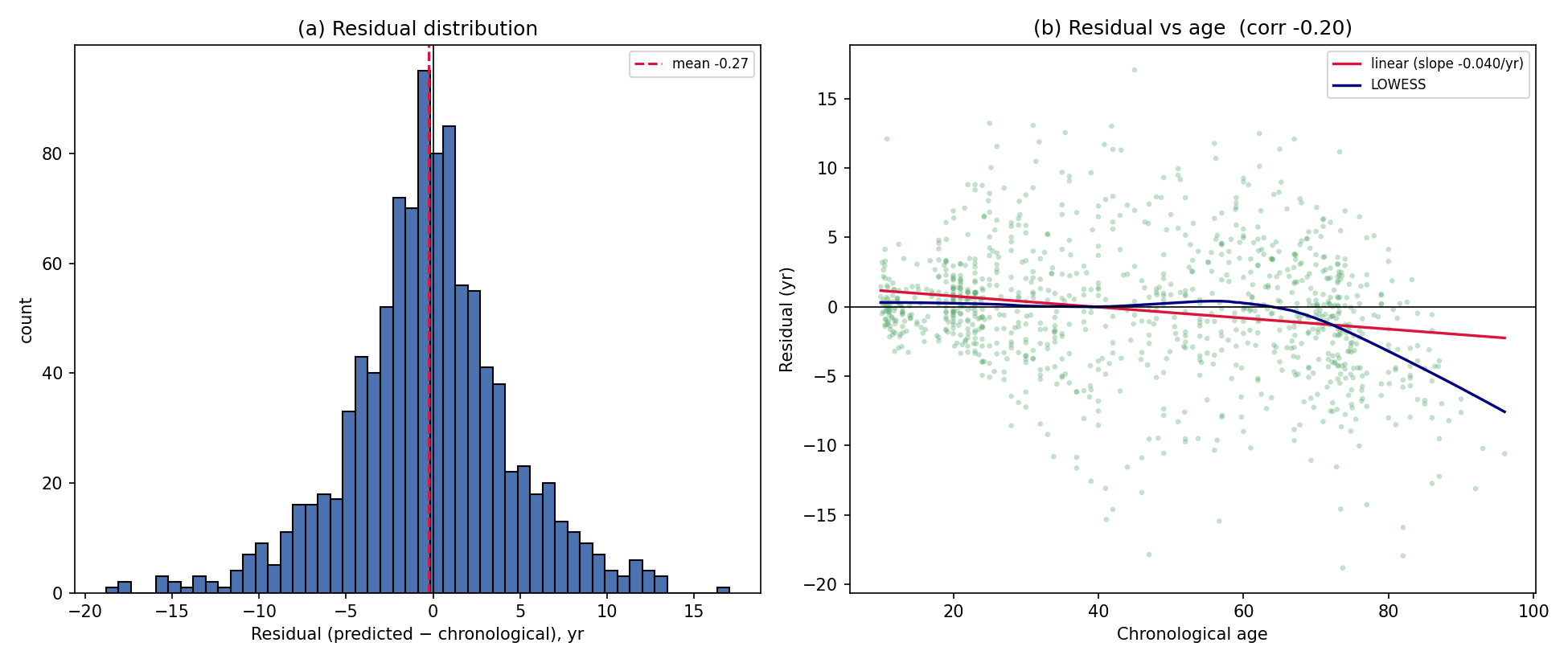}
\caption{(a) Residual distribution. (b) Residual vs chronological age with linear
and LOWESS trends; the trend bends negative beyond $\sim$70~years, indicating
age-dependent bias concentrated at the extremes.}
\label{fig:residuals}
\end{figure}

\begin{figure}[htbp]
\centering
\includegraphics[width=0.72\textwidth]{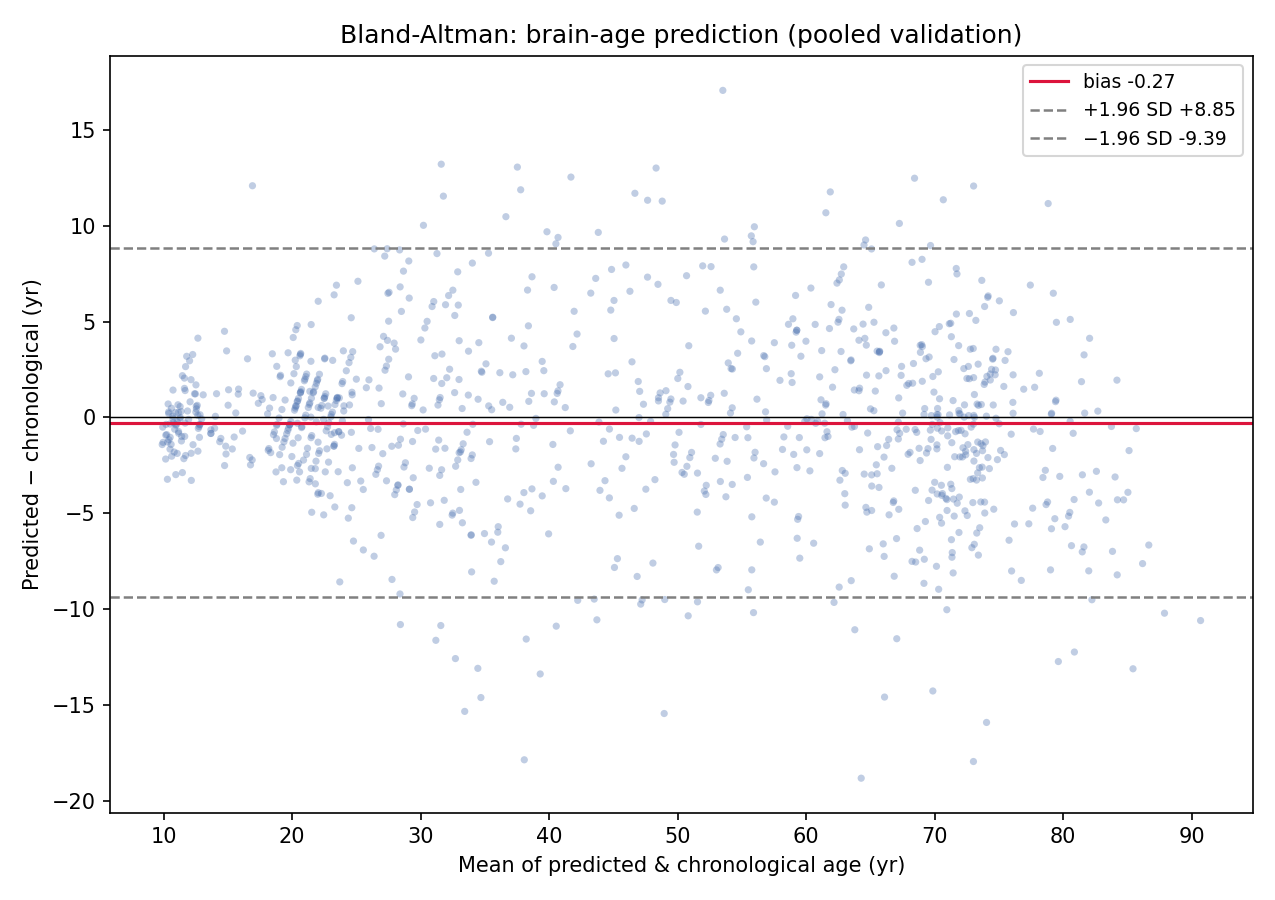}
\caption{Bland--Altman analysis of predicted vs chronological age. Solid line:
mean bias ($-0.27$~yr); dashed lines: $95\%$ limits of agreement
($[-9.39, +8.85]$~yr).}
\label{fig:bland_altman}
\end{figure}

\paragraph{Residual distribution and Bland--Altman analysis.}
Figure~\ref{fig:residuals} presents the residual distribution and the
residual-versus-age relationship with linear and LOWESS trends.
Figure~\ref{fig:bland_altman} gives the Bland--Altman analysis: the mean
difference (bias) is $-0.27$~years with $95\%$ limits of agreement of
$[-9.39, +8.85]$~years, quantifying the individual-level uncertainty that the
summary correlation does not convey.

\begin{figure}[htbp]
\centering
\includegraphics[width=\textwidth]{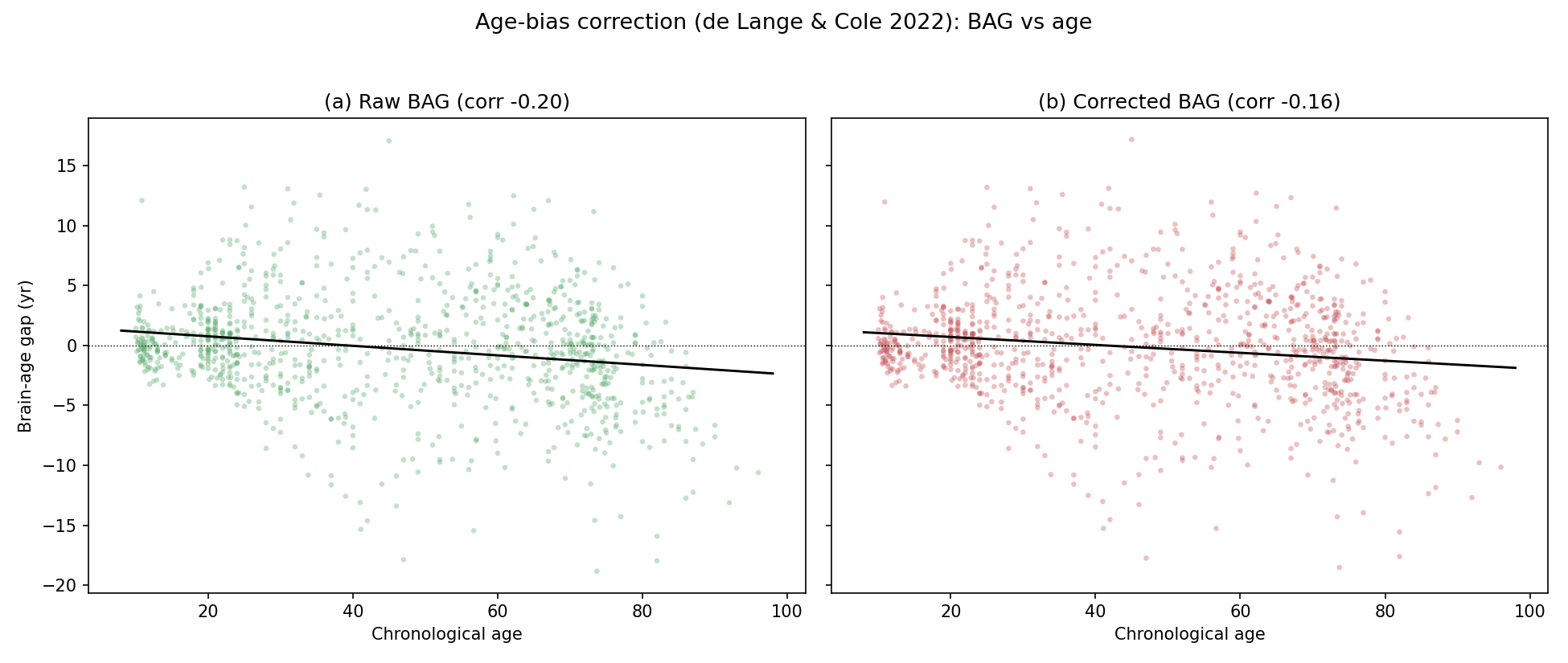}
\caption{Brain-age gap versus chronological age (a) before and (b) after age-bias
correction (trend fit on the training set).}
\label{fig:bag_correction}
\end{figure}

\paragraph{Age-bias-corrected brain-age gap.}
Following de Lange and Cole, we estimate the
brain-age-gap--age trend on the training set (slope $=-0.0067$~yr$^{-1}$) and
subtract it from the validation BAG. As the pooled model is already only mildly
age-biased on average, the correction is modest, reducing the correlation between
BAG and chronological age from $-0.196$ to $-0.164$
(Figure~\ref{fig:bag_correction}); the residual age dependence is concentrated at
the age extremes rather than expressed as a global linear trend.

\section{Preprocessing Pipeline and Quality Control}
\label{app:preprocessing}

\subsection{Pipeline}
All T1-weighted volumes were processed identically across the fifteen source
cohorts. Each raw NIfTI was (i)~skull-stripped and (ii)~corrected for intensity
inhomogeneity with N4 bias-field correction, yielding the
\texttt{*.stripped.N4.nii.gz} volumes used by the model. At load time, each
volume was further (iii)~reoriented to canonical RAS orientation, (iv)~brain-cropped to a tight
bounding box---computed as the largest connected component of the volume
thresholded at $10\%$ of its normalised intensity range---(v)~resampled along the
sagittal axis to a fixed $160$ slices by cubic interpolation, and
(vi)~intensity-normalised per slice to $[0,1]$. For the transformer stage each
slice was replicated to three channels, resized to $224\times224$, and normalised
with the ViT feature-extractor statistics. The model input is therefore a
$160\times224\times224$ (slices$\times$H$\times$W) tensor per subject.

\subsection{Template, Spacing, and Resolution}
Processing was performed in each image's native space, with no alignment to a template; RAS reorientation only permutes and flips axes, and no resampling to a fixed
isotropic voxel spacing was applied before the crop; consequently the bias-corrected volumes
retain their native, dataset-dependent voxel spacings. Spatial
correspondence across subjects is instead imposed at the model interface by the
brain-bounding-box crop followed by fixed-count sagittal resampling
($160$ slices) and in-plane resizing to $224\times224$. The effective model-facing
resolution is thus $160$ sagittal slices of $224\times224$ pixels; physical voxel
spacing is not standardised and varies by cohort.

\subsection{Scanner Harmonization}
No scanner or site harmonization (e.g.\ ComBat or learning-based intensity
harmonization) was applied. Cross-site variability is addressed only implicitly,
through the per-slice intensity normalisation above and the diversity of the
pooled multi-site training data. This is a deliberate design choice to enable deployment on 
unseen external cohorts. However, explicit harmonization can be a direction for future work
to push the model to its limits for predictive performance on the validation set.

\subsection{Quality Control and Exclusion Rules}
The only quality-control exclusion criterion was preprocessing failure: a subject was excluded if
any pipeline stage failed to produce a valid brain volume (e.g.\ skull-stripping
or N4 correction failure, or absence of a valid metadata entry). No additional
manual quality-control rating or automated outlier removal was applied, and no
subjects were excluded at this stage on the basis of age, diagnosis, motion, or model
performance (the subsetting of five cohorts to balance the age distribution of the
training pool is a separate step, described in Section~\ref{supp:subset_selection}). We quantify the preprocessing failure rate per cohort as

\[
\text{failure rate} \;=\;
\frac{N_{\text{original}} - N_{\text{preprocessed}}}{N_{\text{original}}},
\]
where $N_{\text{original}}$ is the number of raw NIfTI volumes provided by the
cohort and $N_{\text{preprocessed}}$ is the number retained for analysis
(Table~\ref{tab:preproc_failure}). Across all fifteen cohorts, $46$ of $11{,}677$
scans ($0.4\%$) failed preprocessing; per-cohort rates ranged from $0.0\%$ to
$5.6\%$, with the highest in sudmex ($5.6\%$), IXI ($2.8\%$), and NIMH ($2.2\%$).

\begin{table}[H]
\centering
\caption{Per-cohort preprocessing failure rates for the training cohorts and the
held-out external test cohorts. $N_{\text{original}}$ is the number of raw
T1-weighted volumes provided; $N_{\text{preprocessed}}$ is the number retained
after skull stripping, N4 correction, and metadata assembly. Excluded scans failed
preprocessing; no other exclusion criteria were applied.}
\label{tab:preproc_failure}
\small
\begin{tabular}{@{}lrrrr@{}}
\toprule
\textbf{Dataset} & $N_{\text{original}}$ & $N_{\text{preprocessed}}$ & \textbf{Excluded} & \textbf{Failure rate} \\
\midrule
\multicolumn{5}{@{}l}{\textit{Training cohorts}} \\
ADNI      & 5896 & 5881 & 15 & 0.3\% \\
IXI       &  581 &  565 & 16 & 2.8\% \\
ABIDE-II  & 1114 & 1114 &  0 & 0.0\% \\
DLBS      &  315 &  315 &  0 & 0.0\% \\
COBRE     &  148 &  146 &  2 & 1.4\% \\
FCON1000  &  870 &  869 &  1 & 0.1\% \\
CoRR      &  377 &  377 &  0 & 0.0\% \\
OASIS-1   &  424 &  424 &  0 & 0.0\% \\
CamCAN    &  653 &  653 &  0 & 0.0\% \\
NIMH      &  185 &  181 &  4 & 2.2\% \\
BOLD      &  158 &  158 &  0 & 0.0\% \\
\addlinespace
\textit{Subtotal} & 10721 & 10683 & 38 & 0.4\% \\
\midrule
\multicolumn{5}{@{}l}{\textit{Held-out test cohorts}} \\
SALD              &  494 &  494 & 0 & 0.0\% \\
TrueCrime         &  133 &  133 & 0 & 0.0\% \\
AgeRisk &  186 &  186 & 0 & 0.0\% \\
SUDMEX-CONN &  143 &  135 & 8 & 5.6\% \\
\addlinespace
\textit{Subtotal} & 956 & 948 & 8 & 0.8\% \\
\midrule
\textbf{Total} & \textbf{11677} & \textbf{11631} & \textbf{46} & \textbf{0.4\%} \\
\bottomrule
\end{tabular}

\vspace{2pt}
\end{table}

% Supplementary subsection: subset selection for the training pool.
% Figures are produced by age_distribution_analysis.py (written to age_distribution/).
% Requires: graphicx, booktabs, subcaption.

\section{Ablation Study: Full Results}
\label{supp:ablation_details}

This section reports the full results and discussion of the twelve ablation experiments summarised in Table~\ref{tab:ablation-summary} and Figure~\ref{fig:ablation-grid} of the main text. All numbers are validation MAE (years) from a single run unless stated otherwise; multi-seed statistics are given in Section~\ref{app:ablations}.

% p1
\subsection{Changing the ViT variant in the backbone}
\label{subsec:ablation-vit-variant}

We replaced the ViT backbone with several popular variants like DeiT \cite{deit}, RopE ViT \cite{heo2024ropevit}, MaxViT \cite{tu2022maxvit} and Swin \cite{liu2021Swin} to observe how sensitive the pipeline is to the transformer choice. The results are summarized in Table~\ref{tab:ablation-vit-variants}.

\begin{table}[h]
\centering
\footnotesize
\setlength{\tabcolsep}{5pt}
\renewcommand{\arraystretch}{1.1}
\caption{Effect of different ViT variants used as the frozen encoder.}
\label{tab:ablation-vit-variants}
\begin{tabular}{lc}
\hline
\textbf{Backbone} & \textbf{MAE}  \\
\hline
ViT-base (ours) & \textbf{3.43} \\
ViT-small & 3.67 \\
ViT-large & 3.71 \\
DeiT & 3.59 \\
DeiT-tiny & 3.66 \\
RoPE ViT & 3.62 \\
MaxViT & 3.48 \\
Swin-tiny & 3.56 \\
\hline
\end{tabular}
\end{table}

The base ViT performs best overall. MaxViT is close but heavier and slower, while DeiT and Swin trail slightly, showing that ViT-base offers the best balance of feature richness and efficiency. DeiT's data-efficient training sacrifices fine-grained spatial encoding needed for volumetric consistency; Swin's hierarchical windows help but still fall short. RoPE ViT's rotational positional invariance does not improve regression accuracy here. Overall, ViT-base remains the most reliable choice.

% p2
\subsection{Age–sex grouping granularity}
\label{subsec:ablation-age-sex-groups}

We examined the effect of changing the bin width for age–sex composite classes during ViT pretraining. Table~\ref{tab:ablation-age-sex-groups} shows the results.

\begin{table}[h]
\centering
\footnotesize
\setlength{\tabcolsep}{5pt}
\renewcommand{\arraystretch}{1.1}
\caption{Effect of age–sex group granularity used during ViT pretraining.}
\label{tab:ablation-age-sex-groups}
\begin{tabular}{lc}
\hline
\textbf{Group Width (years)} & \textbf{MAE}  \\
\hline
5 & 4.72 \\
10 (ours) & \textbf{3.43} \\
20 & 5.26 \\
\hline
\end{tabular}
\end{table}

Age–sex categories were treated as discrete labels under cross-entropy loss during pretraining, encouraging the model to cluster features along group boundaries. Narrow bins (5 years) overfit to minor inter-class variation due to fewer examples per class, while broad bins (20 years) blur morphological aging cues and weaken the classification signal. The 10-year grouping strikes the best balance, giving enough inter-class contrast while preserving intra-class consistency, which improves downstream regression performance.

%p3
\subsection{Number of evenly spaced slices during ViT training}
\label{subsec:ablation-num-slices}

We trained the ViT with different numbers of evenly spaced sagittal slices per subject. The performance is reported in Table~\ref{tab:ablation-slices}.

\begin{table}[h]
\centering
\footnotesize
\setlength{\tabcolsep}{5pt}
\renewcommand{\arraystretch}{1.1}
\caption{Effect of the number of evenly spaced slices sampled per volume during ViT pretraining.}
\label{tab:ablation-slices}
\begin{tabular}{lc}
\hline
\textbf{Slices per Volume} & \textbf{MAE}  \\
\hline
16 & 3.95 \\
32 (ours) & \textbf{3.43} \\
64 & 3.77 \\
\hline
\end{tabular}
\end{table}

Too few slices limit spatial coverage and the ViT's ability to capture subtle structural variation, while too many slices add redundancy and computational cost from highly correlated adjacent slices without improving feature learning. The 32-slice setting balances anatomical coverage with training efficiency and avoids overfitting to redundant information.

% p4
\subsection{Architecture variations in front of ViT}
\label{subsec:ablation-front-arch}

We compared several alternative designs to replace the CNN regression head.
Table~\ref{tab:ablation-front-arch} summarizes the results. In addition to
the Bidirectional GRU, we evaluate a Sequential Transformer operating over
slice embeddings and a Bidirectional Mamba state-space model, providing
representatives of recurrence-, attention-, and SSM-based sequence modeling.

\begin{table}[h]
\centering
\footnotesize
\setlength{\tabcolsep}{5pt}
\renewcommand{\arraystretch}{1.1}
\caption{Effect of different architectures placed after the ViT encoder.}
\label{tab:ablation-front-arch}
\begin{tabular}{lc}
\hline
\textbf{Front End} & \textbf{MAE} \\
\hline
CNN with residual blocks (ours) & \textbf{3.43} \\
ViT & 6.51 \\
MLP only & 8.12 \\
Bidirectional GRU & 3.76 \\
Bidirectional Mamba & 3.81 \\
Sequential Transformer & 3.76 \\
\hline
\end{tabular}
\end{table}

The MLP and ViT-only alternatives perform poorly compared with architectures
that explicitly model relationships across slice embeddings. The
Bidirectional GRU, Sequential Transformer, and Bidirectional Mamba provide
three complementary mechanisms for sequence modeling and substantially
reduce the error. However, the residual CNN
achieves the lowest MAE, suggesting that local inter-slice structure captured
by convolution is particularly useful for this task. Since the sequence models still
approach our method's MAE quite closely, a multi-seed analysis to gauge the 
robustness and statistical significance of this comparison is included in Section~\ref{sup:subsec:frontend}.

% p5
\subsection{Methodological Variants in ViT and CNN Integration}

To further examine the influence of different ways of incorporating sex and age conditioning into the hybrid framework, we conducted additional experiments using the same ViT backbone (ViT-Base) and the best-performing CNN regression head. The goal of these tests was to isolate the contribution of age-sex representation strategies and fusion mechanisms to the final performance. Table~\ref{tab:vit-cnn-variants} summarizes the results.

\begin{table}[h]
\centering
\footnotesize
\setlength{\tabcolsep}{5pt}
\renewcommand{\arraystretch}{1.1}
\caption{Effect of methodological variants using ViT-Base encoder and best CNN regression head. All results are reported on the validation set.}
\label{tab:vit-cnn-variants}
\begin{tabularx}{\columnwidth}{lc}
\hline
\textbf{Model Variant (ViT)} & \textbf{MAE} \\
\hline
\textbf{With age–sex composite classes (ours)} & \textbf{3.43} \\
With age classes only (no sex) & 5.07 \\
With age classes + sex input via FiLM & 5.76 \\
With direct age regression head (no sex) & 5.35 \\
With regression head + sex input via FiLM & 7.06 \\
With sex-aware encoder but CNN without sex input & 3.56 \\
\hline
\end{tabularx}
\end{table}

As shown in Table~\ref{tab:vit-cnn-variants}, jointly encoding age and sex as composite classes during pretraining and fusing them in the CNN head achieves the lowest MAE (3.43 years). Removing sex or using direct age regression raises MAE above 4.9 years, and FiLM-based sex conditioning in either the ViT encoder or CNN head performs worse still (5.76 and 7.06, respectively). Sex-aware encoder supervision without sex input in the CNN head yields a smaller increase (3.56). These results indicate that composite-class encoding produces more robust representations than continuous regression or FiLM conditioning.

\subsection{2D vs.\ 1D Convolution in the CNN}
\label{subsec:ablation-cnn-dimensionality}

To assess whether performance gains arise solely from inter-slice adjacency or also benefit from local structure across the embedding dimension, we compare the proposed 2D CNN with a 1D CNN variant that performs convolution only along the slice axis.

\begin{table}[h]
\centering
\footnotesize
\setlength{\tabcolsep}{4pt}
\renewcommand{\arraystretch}{1.1}
\caption{Comparison of 2D and 1D CNN front-ends on validation performance (seed 701; five-seed statistics in Section~\ref{sup:subsec:2dv1d}).}
\label{tab:ablation-cnn-dimensionality-main}
\begin{tabular}{lc}
\hline
\textbf{Model Variant} & \textbf{MAE} \\
\hline
2D CNN & \textbf{3.43} \\
1D CNN & 3.64 \\
\hline
\end{tabular}
\end{table}

The 2D CNN achieves lower validation MAE across all five random seeds, suggesting that convolution across both the slice and embedding dimensions provides a modest but consistent advantage over convolution restricted to the slice axis alone. This supports the 2D formulation empirically but does not imply that the embedding dimensions possess an intrinsic spatial ordering; the gain may instead reflect the additional weight sharing and local feature mixing introduced by 2D kernels. Full per-seed results are reported in Section~\ref{sup:subsec:2dv1d}.

% p6
\subsection{Number of convolutional layers}
\label{subsec:ablation-num-conv}

We experimented with varying depths for the CNN regression head. The original three-block configuration remains the best performer, as shown in Table~\ref{tab:ablation-num-conv}.

\begin{table}[h]
\centering
\footnotesize
\setlength{\tabcolsep}{5pt}
\renewcommand{\arraystretch}{1.1}
\caption{Effect of varying the number of convolutional blocks in the CNN regression head.}
\label{tab:ablation-num-conv}
\begin{tabular}{lc}
\hline
\textbf{Number of Conv Blocks} & \textbf{MAE}  \\
\hline
1 & 4.23 \\
2 & 3.76 \\
3 (ours) & \textbf{3.43} \\
4 & 3.49 \\
5 & 3.54 \\
6 & 3.73 \\
\hline
\end{tabular}
\end{table}

Shallower networks underfit from limited receptive fields and weak hierarchical feature extraction, while networks deeper than three blocks gain parameters but slightly degrade performance from over-smoothing and redundancy. Three blocks give the best trade-off between expressiveness and generalization, keeping training stable and inference efficient.

%p7
\subsection{CNN kernel sizes}
\label{subsec:ablation-kernels}

We tested multiple kernel configurations in the residual CNN blocks. Results are given in Table~\ref{tab:ablation-kernels}.

\begin{table}[h]
\centering
\footnotesize
\setlength{\tabcolsep}{5pt}
\renewcommand{\arraystretch}{1.1}
\caption{Effect of different kernel size configurations in the residual CNN.}
\label{tab:ablation-kernels}
\begin{tabular}{lc}
\hline
\textbf{Kernel Sizes} & \textbf{MAE}  \\
\hline
(10,60),(5,15),(2,6) (ours) & \textbf{3.43} \\
(12,80),(6,20),(3,8) & 3.47 \\
(6,40),(4,12),(2,6) & 3.82 \\
(8,50),(5,15),(1,4) & 3.54 \\
Dilated variant & 3.50 \\
\hline
\end{tabular}
\end{table}

The selected kernel sizes best balance cross-slice coverage and fine-grained precision. Larger kernels extend the receptive field but blur subtle anatomical differences; smaller kernels underfit from limited context. The chosen configuration preserves both local structure and inter-slice coherence.

%p8
\subsection{Activation functions}
\label{subsec:ablation-activations}

We replaced the activation used in the CNN and fully connected layers to test the effect of nonlinearity. Results are shown in Table~\ref{tab:ablation-activations}.

\begin{table}[h]
\centering
\footnotesize
\setlength{\tabcolsep}{5pt}
\renewcommand{\arraystretch}{1.1}
\caption{Effect of different activation functions in the CNN and FC layers.}
\label{tab:ablation-activations}
\begin{tabular}{lc}
\hline
\textbf{Activation} & \textbf{MAE}  \\
\hline
SiLU (ours) & \textbf{3.43} \\
ReLU & 3.48 \\
Leaky ReLU & 3.52 \\
GELU & 3.71 \\
\hline
\end{tabular}
\end{table}

SiLU yields the lowest error, likely due to smooth gradient flow and compatibility with residual layers. ReLU performs reasonably well but lacks SiLU's smoother transitions; Leaky ReLU offers no clear benefit over ReLU here; GELU trails all three. SiLU provides the most stable convergence and generalization.

%p9

\subsection{Impact of Late Sex Fusion}
\label{subsec:ablation-sex-fusion}
To evaluate the contribution of sex information in the CNN regression stage, we compare models with and without late fusion of the binary sex variable at the final fully connected layer. As shown in Table~\ref{tab:sex_fusion}, sex fusion improves accuracy, lowering validation MAE from \textbf{3.56} to \textbf{3.43} years. This confirms that sex-specific differences in brain structure matter for accurate age estimation. Late fusion lets earlier layers learn generalized, sex-independent representations while incorporating sex only at the final stage, improving robustness and interpretability.

\begin{table}[h!]
\centering
\caption{Effect of incorporating biological sex information in the CNN regression head.}
\label{tab:sex_fusion}
\begin{tabular}{l c}
\hline
\textbf{Configuration} & \textbf{Validation MAE (years)} \\
\hline
With Sex Fusion & \textbf{3.43} \\
Without Sex Fusion & 3.56 \\
\hline
\end{tabular}
\end{table}

%p10
\subsection{Effect of Learning Rate Variation}
\label{subsec:ablation-lr}

We further study the influence of the learning rate on convergence behavior and final predictive performance. The proposed model was originally trained with a learning rate of $5\times10^{-4}$, which achieved the best trade-off between stability and accuracy. To explore the sensitivity of the model to this hyperparameter, we vary the learning rate across an order of magnitude and interpolate the results around the optimal configuration.

\begin{table}[h]
\centering
\footnotesize
\setlength{\tabcolsep}{3pt} % reduce horizontal spacing
\renewcommand{\arraystretch}{1.1} % adjust vertical spacing
\caption{Effect of learning rate on model performance (validation set).}
\label{tab:ablation-lr}
\begin{tabular}{lccccc}
\hline
\textbf{\makecell{Learning \\ Rate}} 
    & $1\times10^{-4}$ 
    & \textbf{\makecell{$5\times10^{-4}$ \\ (ours)}} 
    & $1\times10^{-3}$ 
    & $5\times10^{-3}$ 
    & $1\times10^{-2}$ \\
\hline
\textbf{MAE}    & 4.52 & \textbf{3.43} & 3.83 & 5.10 & 6.13 \\
\textbf{r}      & 0.96 & \textbf{0.98} & 0.97 & 0.94 & 0.92 \\
\textbf{R$^{2}$} & 0.90 & \textbf{0.96} & 0.92 & 0.86 & 0.80 \\
\hline
\end{tabular}
\end{table}

As shown in Table~\ref{tab:ablation-lr}, both very low and very high learning rates hurt performance: lower rates slow convergence and risk local minima, while higher rates cause unstable updates and overfitting. A rate of $5\times10^{-4}$ yields the best overall metrics, indicating a stable optimization regime for the hybrid architecture.

%p11
\subsection{Effect of Removing Residual Connections in CNN}
\label{subsec:ablation-residual}

Residual connections are known to facilitate gradient propagation and stabilize deep convolutional architectures. To verify their impact, we train a variant of the CNN regression head where skip connections are removed, forcing the network to learn purely sequential transformations.

\begin{table}[h]
\centering
\footnotesize
\setlength{\tabcolsep}{3pt} % tighten horizontal spacing
\renewcommand{\arraystretch}{1.1} % adjust vertical spacing
\caption{Impact of removing residual connections in the CNN on validation performance.}
\label{tab:ablation-residual}
\begin{tabular}{lcccc}
\hline
\textbf{\makecell[l]{Model Variant}} & 
\textbf{MAE}  \\
\hline
\makecell[l]{With Residual Connections} & 
\textbf{3.43} \\
\makecell[l]{Without Residual Connections} & 
3.70 \\
\hline
\end{tabular}
\end{table}

Removing residual pathways moderately increases MAE (from 3.43 to 3.70), confirming that residual links stabilize learning and preserve fine-grained spatial representations while avoiding vanishing gradients.

\section{Statistical Details of Ablation Study}
\label{app:ablations}

For each ablation, we retrain the model with $n \geq 3$ random seeds, holding
the evaluation split fixed across all runs. We report the mean $\pm$ standard
deviation (SD) of the mean absolute error (MAE, in months) across seeds, and
assess significance with a paired $t$-test on seed-matched runs (same seed
used to initialize and shuffle training data in both arms of each
comparison). 

\subsection{Residual connections}

Table~\ref{tab:ablation_residual} compares the model with and without
residual connections across five seeds. Performance without residual
connections is not merely lower on average but \emph{unstable}: three of five
seeds (700, 701, 702) yield MAE within 0.3 of the with-residual condition,
while two seeds (703, 704) diverge sharply (MAE 5.01 and 4.53), driving the
SD from 0.047 (with residual) to 0.563 (without) --- a more than tenfold
increase in run-to-run variability. This pattern is consistent with residual
connections stabilizing optimization rather than simply shifting the mean
error.

\begin{table}[h]
\centering
\caption{MAE (months) with and without residual connections, across 5 seeds.}
\label{tab:ablation_residual}
\begin{tabular}{lcc}
\toprule
Seed & With residual & Without residual \\
\midrule
700 & 3.51 & 3.92 \\
701 & 3.43 & 3.70 \\
702 & 3.48 & 3.78 \\
703 & 3.56 & 5.01 \\
704 & 3.50 & 4.53 \\
\midrule
Mean $\pm$ SD & $3.496 \pm 0.047$ & $4.188 \pm 0.563$ \\
\bottomrule
\end{tabular}

\vspace{4pt}
\small
Paired $t$-test: $t(4) = -2.95$, $p = 0.042$. 
\end{table}

\subsection{Front-end architecture: CNN vs.\ sequence models}
\label{sup:subsec:frontend}

Table~\ref{tab:ablation_frontend} compares the convolutional front-end against
three sequence-modeling alternatives operating over the slice embeddings:
a Bidirectional GRU, a Sequential Transformer, and a Bidirectional Mamba
state-space model. These alternatives represent recurrence, attention, and
state-space modeling, respectively. Across five seeds, the CNN consistently
achieves the lowest mean validation MAE.

\begin{table}[h]
\centering
\caption{MAE (years) for alternative front-end architectures across 5 seeds.}
\label{tab:ablation_frontend}
\begin{tabular}{lcccc}
\toprule
Seed & CNN front-end & Bidirectional GRU & Bidirectional Mamba & Sequential Transformer \\
\midrule
700 & 3.51 & 3.83 & 3.805 & 3.744 \\
701 & 3.43 & 3.76 & 3.814 & 3.758 \\
702 & 3.48 & 3.81 & 3.599 & 3.774 \\
703 & 3.56 & 3.78 & 3.688 & 3.654 \\
704 & 3.50 & 3.73 & 3.642 & 3.880 \\
\midrule
Mean $\pm$ SD
& $3.496 \pm 0.047$
& $3.782 \pm 0.040$
& $3.710 \pm 0.097$
& $3.762 \pm 0.081$ \\
\bottomrule
\end{tabular}

\vspace{4pt}
\small
Paired $t$-tests against the CNN front-end:
Bidirectional GRU, $t(4)=-11.43$, $p=0.0003$;
Bidirectional Mamba, $t(4)=-4.00$, $p=0.0161$;
Sequential Transformer, $t(4)=-5.42$, $p=0.0056$.
\end{table}

The three sequence-modeling alternatives all yield higher mean MAE than the
CNN front-end. Among them, Bidirectional Mamba has the lowest mean MAE
($3.710 \pm 0.097$), followed by the Sequential Transformer
($3.762 \pm 0.081$) and Bidirectional GRU ($3.782 \pm 0.040$).
The difference between Bidirectional Mamba and the Sequential Transformer is
small relative to their seed-to-seed variation, and their ordering is not
consistent across seeds: the Sequential Transformer performs better for
seeds 700, 701, and 703, whereas Bidirectional Mamba performs better for
seeds 702 and 704. Thus, neither attention nor state-space modeling shows a
uniform advantage over the other, while the CNN remains consistently stronger
across these runs.

\subsection{Front-end architecture: Sex fusion}

Table~\ref{tab:ablation_sex} compares the model with and without sex
incorporated as an input feature, across three seeds.

\begin{table}[h]
\centering
\caption{MAE (months) with and without sex incorporated, across 3 seeds.}
\label{tab:ablation_sex}
\begin{tabular}{lcc}
\toprule
Seed & Sex incorporated & Sex not incorporated \\
\midrule
700 & 3.51 & 3.71 \\
701 & 3.43 & 3.56 \\
702 & 3.48 & 3.74 \\
\midrule
Mean $\pm$ SD & $3.473 \pm 0.040$ & $3.670 \pm 0.096$ \\
\bottomrule
\end{tabular}

\vspace{4pt}
\small
Paired $t$-test: $t(2) = -5.24$, $p = 0.035$. 
\end{table}

\subsection{Front-end architecture: 2D vs.\ 1D convolution}
\label{sup:subsec:2dv1d}
Table~\ref{tab:ablation_cnn_dimensionality} compares the 2D CNN against a
1D CNN variant. Both architectures show low seed-to-seed variance, but the
2D CNN achieves lower validation MAE across all five seeds.

\begin{table}[H]
\centering
\caption{MAE (years) for 2D CNN vs.\ 1D CNN, across 5 seeds.}
\label{tab:ablation_cnn_dimensionality}
\begin{tabular}{lcc}
\toprule
Seed & 2D CNN & 1D CNN \\
\midrule
700 & 3.51 & 3.55 \\
701 & 3.43 & 3.64 \\
702 & 3.48 & 3.61 \\
703 & 3.56 & 3.59 \\
704 & 3.50 & 3.58 \\
\midrule
Mean $\pm$ SD & $3.496 \pm 0.047$ & $3.593 \pm 0.039$ \\
\bottomrule
\end{tabular}

\vspace{4pt}
\small
Paired $t$-test: $t(4) = -2.963$, $p = 0.041$.
\end{table}

\section{Dataset Descriptions}
\label{supp:datasets}

Demographic summaries of all datasets are in Table~\ref{tab:dataset_info} of the main text. Here we describe each dataset and the subset used.

\subsection{Training and validation datasets}

\textbf{The Autism Brain Imaging Data Exchange II (ABIDE II)} \cite{abideII} is a large, multisite, open-access case–control dataset advancing research on brain connectivity and structure in Autism Spectrum Disorder (ASD), comprising scans from individuals with ASD and typically developing controls in roughly equal proportions, ages 5 to 64. We use a curated subset of 750 high-quality structural MRI scans for brain age modeling. For disease-association analyses we instead use the complete ABIDE-II cohort (N = 1,114), rather than the balanced subset used for the main multi-site model, to maximize diagnostic-group sample size.\\

\textbf{The Alzheimer’s Disease Neuroimaging Initiative (ADNI)} \cite{Mueller2005} is a longitudinal multicenter study identifying biomarkers of Alzheimer’s disease progression, launched in 2003 as a public-private partnership led by Principal Investigator Michael W. Weiner, MD. It provides multimodal neuroimaging data alongside clinical, genetic, and cognitive assessments. We use 900 T1-weighted MRI scans of unique subjects; for disease-association analyses we use the complete ADNI cohort (N = 5,881; subject-level split) rather than the balanced subset.\\

The dataset \cite{rieck2024boldvar} from the \textbf{BOLD variability study} by Rieck et al. (2021) \cite{rieck2021reconfiguration, rieck2021contributions} includes MRI data from 158 healthy, cognitively normal adults (aged 20–86), recruited from the greater Toronto area, investigating age-related differences in cognitive control (inhibition, task switching, working memory). We incorporate structural MRI scans from all participants.\\

\textbf{The Cambridge Centre for Ageing and Neuroscience (Cam-CAN)} \cite{taylor2016camcan} \cite{shafto2014camcan} project studies the neural mechanisms underlying healthy cognitive aging in a population-based cohort aged 18–90. We use T1-weighted MRI volumes from 653 adults.\\

\textbf{The Center for Biomedical Research Excellence (COBRE)} dataset \cite{cobre} comprises anatomical and resting-state functional MRI data from 72 schizophrenia patients and 75 healthy controls, aged 18–65. We include 146 samples.\\

\textbf{The Consortium for Reliability and Reproducibility (CORR)} \cite{Zuo2014_CoRR} dataset aggregates resting-state MRI and anatomical data from multiple international sites to assess and improve reliability of brain imaging findings. We include 377 anatomical scans, enhancing protocol variability and robustness.\\

\textbf{The Dallas Lifespan Brain Study (DLBS)} \cite{dlbs, park2025dlbs} investigates age-related changes in brain structure and function across the adult lifespan (20–89 years). We use T1-weighted MRI volumes from 315 participants.\\

\textbf{The 1000 Functional Connectomes Project (FCON1000)} \cite{re3data_1000functionalconnectomes_2025} includes data from multiple independent studies focused on resting-state functional connectivity. We use MRI volumes from 763 adults pooled across sites, adding variability valuable for robust machine learning.\\

\textbf{The IXI (Information eXtraction from Images) dataset} \cite{ixi_dataset} is a multi-center, open-access collection of structural MRIs from cognitively healthy adults scanned at three London hospitals. We take 565 samples aged 18–88.\\

\textbf{The National Institute of Mental Health (NIMH) Healthy Research Volunteer (RV)} dataset \cite{nimh, nugent2022nimh} includes structural MRI, functional MRI, diffusion tensor imaging, and detailed clinical/cognitive assessments, characterizing healthy adults while excluding any history of mental illness. We use a subset of \( n = 181 \) participants.\\

\textbf{The Open Access Series of Imaging Studies (OASIS)} includes a cross-sectional MRI dataset (OASIS-1, \( n = 416 \)) \cite{oasis_data}, covering ages 18–96. We use \( n = 300 \) individuals from OASIS-1.\\

\subsection{Independent test cohorts}

\textbf{The Southwest University Adult Lifespan Dataset (SALD)} \cite{sald_dataset} is a large, cross-sectional, multimodal neuroimaging collection of 494 healthy adults aged 19–80, mapping structural and functional changes across adulthood. We include all 494 participants to enhance adult lifespan coverage.\\

\textbf{The SUDMEX-CONN} dataset \cite{sudmex, angelesvaldez2022sudmex} is a case-control study of cocaine use disorder patients from the National Institute of Psychiatry in Mexico City, with T1-weighted, 10-minute resting-state fMRI, and HARDI-DWI multishell sequences. We include 135 subjects.\\

The \textbf{TrueCrime} dataset \cite{truecrime} provides structural and resting-state MRI scans for 133 participants from a University of Graz study on the psychology of true crime consumption; no tasks were performed during scanning. We use the T1-weighted structural scans.\\

The \textbf{AgeRisk} \cite{agerisk} dataset provides cross-sectional MRI data from healthy participants aged 16–81, collected as part of a study on risk preference and impulsivity, including a T1-weighted structural scan and functional scans. We utilize 186 skull-stripped T1-weighted MRI scans.\\

\section{Subset selection and age balancing of the training pool}
\label{supp:subset_selection}

The combined training pool contains 5{,}108 T1-weighted scans from eleven
cohorts (ages 8--96; mean 46.1, median 47.0 years; 2{,}557 male, 2{,}551 female).
Six cohorts (IXI, DLBS, COBRE, Cam-CAN, NIMH, BOLD) contribute every scan that
passed our preprocessing pipeline (skull stripping and N4 bias-field correction)
and had age and sex recorded. For the remaining five cohorts we used only part
of the available data (Table~\ref{tab:supp_subsets}). The aim was to keep any
one cohort from dominating the age distribution of the pool.

\begin{figure}[H]
\centering
\includegraphics[width=\linewidth]{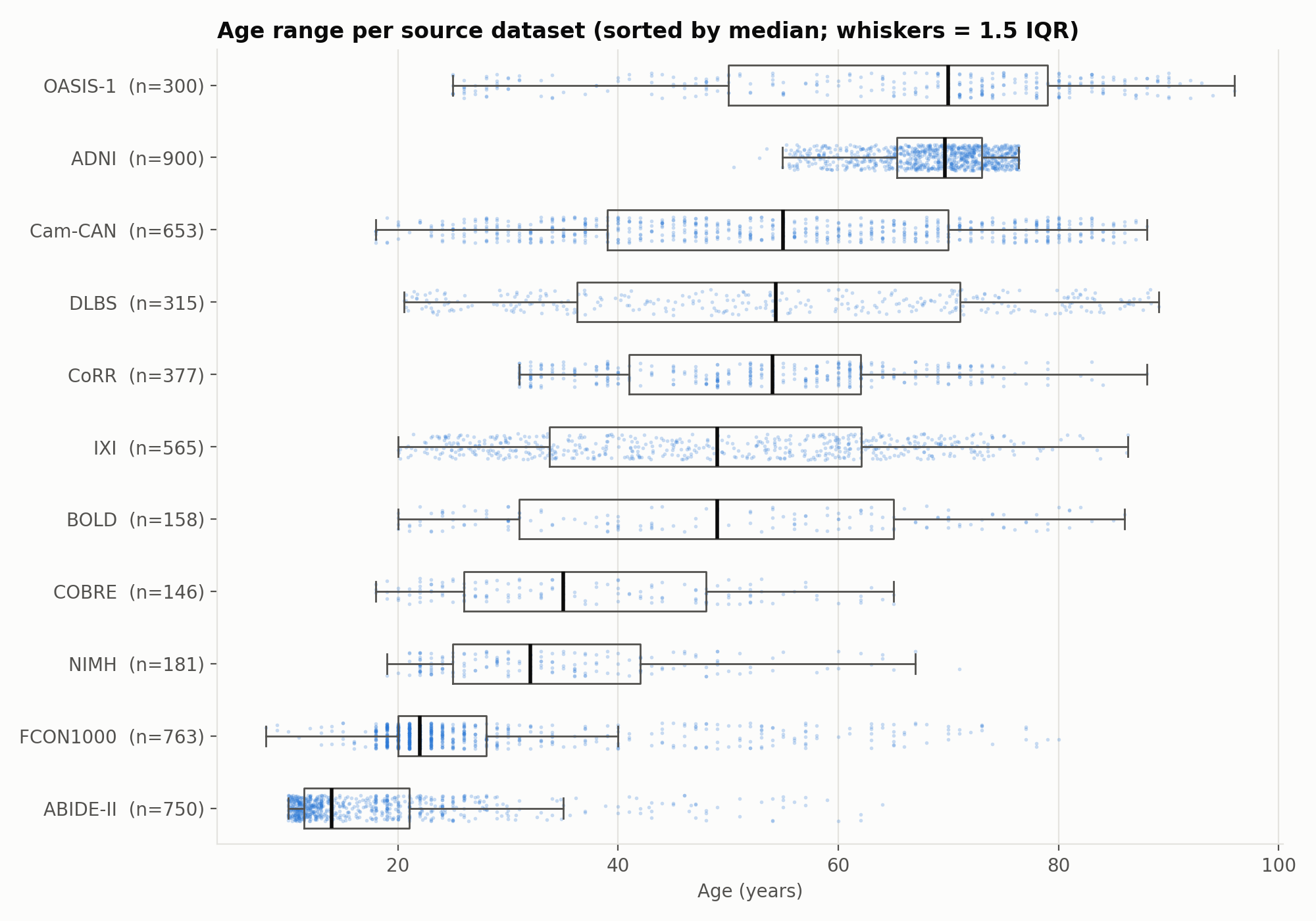}
\caption{Age range of each cohort after subset selection, sorted by median age.
Boxes show the interquartile range, whiskers extend to 1.5\,IQR, and points are
individual scans.}
\label{fig:supp_age_box}
\end{figure}

\begin{figure}[H]
\centering
\includegraphics[width=\linewidth]{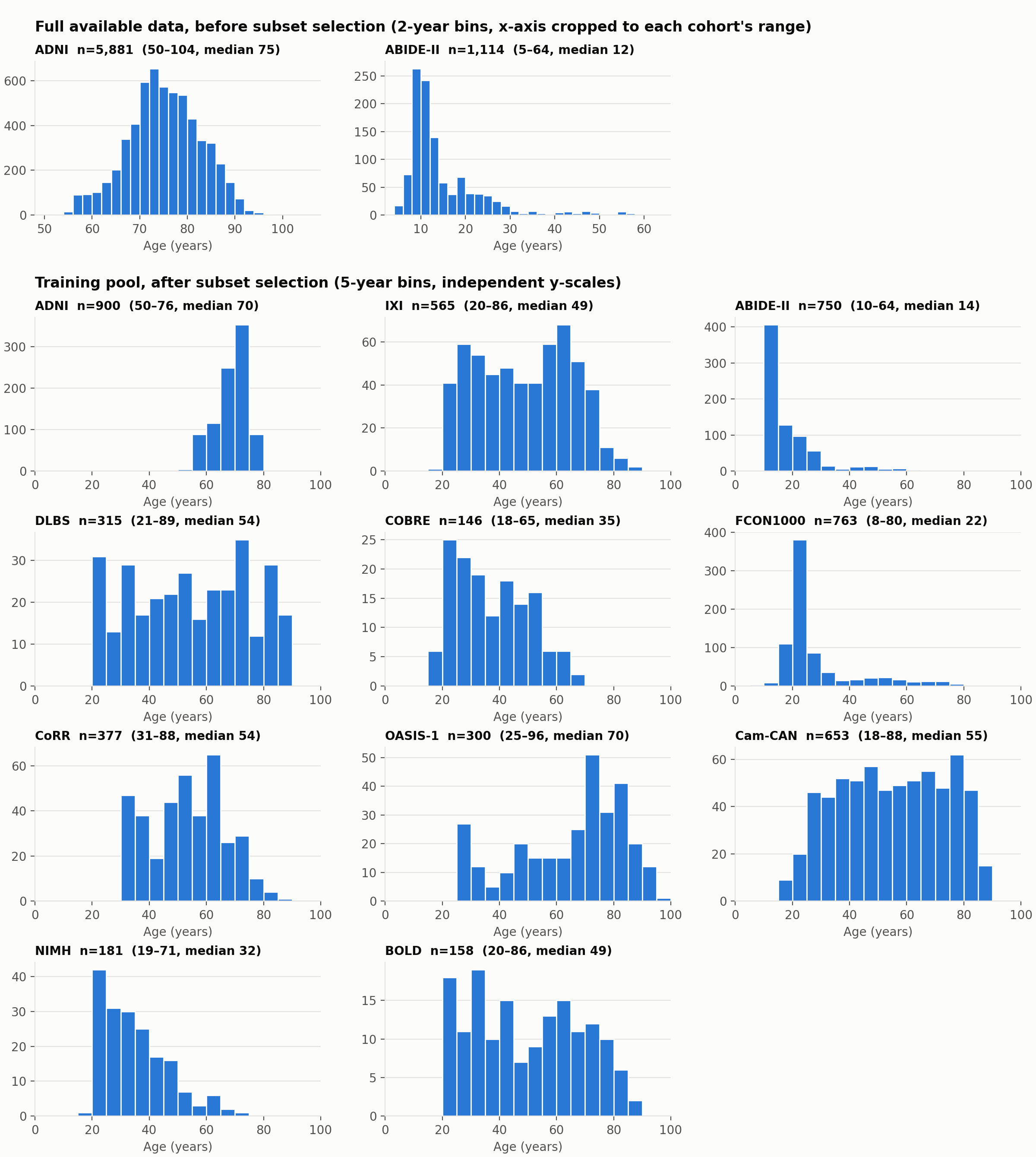}
\caption{Age distribution of each cohort after subset selection (5-year bins;
y-axes scaled independently).}
\label{fig:supp_age_per_dataset}
\end{figure}

\begin{figure}[H]
\centering
\includegraphics[width=\linewidth]{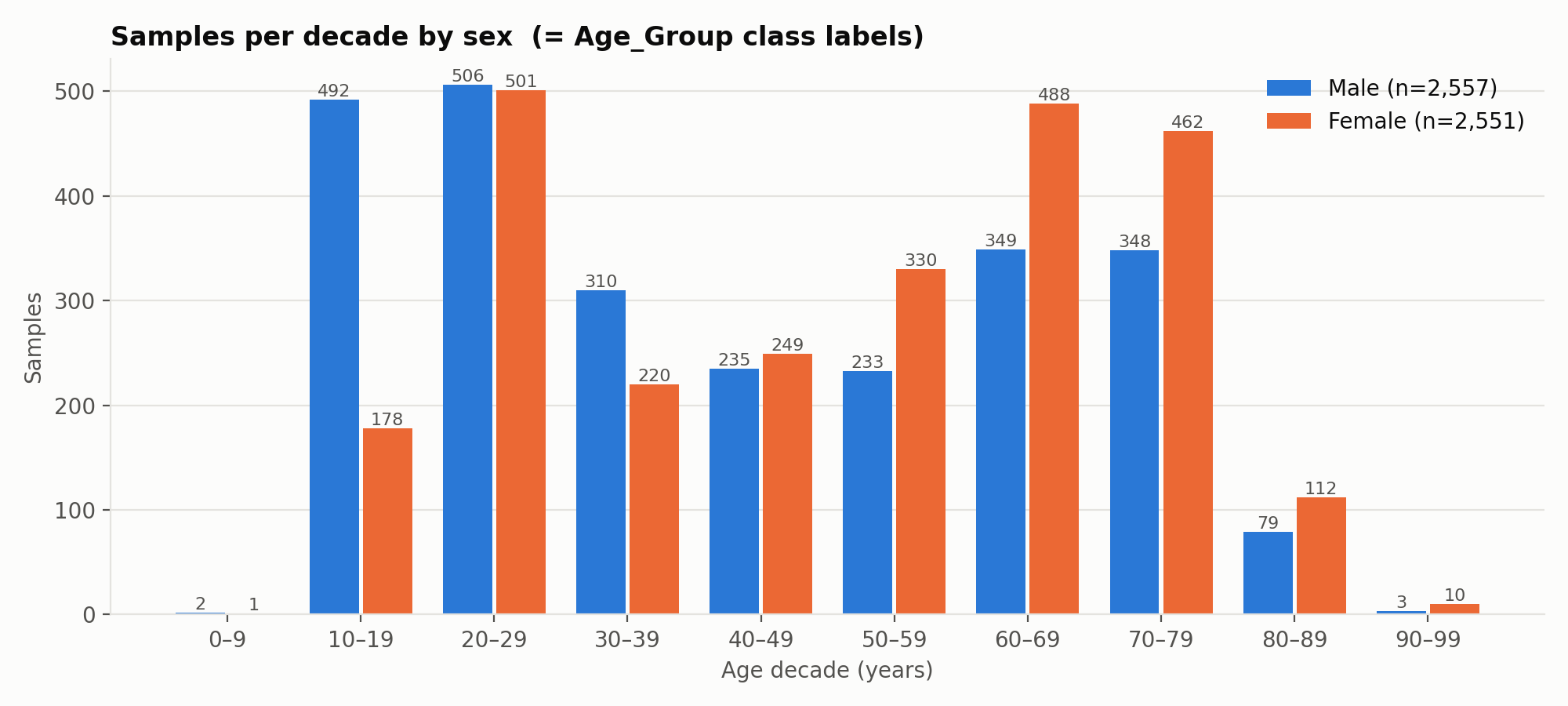}
\caption{Scans per age decade by sex. The decade--sex combinations are the
\texttt{Age\_Group} classes used as targets in the ViT stage. Males outnumber
females in the teens because of ABIDE-II; females outnumber males above 50.}
\label{fig:supp_age_sex}
\end{figure}

The two largest cohorts have strongly skewed age distributions: ADNI is
concentrated in older adults and ABIDE-II in children and adolescents. Using
either in full would have tilted the pool toward that end of the lifespan, and
the model would have learned site identity as a proxy for age. We therefore
capped their contributions and kept all usable scans from the cohorts whose ages
spread across mid-life (Cam-CAN, DLBS, IXI, CoRR, BOLD).
Figure~\ref{fig:supp_age_sex} shows the resulting age distribution, and
Figures~\ref{fig:supp_age_per_dataset} and~\ref{fig:supp_age_box} show each
cohort's contribution.

\begin{table}[h]
\centering
\small
\caption{Cohorts from which only part of the available data was used.
``Available'' is the number of scans in the public release; ``Used'' is the
number of scans in the training pool.}
\label{tab:supp_subsets}
\begin{tabular}{lrrrl}
\toprule
Cohort & Available & Used & Unique subjects & Selection rule \\
\midrule
ADNI     & 5{,}881 & 900 & 900 & 1 scan/subject; youngest 900 subjects (50.5--76.4 y) \\
ABIDE-II & 1{,}114 & 750 & 749 & oldest 750 scans (10--64 y) \\
FCON1000 & 1{,}355 & 763 & 763 & all processed scans with age and sex recorded \\
CoRR     & 3{,}357 & 377 & 256 & all scans meeting the MPRAGE T1w criterion \\
OASIS-1  &     416 & 300 & 294 & oldest 300 scans (25--96 y) \\
\bottomrule
\end{tabular}
\end{table}

\paragraph{ADNI.} The ADNI collection used in the study contains 5{,}881 scans from 1{,}378
subjects, most of whom were scanned longitudinally. We first reduced each subject
to a single scan: the session whose age is closest to that subject's median age
across visits. We then kept the 900 youngest subjects (50.5--76.4 years). This
removes the oldest part of the ADNI range, which would otherwise dominate the
65+ age bands; older ages are instead covered by OASIS-1, Cam-CAN and DLBS. All
900 ADNI scans come from distinct subjects. The selected subjects are
514 cognitively normal (CN), 262 with mild cognitive impairment (MCI) and 124
with Alzheimer's disease (AD).

\paragraph{ABIDE-II.} Most ABIDE-II participants are children. We kept the 750
oldest scans (10--64 years; 395 controls, 355 ASD) and dropped the youngest
participants, who would otherwise have outnumbered all other cohorts below age 20.

\paragraph{FCON1000.} FCON1000 is also skewed young (median 22 years in our
pool). We did not apply an age cut. We used every scan that passed
preprocessing (869) and excluded the 106 whose age or sex is missing in the
release, leaving 763 scans. FCON1000 provides one anatomical scan per
participant, and all 763 scans in the pool come from distinct subjects
(13 sites).

\paragraph{CoRR.} CoRR is large, but only a small part of its NITRC anatomical
data met our acquisition criterion (MPRAGE T1-weighted). All 377 qualifying
scans were used.

\paragraph{OASIS-1.} We kept scans from the 300 oldest OASIS-1 subjects (25--96 years), which
strengthens the sparsely sampled 80+ range. 

\pagebreak

\section{Learning and Interpretability Pipeline Algorithms} \label{supp:algos}
\begin{algorithm*}
\caption{Proposed BrainRotViT Framework}
\label{alg:brainrot}
\DontPrintSemicolon
\SetAlgoLined
\LinesNumbered
\KwIn{MRI volume $\mathcal{V}$, chronological age $a$, sex $s \in \{0,1\}$}
\KwOut{Predicted brain age $\hat{y}$, Brain-Age-Gap $\mathrm{BAG} = \hat{y} - a$}
\BlankLine
\textbf{Stage 1: Vision Transformer (ViT) Representation Learning}\;
\Indp
Sample $32$ evenly spaced sagittal slices $\{S_i\}_{i=1}^{32}$ from $\mathcal{V}$\;
\For{$i \gets 1$ \KwTo $32$}{
    Divide slice $S_i$ into non-overlapping patches, project them into embeddings, and add positional encodings\;
    Forward the embeddings through $L$ Transformer encoder layers consisting of multi-head self-attention and feed-forward sublayers with residual connections\;
    Apply a linear classification head with softmax to predict age-sex composite class for slice $S_i$\;
}
Compute cross-entropy loss:
\[
\mathcal{L}_{\mathrm{CE}} = -\frac{1}{B}\sum_{i=1}^{B}\sum_{c=1}^{C}y_{i,c}\log\hat{y}_{i,c}
\]
Train until convergence, then discard classification head and freeze $\mathrm{Enc}_{\text{ViT}}$\;
\Indm
\BlankLine
\textbf{Stage 2: Feature Map Construction}\;
\Indp
\For{$i \gets 1$ \KwTo $160$}{
    Extract 768-dimensional feature embedding $z_i$ from slice $S_i$ using the frozen ViT encoder\;
}
Stack embeddings in spatial order to form feature tensor:
\[
Z = [z_1, z_2, \dots, z_{160}]^\top \in \mathbb{R}^{160 \times 768}
\]
\Indm
\BlankLine
\textbf{Stage 3: Residual CNN Regression}\;
\Indp
Input feature map $Z$ into a residual CNN with three convolutional blocks\;
Each block applies convolution, SiLU activation, and normalization, with residual paths to preserve information\;
Flatten the final feature map $F_3$ into vector $h$\;
Pass $h$ through two fully connected layers with layer normalization, SiLU activation, and dropout, producing $h_2$\;
Concatenate sex information: $h_c = [h_2; s]$\;
Predict brain age with a linear layer:
\[
\hat{y} = W_3 h_c + b_3
\]
Compute mean squared error:
\[
\mathcal{L}_{\mathrm{MSE}} = \frac{1}{N}\sum_{i=1}^{N}(a_i - \hat{y}_i)^2
\]
\Indm
\BlankLine
\textbf{Output:} Predicted age $\hat{y}$ and Brain-Age-Gap $\mathrm{BAG} = \hat{y} - a$\;
\end{algorithm*}

\begin{algorithm*}[h]
\caption{Model Interpretability Pipeline}
\label{alg:interpretability}
\DontPrintSemicolon
\SetAlgoLined
\LinesNumbered
\KwIn{Trained ViT Encoder $\mathrm{Enc}_{\text{ViT}}$, Trained CNN Regressor $\mathrm{CNN}_{\text{reg}}$}
\KwIn{Validation set subjects $\{(\mathcal{V}_j, s_j)\}_{j=1}^{N_{\text{val}}}$, AAL3 Atlas $\mathcal{A}$ (Automated Anatomical Labeling v3, used for mapping voxels to brain regions)}
\KwOut{Aggregated 3D Attention Volume $\mathbf{A}_{\text{agg}}$, ROI Importance Scores}
\BlankLine
\textbf{Initialize:} Aggregated Volume $\mathbf{A}_{\text{agg}} \leftarrow \mathbf{0} \in \mathbb{R}^{S \times H \times W}$ (where $S=160$) \;
\BlankLine
\textbf{Iterate through each subject in the validation set} \\
\For{$j \gets 1$ \KwTo $N_{\text{val}}$}{
    \textbf{1. Guided Backpropagation of CNN Features} \\
    Generate feature map $\mathbf{Z}_j \in \mathbb{R}^{160 \times 768}$ from $\mathcal{V}_j$ using frozen $\mathrm{Enc}_{\text{ViT}}$\;
    Perform forward pass: $\hat{y}_j = \mathrm{CNN}_{\text{reg}}(\mathbf{Z}_j, s_j)$\;
    Compute guided gradient $\mathbf{G}_j = \nabla_{\mathbf{Z}_j} \hat{y}_j$, propagating only positive gradients through active (Si)LU units\;
    Normalize $\mathbf{G}_j \in \mathbb{R}^{160 \times 768}$\;
    \BlankLine
    \textbf{2. Projection, Fusion, and Slice-Level Heatmap Generation} \\
    \For{$i \gets 1$ \KwTo $160$ (each sagittal slice $S_i$ in $\mathcal{V}_j$)}{
        Pass slice $S_i$ through $\mathrm{Enc}_{\text{ViT}}$ to get patch embeddings $\mathbf{P}_i \in \mathbb{R}^{N_p \times 768}$ and CLS attention $\boldsymbol{\alpha}_{\text{CLS}, i} \in \mathbb{R}^{N_p}$\;
        Extract slice gradient vector $\mathbf{g}_{j,i} \in \mathbb{R}^{768}$ from $\mathbf{G}_j[i, :]$ \;
        Compute patch-wise importance scores: $\mathbf{s}_{j,i} = \mathbf{P}_i \cdot \mathbf{g}_{j,i}^T \in \mathbb{R}^{N_p}$\;
        Normalize scores: $\mathbf{s}^{\text{norm}}_{j,i} \leftarrow \text{Normalize}(\mathbf{s}_{j,i}, [0, 1])$ \;
        Normalize attention: $\boldsymbol{\alpha}^{\text{norm}}_{j,i} \leftarrow \text{Normalize}(\boldsymbol{\alpha}_{\text{CLS}, i}, [0, 1])$ \;
        Fuse maps: $\mathbf{f}_{j,i} = \mathbf{s}^{\text{norm}}_{j,i} \odot \boldsymbol{\alpha}^{\text{norm}}_{j,i}$ (element-wise multiplication)\;
        Reshape $\mathbf{f}_{j,i}$ to patch grid and upsample to slice resolution $(H, W)$ using bilinear interpolation $\rightarrow \mathbf{H}_{j,i}$\;
        $\mathbf{A}_{\text{agg}}[i, :, :] \leftarrow \mathbf{A}_{\text{agg}}[i, :, :] + \text{Normalize}(\mathbf{H}_{j,i})$\;
    }
}
\BlankLine
\textbf{3. 3D Volume Aggregation and ROI Identification} \\
Average across subjects: $\mathbf{A}_{\text{agg}} \leftarrow \mathbf{A}_{\text{agg}} / N_{\text{val}}$\;
Normalize the final 3D volume $\mathbf{A}_{\text{agg}}$\;
Align $\mathbf{A}_{\text{agg}}$ with AAL3 atlas $\mathcal{A}$ via cropping and affine registration\;
Initialize empty list $\text{ROI Scores}$\; 

\For{each region $R$ in $\mathcal{A}$}{
    Get voxel indices $\mathcal{V}_R$ for region $R$\;
    Calculate mean intensity: $I_R = \frac{1}{|\mathcal{V}_R|} \sum_{v \in \mathcal{V}_R} \mathbf{A}_{\text{agg}}[v]$\;
    Store $(R, I_R, |\mathcal{V}_R|)$ in $\text{ROI Scores}$\;
}
Group ROIs (e.g., all cerebellum parts) and compute weighted average intensity based on voxel count $|\mathcal{V}_R|$\;
\BlankLine
\Return $\mathbf{A}_{\text{agg}}$ and Grouped ROI Scores\;
\end{algorithm*}

\end{document}